\documentclass{ws-rv975x65}
\usepackage{ws-rv-van}             
\usepackage{makeidx}
\makeindex

\begin{document}

\chapter{Extracting Parts of 2D Shapes Using Local and Global Interactions Simultaneously\label{ch1}}

\author[S. Tari]{Sibel Tari\footnote{Work is supported in part by TUBITAK 108E015.}}
\index{skeleton|see{medial axis}}
\index{local symmetry|see{medial axis}}
\index{energy!non-local|see{non-local}}
\index{interaction!non-local|see{non-local}}
\index{PDE!curve evolution|see{curve evolution}}
\index{shape}
\index{interaction}
\index{shape!peripheral structure}
\index{shape!primitives}
\index{shape!boundary texture}
\index{shape!gross structure}
\index{shape!PDE based representation}
\index{parts of visual form}
\index{shape!parts}
\index{shape!parts!the least deformable}
\index{human visual system}

\index{features!region-based}
\index{features!boundary-based}

\index{non-local}
\index{medial axis}

\address{Middle East Technical University, Department of Computer Engineering,\\
Ankara, Turkey 06531, \\
stari@metu.edu.tr}

\begin{abstract}
Perception research provides strong evidence in favor of part based
representation of shapes in human visual system. Despite
considerable differences among different theories in terms of how
part boundaries are found, there is substantial agreement on that
the process  depends on many local and global geometric factors.
This poses an important challenge from the computational point of
view. In the first part of the chapter, I
present a novel decomposition method by taking both local and global
interactions within the shape domain into account. At the top of the partitioning hierarchy, the
shape gets split into two parts capturing, respectively, the gross
structure and the peripheral structure. The gross structure may be
conceived as the least deformable part of the shape which remains
stable under visual transformations.  The peripheral structure
includes  limbs,  protrusions, and  boundary texture. Such a
separation is in accord with the behavior of the artists who start
with a gross shape and enrich it with details. The method is
particularly interesting from the computational point of view as it
does not resort to any geometric notions (e.g. curvature, convexity)
explicitly. In the second part of the chapter, I relate the new method to PDE
based shape representation schemes.
\end{abstract}

\section{Introduction}\label{sec1.1}
Perception research provides strong evidence in favor of part based
representation of shapes in human visual system~\cite{BarenholtzFeldman}. Recent work using
single cell recordings in area V4 in the primate visual cortex
supports part based coding at intermediate levels~\cite{Pasu2002}. Many influential
shape representation theories either explicitly or implicitly assume
an organization in terms of constituent components. In
Binford~\cite{Binford71}; Marr and Nishiara~\cite{MarrNish}; and
Biederman~\cite{Biederman}, shape is represented via pre-defined
simple shapes (primitives) and their spatial layout.  Medial axis or
local symmetry set (one of the  influential ideas in shape
representation) is closely connected with the notion of parts.
 Large number of\vadjust{\pagebreak} computational methods utilize medial axis to infer part
structure~\cite{Zeng07,Mi07,Svensson02,Arcelli97,Aslan05,Rom}.
In a recent work, Super~\cite{Super07} presents a quite successful part based recognition scheme.

There are powerful theories accompanied by computational
implementations on what constitutes a good partition without
resorting to predefined shapes or categorical units,
e.g.~\cite{Cohen07,HofmannRichards,Latecki99,Renninger03,Siddiqi95,SiddiqiTresness,Wagemans99}.
Despite considerable differences among different theories in terms
of how part boundaries are found, there is substantial agreement
on that the process depends on many geometric factors both at the
global and the local
levels~\cite{Cohen07,BarenholtzFeldman,SiddiqiTresness,Wagemans99,Burbeck95,Navon77}.
Indeed, part of the difficulty in devising computational mechanisms
for shape decomposition lies in the difficulty in integrating local
boundary  concavities with  non-local shape descriptions. Recent
works  by Mi and DeCarlo~\cite{Mi07} and Zeng et al.~\cite{Zeng07}
address this challenging issue using contour curvature and local
symmetry axis simultaneously. In another recent work, Xu, Liu and
Tang~\cite{Xu09} combine effectiveness of both local and global
features for matching shapes. One of the successful  recognition
schemes, the shape context by Belongie, Malik and
Puzicha~\cite{Belongie00}, is based on  quantifying non-local
interactions among boundary points. Moreover, some of the  successful
methods for partitioning images into pixel
groups~\cite{Ncut,NLMeans} are based on non-local relations.
\index{features!local}
\index{features!global}
\index{non-local}

The rest of the chapter is organized as follows. In \ref{S1}, the new decomposition method is presented.
The new method takes both local and global interactions within the shape domain into account. At the top of the
partitioning hierarchy, the shape gets split into two parts capturing, respectively,
 the gross structure and the peripheral structure. The
gross structure may be conceived as the least deformable part of the
shape which remains stable under visual transformations.  The
peripheral structure includes  limbs,  protrusions, and
boundary texture. Such a separation is in accord with the experimental
studies suggesting that the global shape is processed before the
details~\cite{Navon77,Burbeck95}; and with the behavior of the
artists who start with a gross shape and enrich it with
details~\cite{Koenderink82}.

\index{shape!global}
\index{shape!PDE based representation}
\index{medial axis}
In \ref{S2}, the new method  formulated in a discrete setting is related to
PDE  based shape representation approach with particular emphasis given to a recent
skeleton based scheme  which I had previously developed with my student Cagri Aslan\cite{Aslan05,Aslan08,Aslantez}.

\section{The New Method}
\label{S1}
\index{energy!minimizer}
\index{energy!region-based}
\index{energy!boundary-based}
\index{emergent structures}

The basic idea is to create a field within the shape domain with emergent structures
capturing the parts automatically.
This field is
computed by minimizing an energy which captures both local and
global; and both region and boundary based interactions among
shape points.

Let us define  a function $\omega$ defined on a  discrete shape
domain $\Omega$ as the minimizer of an energy $E(\omega)$. Let this
energy be a sum of a region based energy $E_{Reg}(\omega)$ and a
boundary based energy $E_{Bdy}(\omega)$; and the region based energy
$E_{Reg}(\omega)$ is a sum of  two energies which model the global
({\sl G}) and the local ({\sl L}) interactions within the shape
domain $\Omega$:
\begin{eqnarray}
E(\omega) &=& E_{Reg}(\omega) + w_{Bdy} E_{Bdy}(\omega) \nonumber \\
      &=& E_{Reg}^G(\omega) + E_{Reg}^L(\omega) + w_{Bdy} E_{Bdy}(\omega)  \label{eq:eq1}
\end{eqnarray}

Assuming that each of the three terms  in (\ref{eq:eq1}) can be expressed as a sum of
energies defined at each pixel $i,j$, the following form is
obtained:
\begin{equation}
 E(\omega)= \sum_{i,j\in \Omega}
E_{Reg}^G(\omega_{i,j}) + E_{Reg}^L(\omega_{i,j}) +w_{Bdy}
E_{Bdy}(\omega_{i,j}) \label{eq:energy}
\end{equation}

In the absence of any specific purpose or bias, equal importance can
be given to both the region and the boundary by setting $w_{Bdy}=2$.
For computational reasons, it is preferable to choose a quadratic
form for each energy.  Let $E_{Reg}^G(\omega_{i,j})$   be:
\begin{equation}
E_{Reg}^G(\omega_{i,j}) = \frac{1}{\left| \Omega \right|}
\left(\sum_{k,l \in \Omega} \omega_{k,l} \right)^2 \label{eq:G}
\end{equation}

The minimizer $\omega_{i,j}$ of (\ref{eq:G}) satisfies:
\begin{equation}
\frac{1}{\left| \Omega \right|}\sum_{k,l \in \Omega} \omega_{k,l}=0
\label{eq:turevG}
\end{equation}

The condition satisfied by the minimizer of
$E_{Reg}^G(\omega_{i,j})$ is independent of the pixel location
$(i,j)$ and it  explicitly states that the global average over the
shape domain should be zero. 
It forces $\omega$ to attain both positive and negative values
within the shape domain $\Omega$. This behavior when complemented with
the behavior induced by the other terms will be shown to be quite
instrumental in obtaining robust and parameter free separation of
the gross structure and
the peripheral structure. %

The second term of the region based energy,  $E_{Reg}^L$, has the
following form:
\begin{equation}
E_{Reg}^L(\omega_i,j) = - \left(  \omega_{i+1,j} \cdot \omega_{i-1,j} +
\omega_{i,j+1} \cdot \omega_{i,j-1} \right)
\label{eq:L}
\end{equation}

 Clearly, $E_{Reg}^L(\omega_i) $   is
minimized when the values of the neighboring pixels are similar.
Thus the second term of the energy
 imposes  smoothness on $\omega$. The condition for the minimizer of  $E_{Reg}^L$
is not  straightforward to calculate  as that of  $E_{Reg}^G$.
First, a local continuous approximation at location $i,j$ is
considered with the help of Taylor series. Second, the Gateaux
derivative is calculated. Third, the Gateaux derivative is
discretized and set to {\sl zero}, to obtain the condition for the
minimizer:
\begin{equation}
(-2+ 4)*\omega_{i,j} - \omega_{i-1,j}
-\omega_{i+1,j}-\omega_{i,j-1}-\omega_{i,j+1} =0
\label{eq:turevL}
\end{equation}

\index{interaction!pairwise}
\index{shape!minima rule}
The boundary based energy $E_{Bdy}(\omega)$ is chosen as a  measure
of pairwise interaction between two properly chosen boundary points
such that the pairs indicate parts. One motivation to consider
pairwise interaction between two boundary points comes from the
well-established minima rule~\cite{HofmannRichards,Cohen07} which is
used in many computational procedures for shape decomposition.
However, computationally, it is not an easy task to model pairwise
interactions among boundary points.  These interactions are neither
local nor global. A simple alternative is constructed by exploiting
the connection between the concept  of pairwise interaction among
boundary points and the shape skeleton~\cite{Blum73}. With the help
of this connection, $E_{Bdy}$ is expressed as an energy defined over
the entire shape domain.
\index{medial axis!grass-fire}
\index{medial axis!Blum}
\index{distance transform}

\begin{figure}[!h]
\centering
\includegraphics[width=5cm]{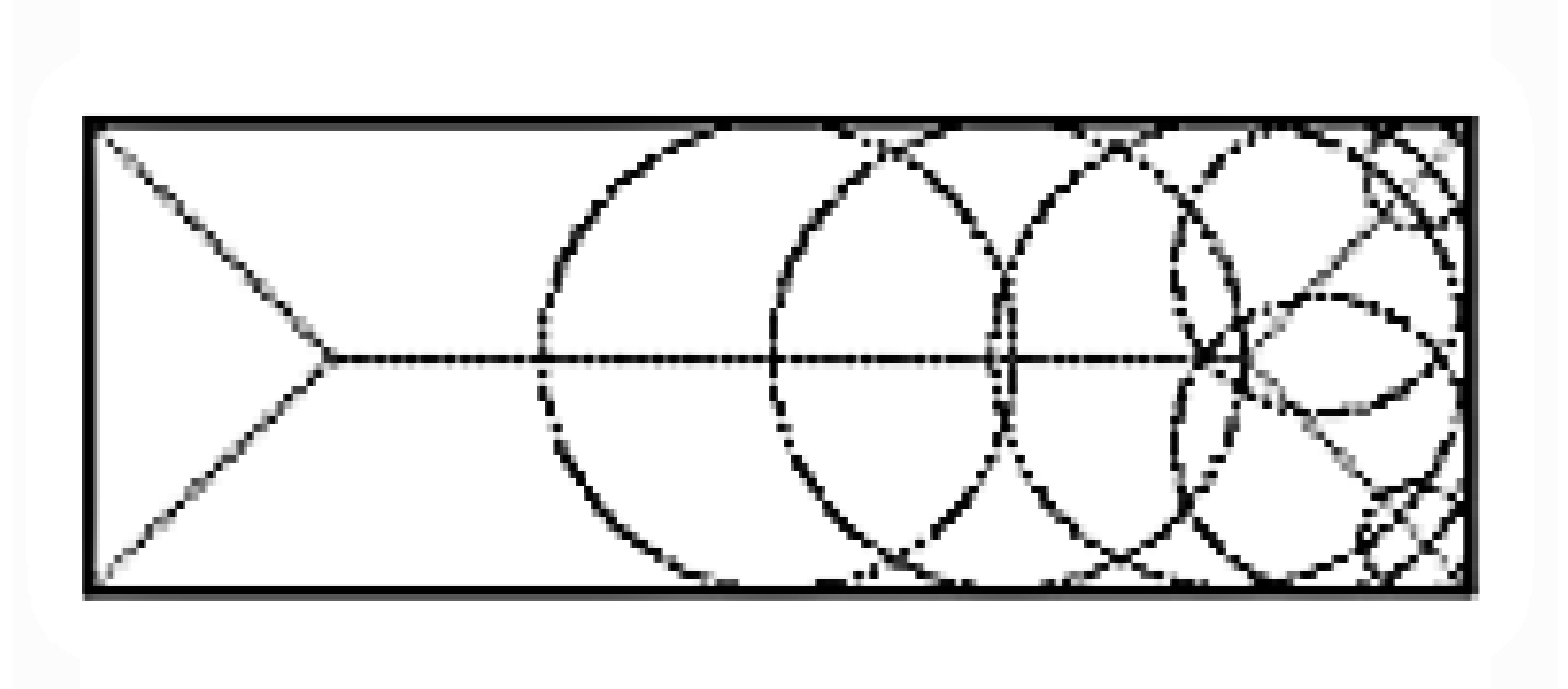} 
\vglue -4pt
\caption {Each skeleton point is formed as a result of interaction
between two or more boundary points. Skeleton points are the ones
which are equidistant from at least two boundary
points~\cite{Blum73}. } \label{fig:Blum}
\end{figure}

The connection can be  explained with the help of the grass-fire
model by Blum~\cite{Blum73}. Assume that one initiates fire fronts
at time $t = 0$ along all the points of the shape boundary and lets
these fronts propagate toward the center of the shape at a uniform
speed. The locus of  points where these fronts meet and extinguish 
defines the shape skeleton. Each skeleton point is formed as a
result of interaction between two or more boundary points. During the course
of the propagation, the time $t$ may be thought of as a function
$t_{i,j}$ defined over the shape domain by setting the value to the
time when the propagating fronts passes through the pixel $(i,j)$.
The value of $t_{i,j}$  will be proportional to the minimum distance
from $(i,j)$ to the nearest boundary points. Skeleton  points are
the ones which are equidistant from at least
two boundary points~\cite{Blum73}~(Fig.~\ref{fig:Blum}.)
%

This insight  enables the expression of   $E_{Bdy}(\omega)$ as a
quadratic energy defined over the entire shape region $\Omega$ as
 the following form: 
\begin{equation}
E_{Bdy}(\omega_{i,j}) = \left( \omega_{i,j} - t_{i,j} \right)^2
\label{eq:B}
\end{equation}

It is straightforward to calculate the condition for the minimizer
of the $E_{Bdy}(\omega_{i,j})$ given in (\ref{eq:B}) as:
\begin{equation}
\left( \omega_{i,j} - t_{i,j} \right) =0
\label{eq:turevB}
\end{equation}

In the absence of other terms, (\ref{eq:turevB}) states that the field $\omega$ should be
equal to the
distance transform defined over $\Omega$. Putting together all three
competing terms, (\ref{eq:turevG},\ref{eq:turevL},\ref{eq:turevB}),
the first order derivative w.r.t. each unknown $\omega_{i,j}$ takes
the following form:
\begin{eqnarray}
\frac{ \partial E}{\partial \omega_{i,j}}&=& \frac{1}{\left|\Omega\right|} \left( \sum_{k,l \in \Omega} \omega_{k,l} \right) \nonumber + (w_{Bdy}-2 +4)\omega_{i,j} - w_{Bdy}t_{i,j} \nonumber\\
 &-&   \omega_{i-1,j} -\omega_{i+1,j}-\omega_{i,j-1}-\omega_{i,j+1}
\label{eq:turev_all}
\end{eqnarray}

Setting the derivative equal to {\sl zero} yields that the minimizer
 of $E(\omega)$ satisfies the following condition at each pixel $({i,j})$:
\pagebreak

\noindent
\begin{eqnarray}
 w_{Bdy}t_{i,j} &=& (w_{Bdy}-2 +4)\omega_{i,j} -\omega_{i-1,j} -\omega_{i+1,j}
-\omega_{i,j-1}-\omega_{i,j+1} \nonumber \\
&+& \frac{1}{\left|\Omega\right|} \left( \sum_{k,l \in \Omega} \omega_{k,l}\right)
\label{eq:method_withw}
\end{eqnarray}

Recall that, in the absence of any specific purpose or bias, the
weight $w_{Bdy}$ is set to $2$ to give equal importance to both the
region and the  boundary. Thus,   $\omega_{i,j}$ is computed by
solving (\ref{eq:method}) given below, at all the pixels simultaneously,
assuming that the values are zero at the boundary pixels.
\begin{equation}
 t_{i,j} = 4\omega_{i,j} -\omega_{i-1,j} -\omega_{i+1,j}-\omega_{i,j-1}-\omega_{i,j+1}
 + \frac{1}{\left|\Omega\right|} \left( \sum_{k,l\in \Omega} \omega_{k,l}\right)
\label{eq:method}
\end{equation}

\begin{figure}[t]
\vglue -2pt
\centering
{\footnotesize
\begin{tabular}{cc}
\includegraphics[height=4cm]{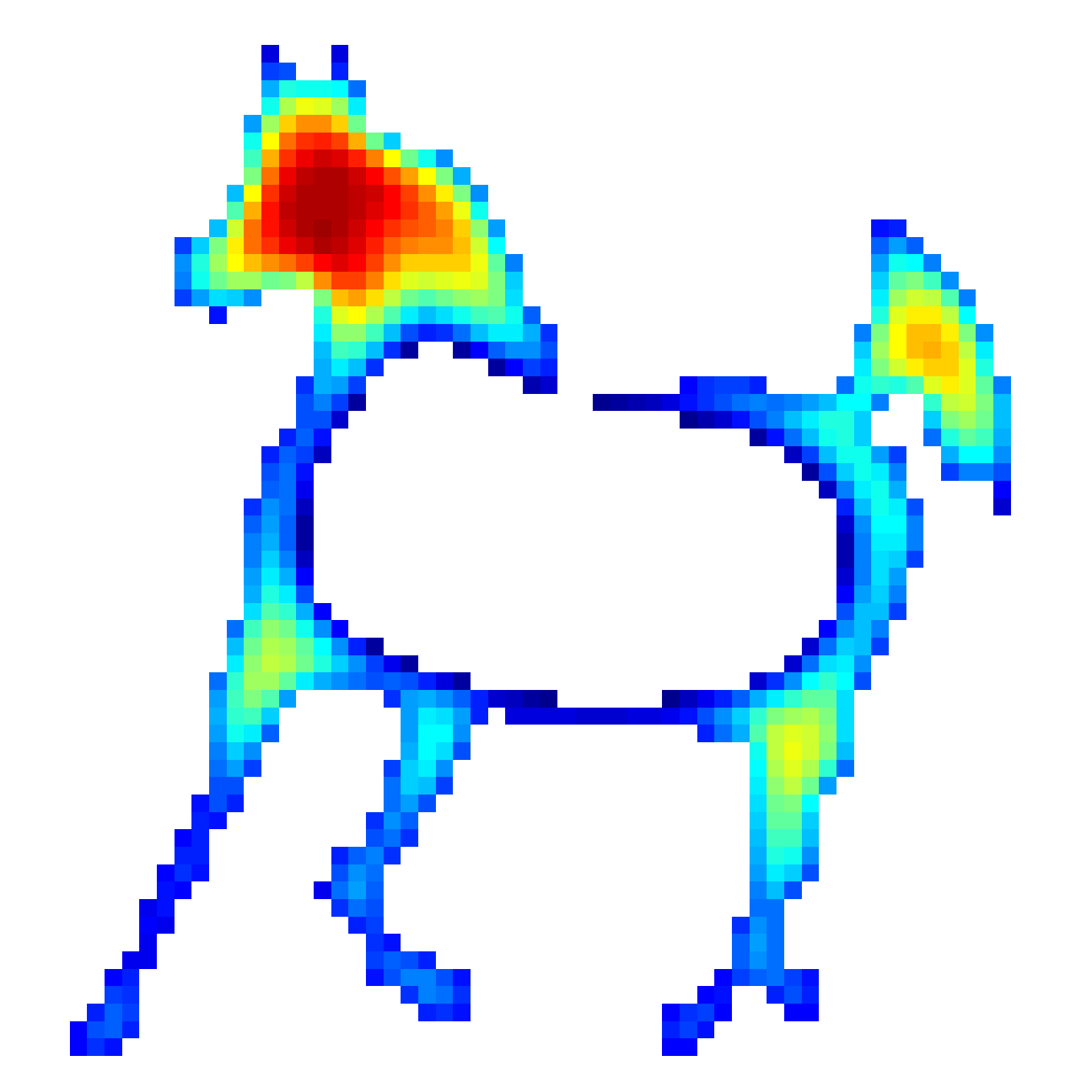}&
\includegraphics[height=4cm]{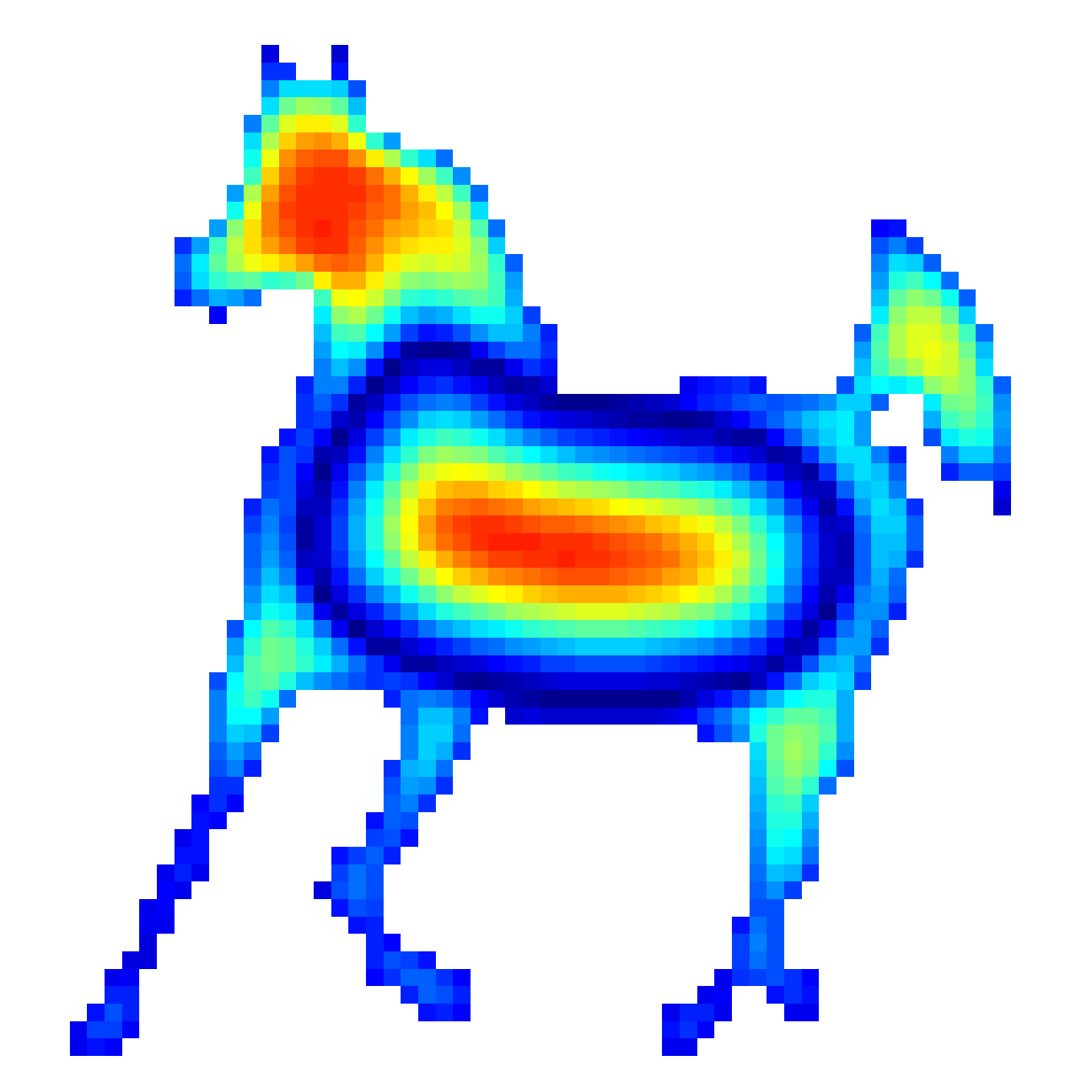}\\
\includegraphics[height=4cm]{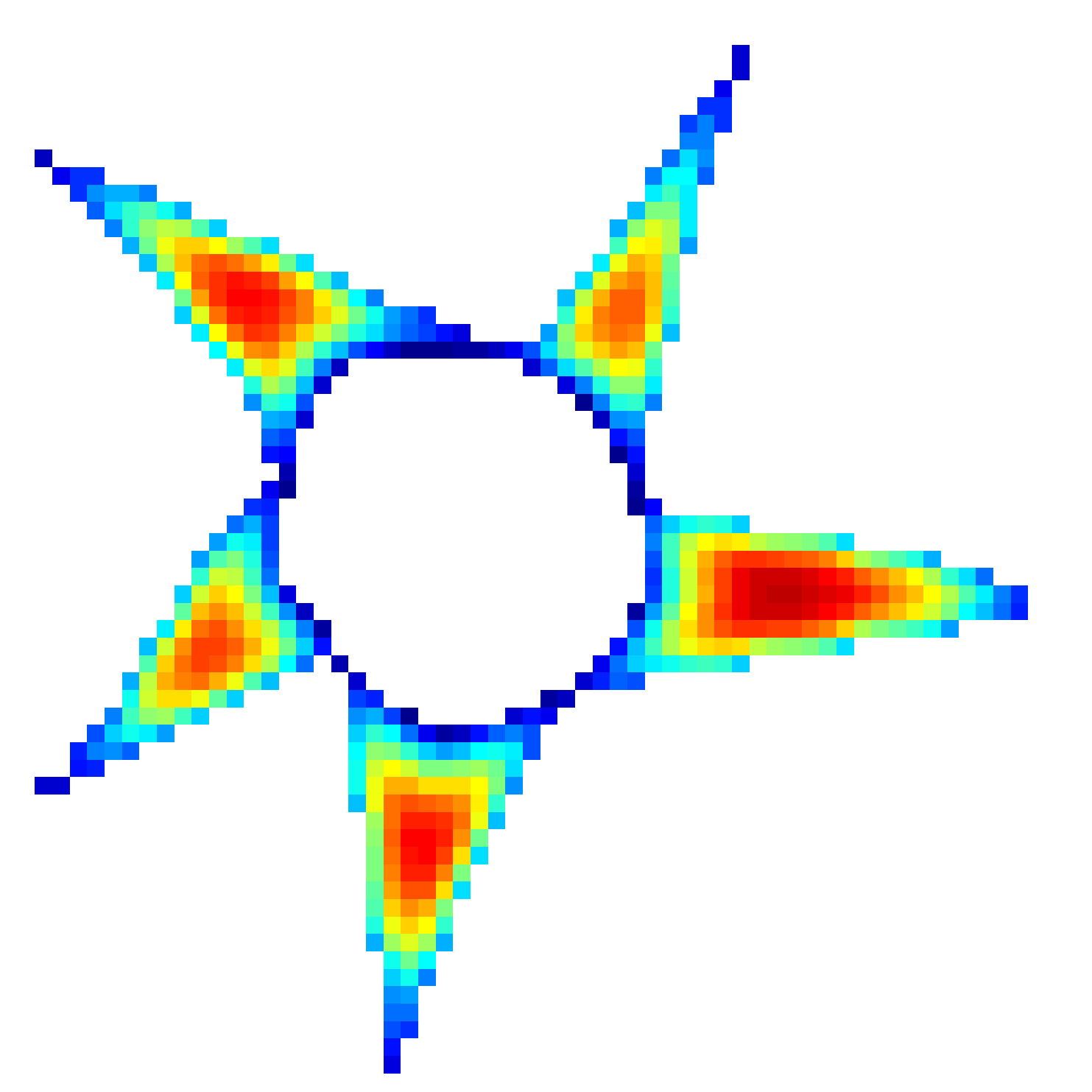}&
\includegraphics[height=4cm]{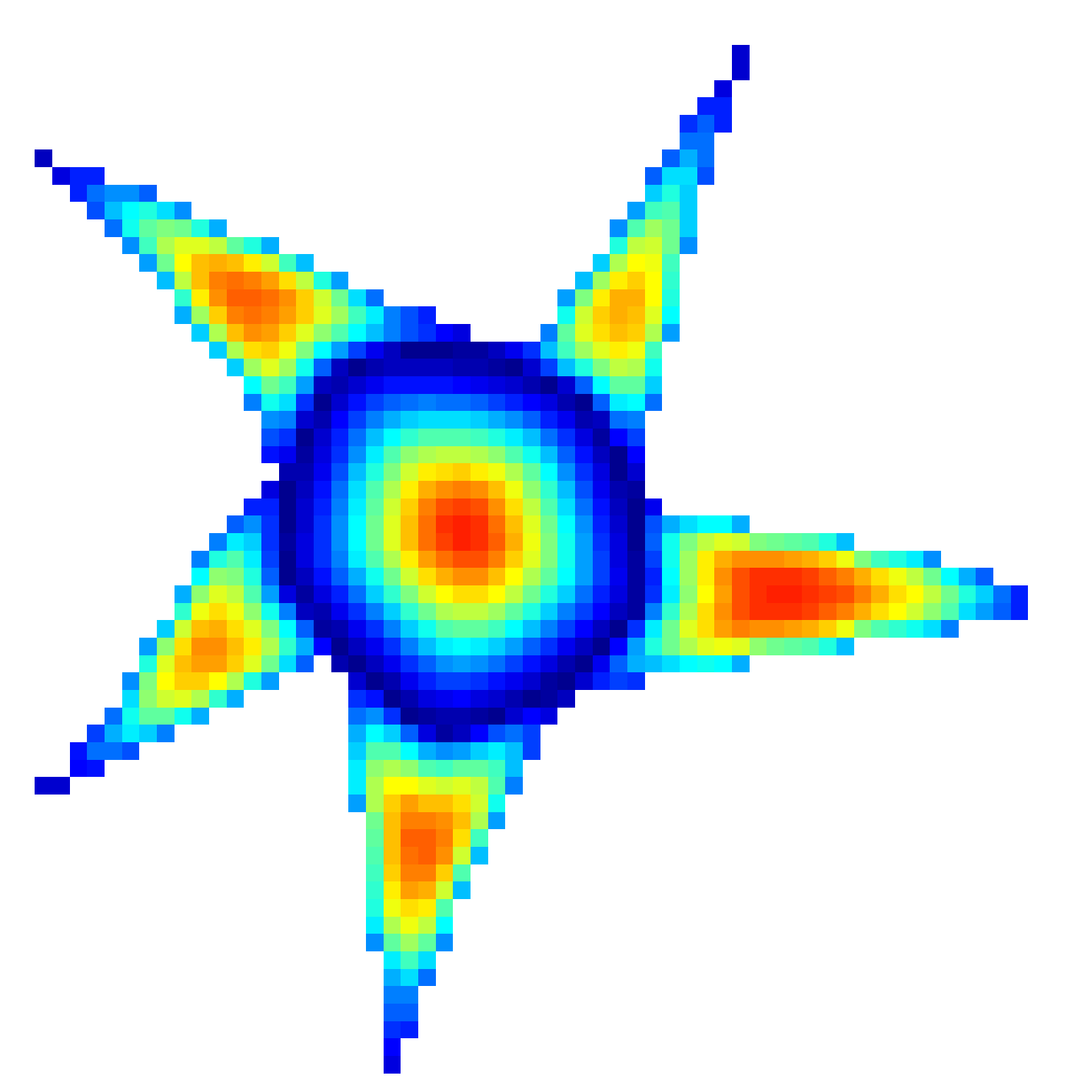}\\
(a) & (b)
\end{tabular}}
\caption {The field $\omega$ computed by
solving~(\ref{eq:method}).
 For visualization purposes, the values are
normalized.  (a) The restriction of $\omega$ to  areas where its
values are negative. This part denotes the peripheral structure {\sl
i.e.} protrusions, limbs, and boundary texture. The removed inner
part on which the values are positive is the gross structure. (b)
The absolute value of $\omega$.} \label{fig:w}

\vglue 12pt

\centering
\begin{tabular}{cc}
\includegraphics[height=4.4cm]{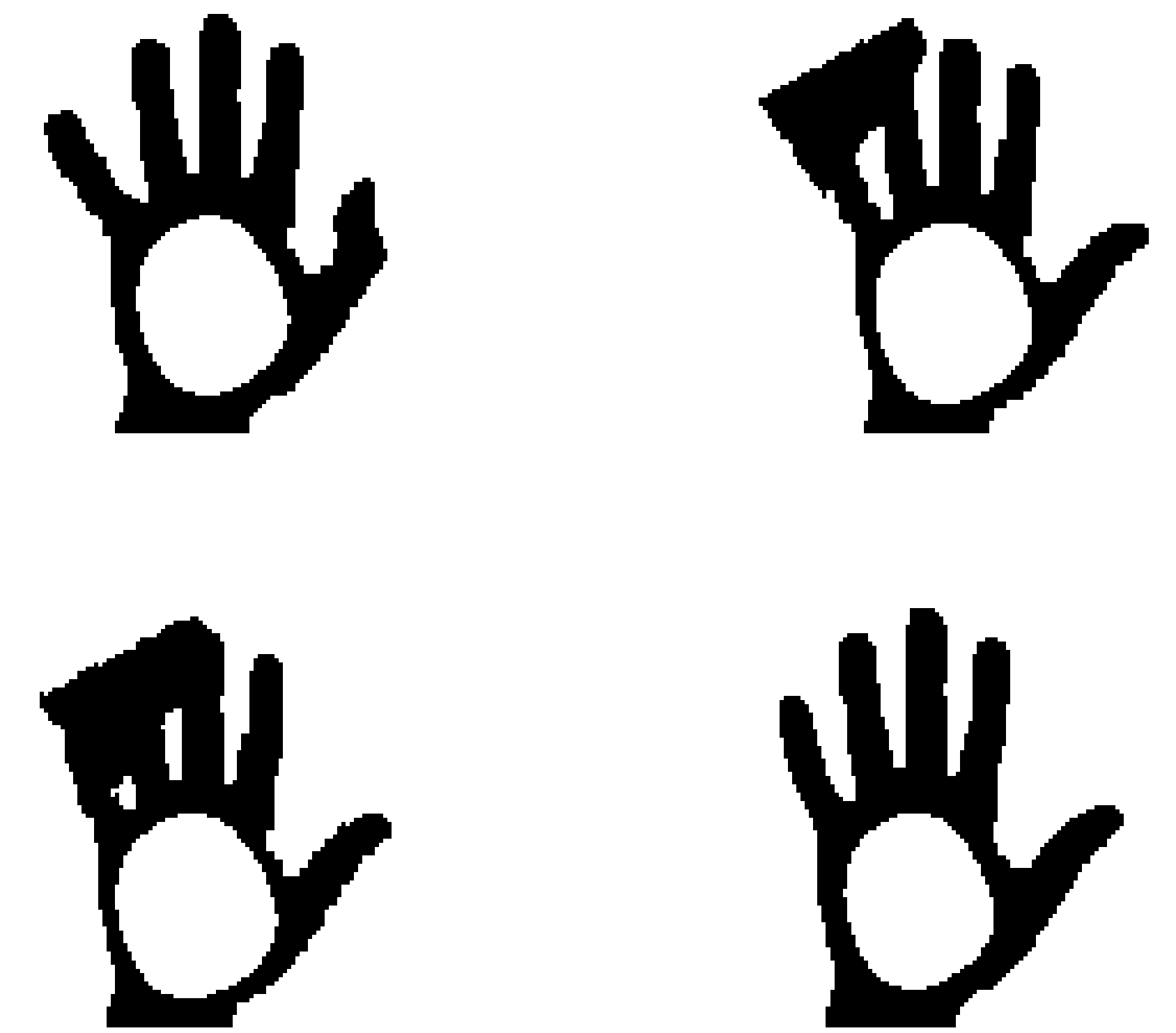}&
\includegraphics[height=4.4cm]{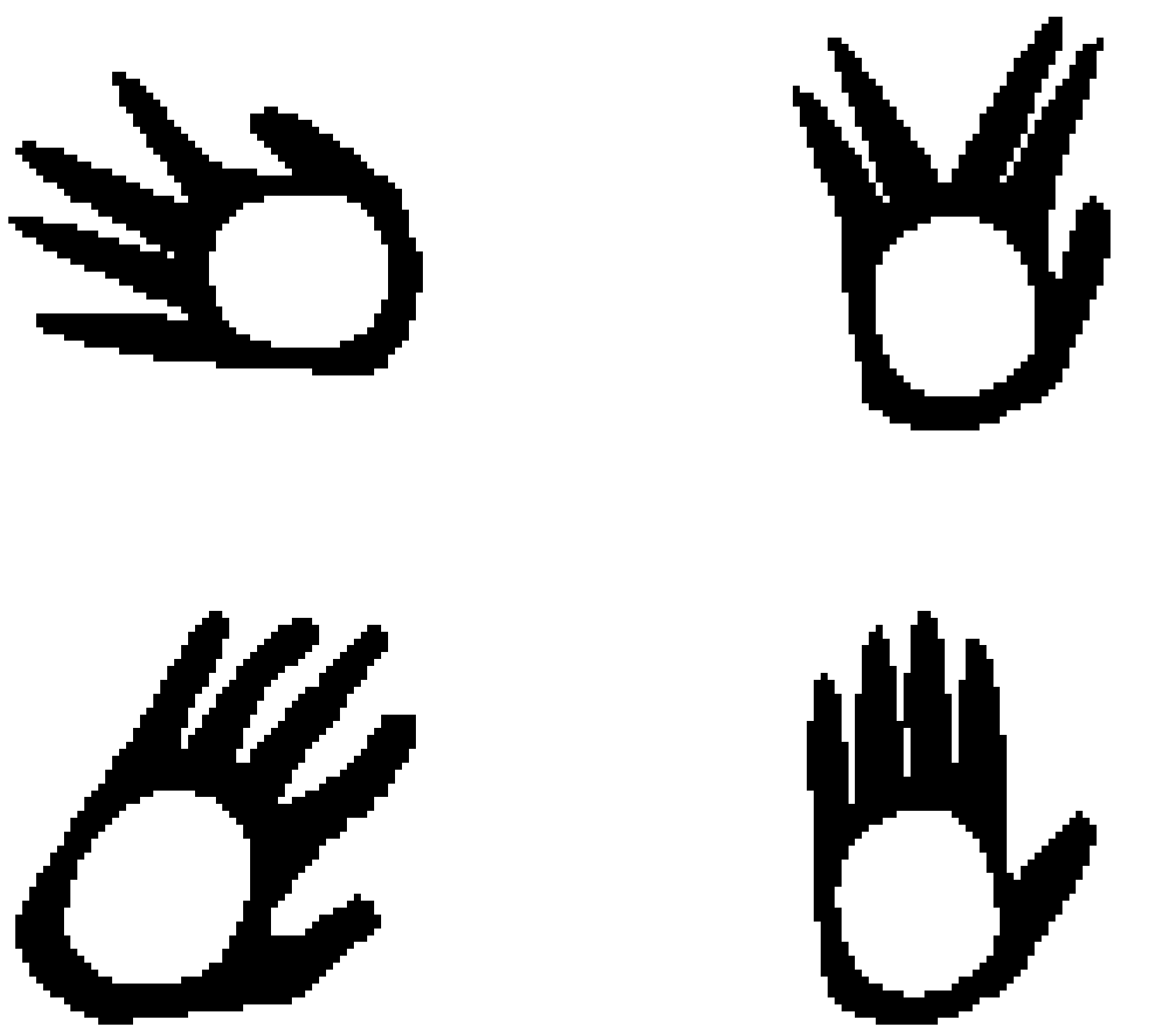}
\end{tabular}
\caption {The areas where $\omega<0$ is shown in black. The gross
structure (inner white blob) may be conceived as the least
deformable part of the shape. It remains stable under a variety of
changes. } \label{fig:handgross}
\end{figure}

The field $\omega$,  computed by solving (\ref{eq:method}),  is
depicted in Fig.~\ref{fig:w} for two sample shapes.  It attains both
negative and positive values. This behavior is dictated by the
global region energy $E_{Reg}^G$ which explicitly states that the
global average over the shape domain should be small. 

In Fig.~\ref{fig:w} (a), the restriction of $\omega$ where it is
negative is displayed. This set is denoted by $\Omega^-$. It
captures the peripheral structure (protrusions, limbs, boundary
texture). The darkest blue denotes {\sl zero}; the darkest red
denotes
 the lowest negative value.
The removed inner part is the part on which $\omega$ is positive;
and  is denoted by $\Omega^+$. This blob-like part captures the
gross structure. The gross structure is the least deformable part of
the shape which remains stable under a variety of visual changes as
demonstrated in Fig.~\ref{fig:handgross} using eight different instances of a hand silhouette.

In Fig.~\ref{fig:w} (b), the absolute value of $\omega$ is displayed
on the entire shape domain.  For visualization purposes, the
negative values and the positive values are normalized, separately,
to the $[0,1]$ interval. The darkest blue denotes {\sl zero}; the
darkest red denotes {\sl one}. Notice that various local maxima
capture intuitive parts such as the body, the head, the tail, and
the legs of the horse. These parts can be easily extracted by
considering a growth starting from each local maxima. Separate
growths from each pair of neighboring maxima meet at a saddle point.

\subsection{Experimental Results and Discussion}

The method is is discussed via a set of highly illustrative
examples. These examples are silhouettes collected from various
sources~\cite{LEMS,Aslan05,Gorelick06,Zeng07}. Some of the original
silhouettes  are modified  by the author to obtain shapes with
holes, missing and/or extra parts.

\begin{figure}[t]
\centering
\begin{tabular}{ccc}
\includegraphics[height=3.5cm]{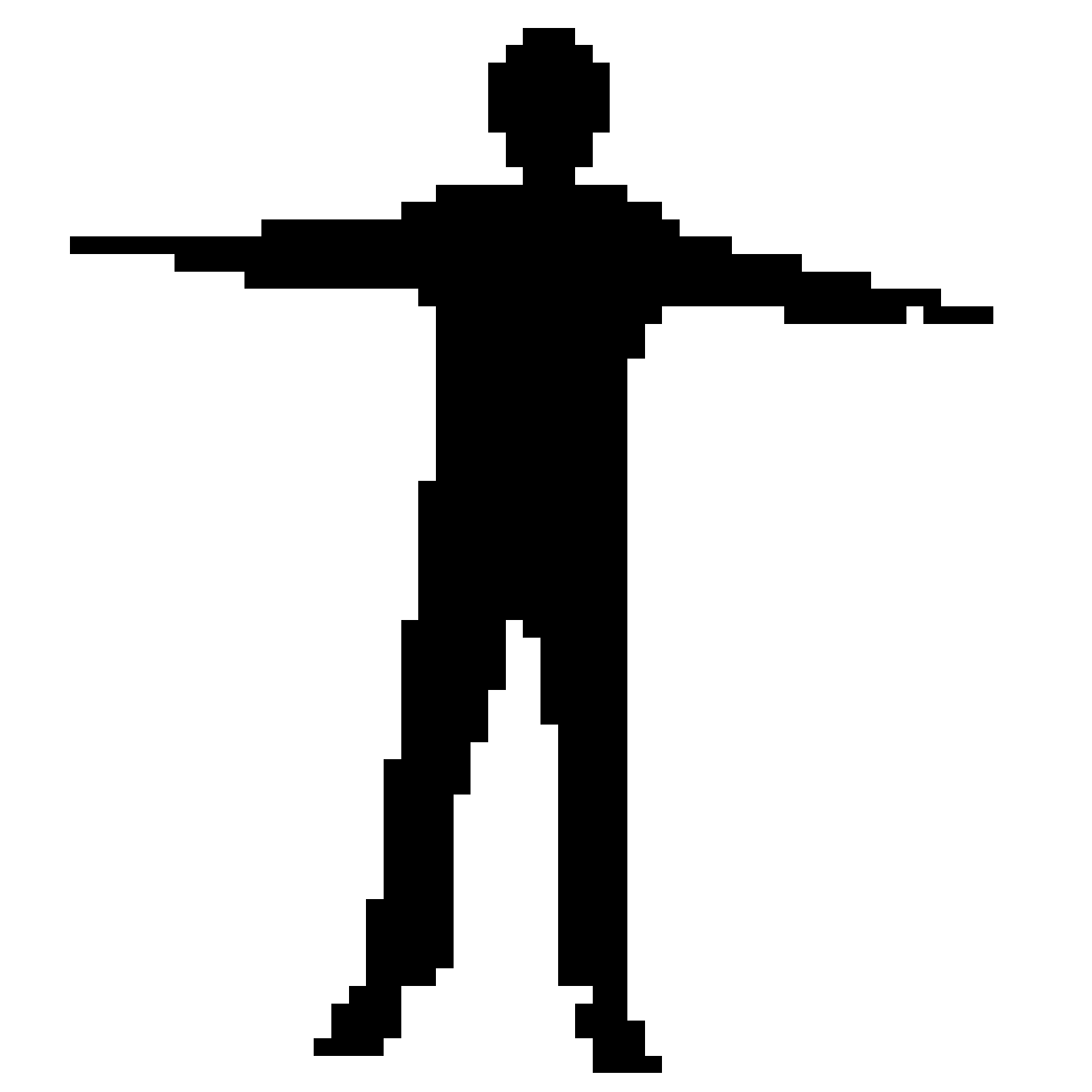}&
\includegraphics[height=3.5cm]{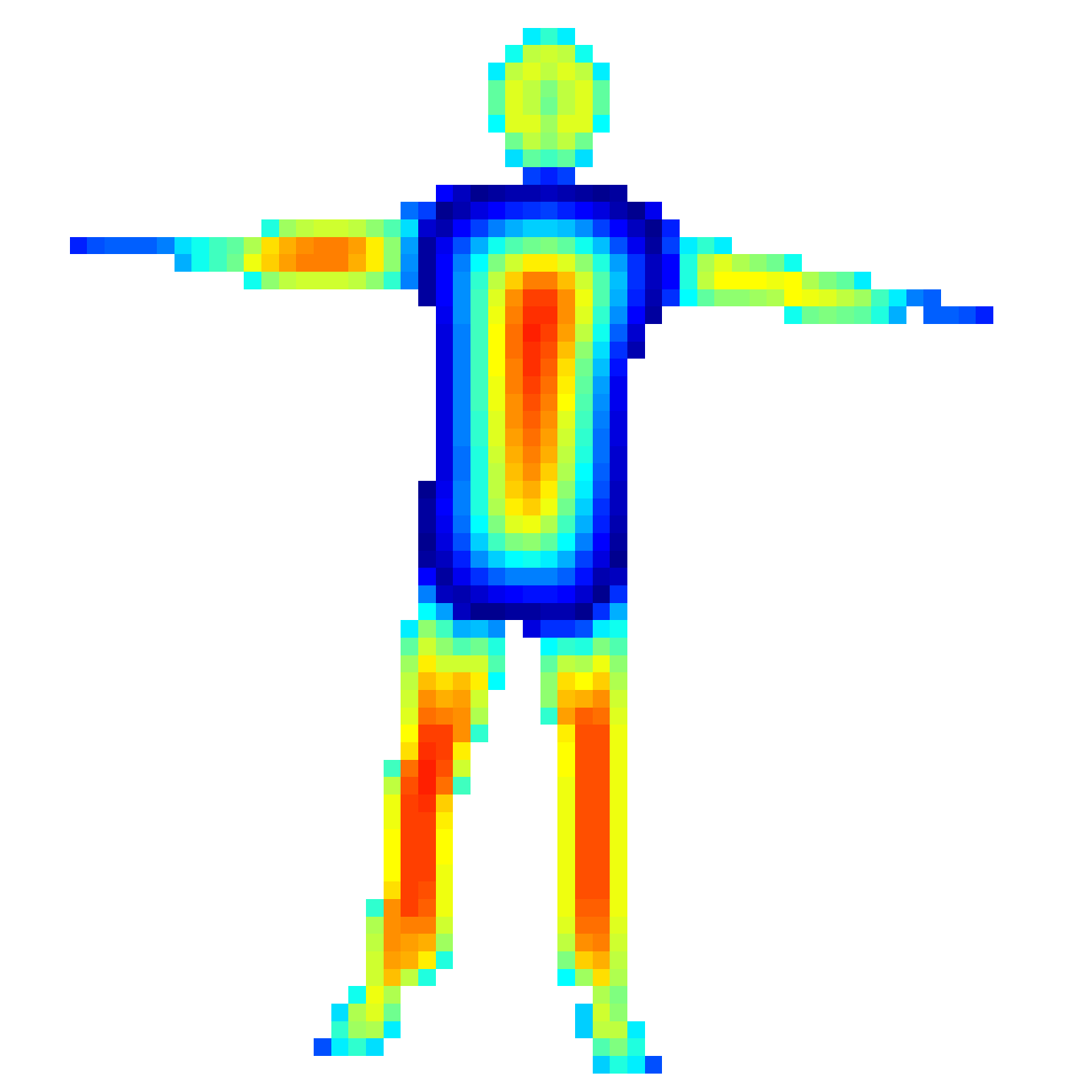}&
\includegraphics[height=3.5cm]{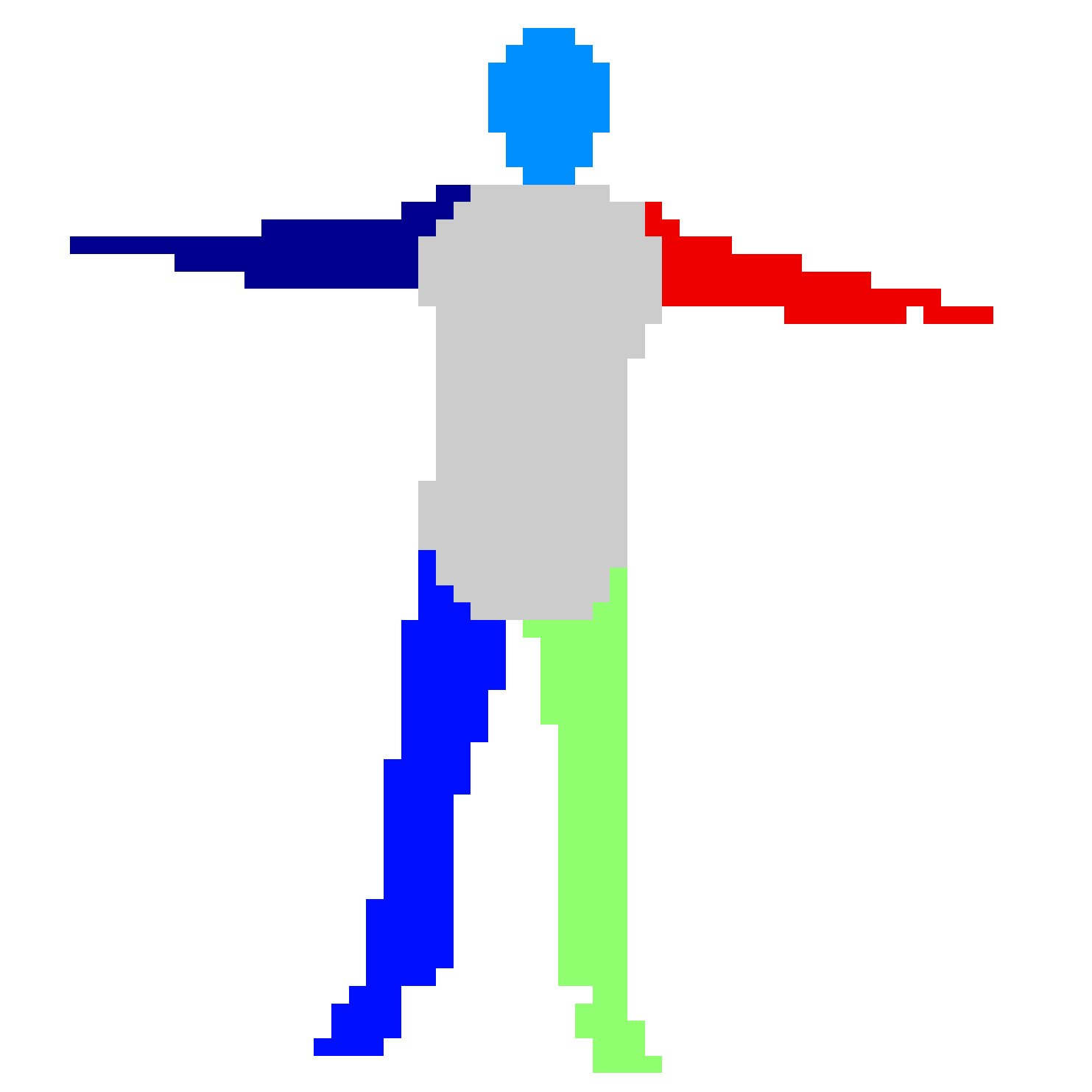}\\
\includegraphics[height=3.5cm]{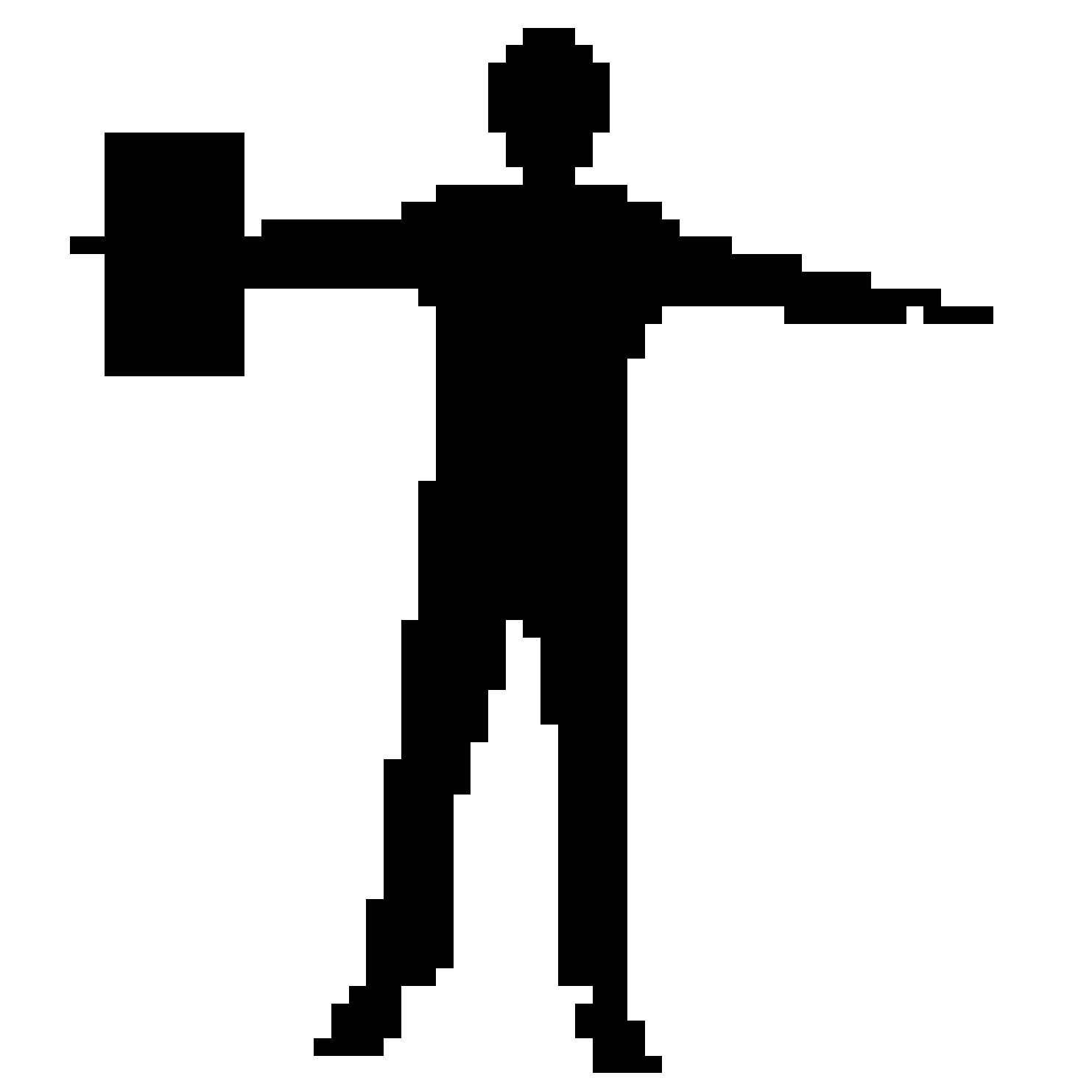}&
\includegraphics[height=3.5cm]{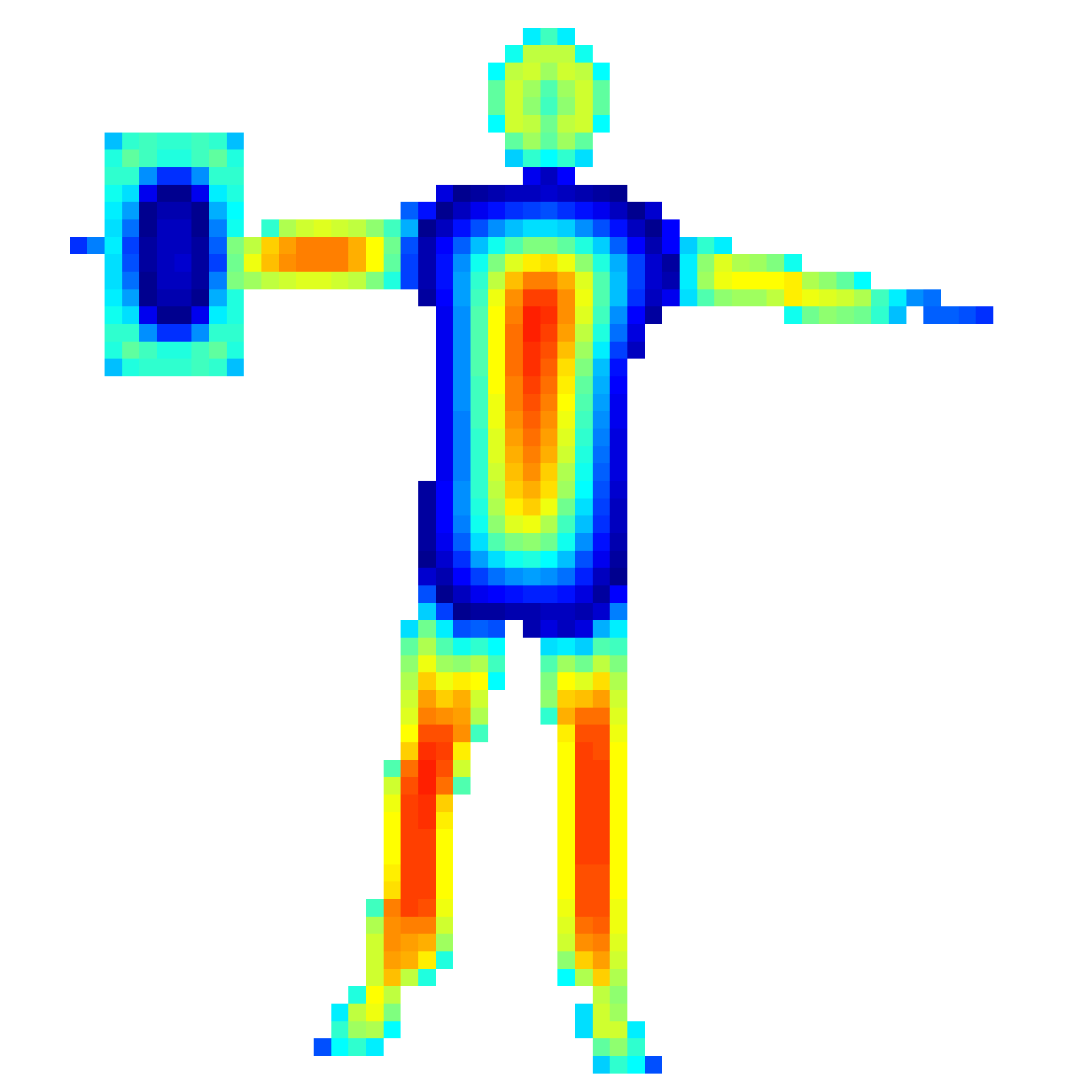}&
\includegraphics[height=3.5cm]{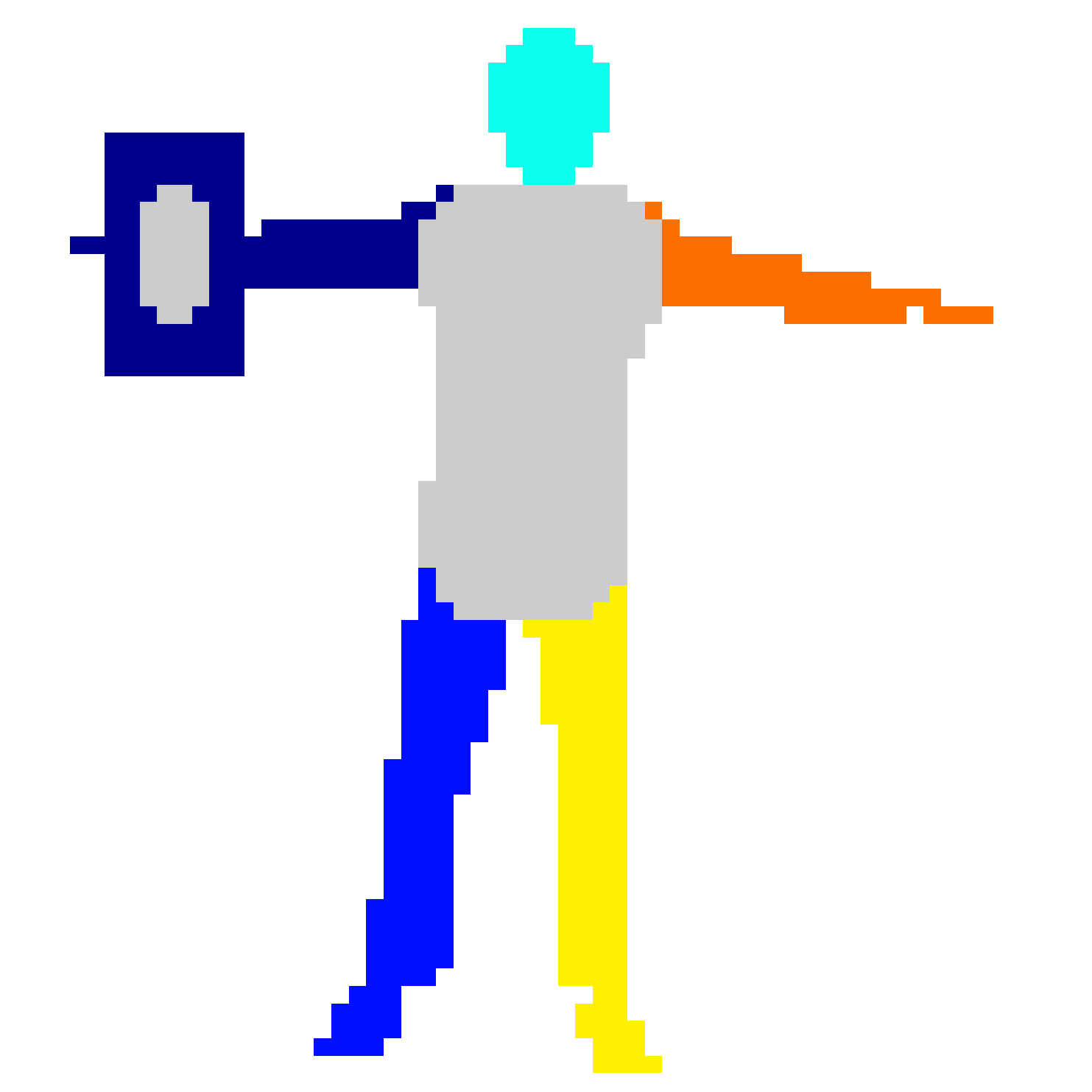}\\
\includegraphics[height=3.5cm]{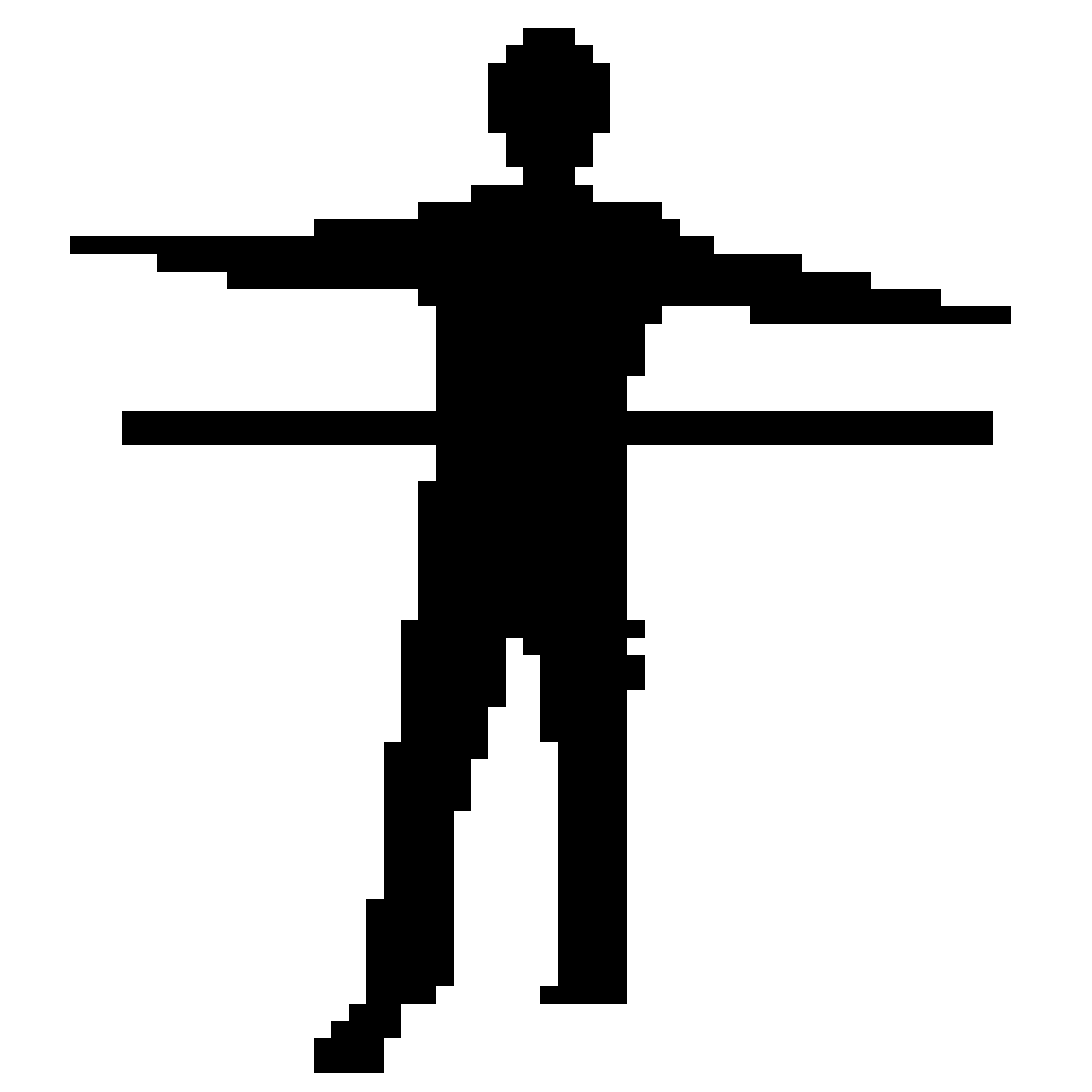}&
\includegraphics[height=3.5cm]{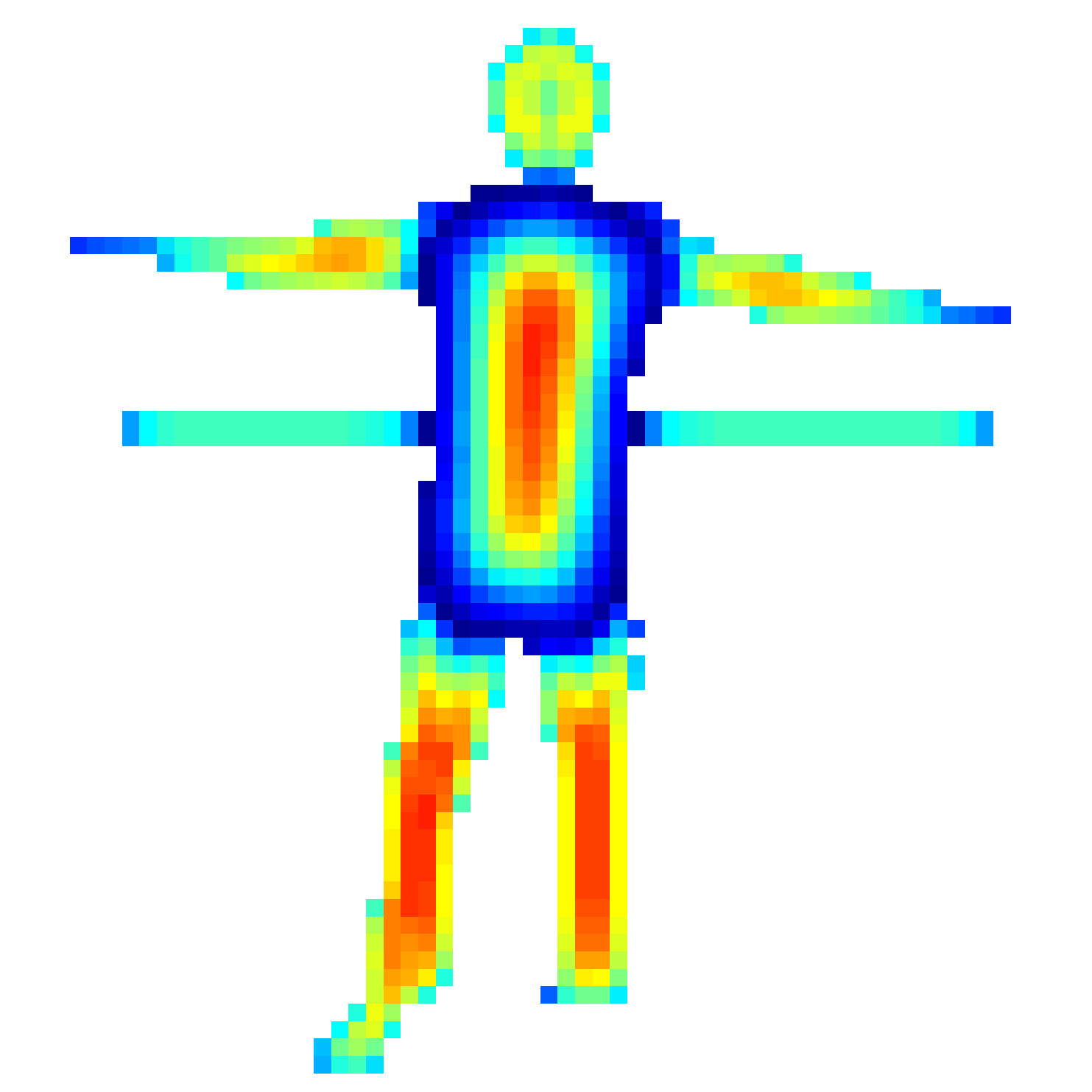}&
\includegraphics[height=3.5cm]{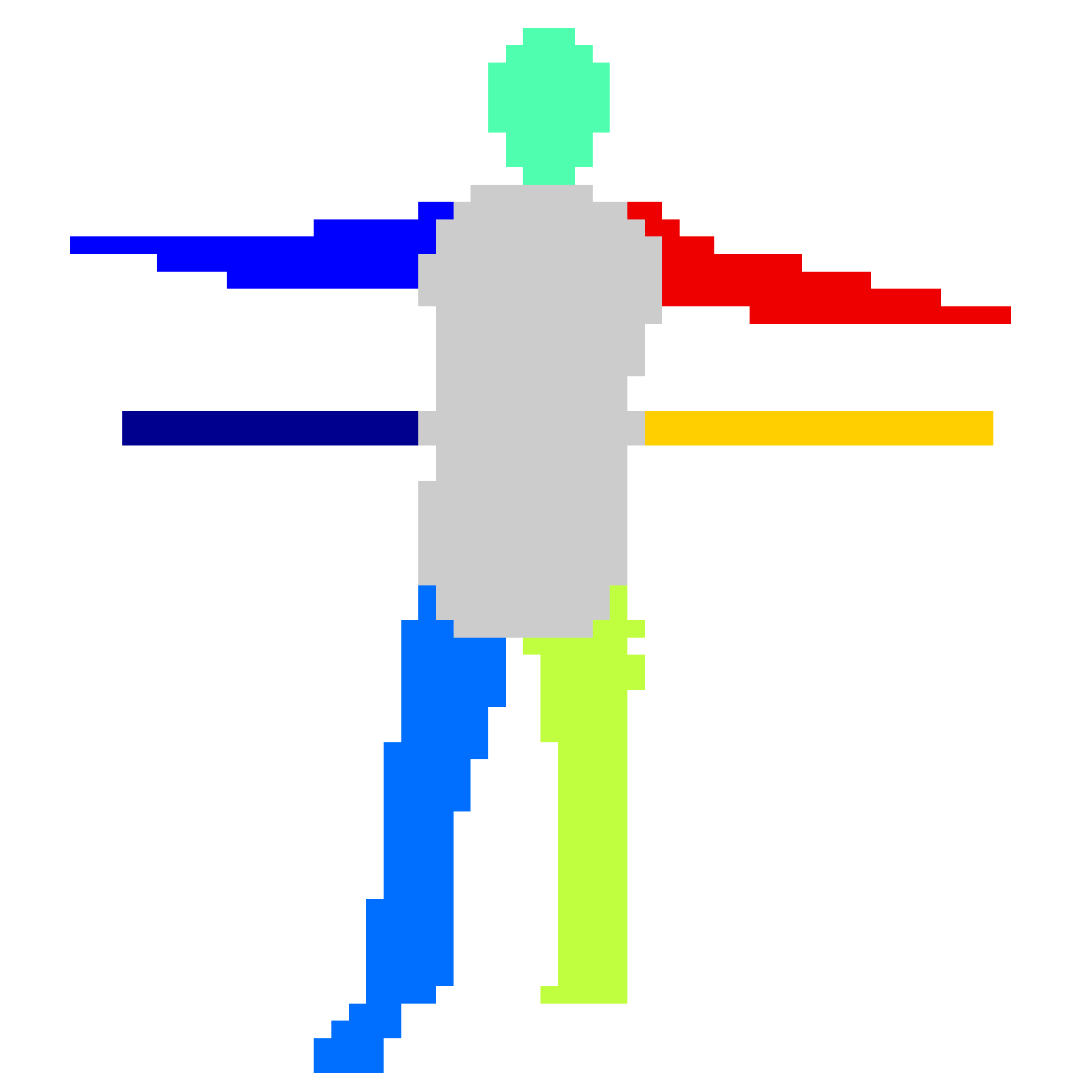}\\
\includegraphics[height=3.5cm]{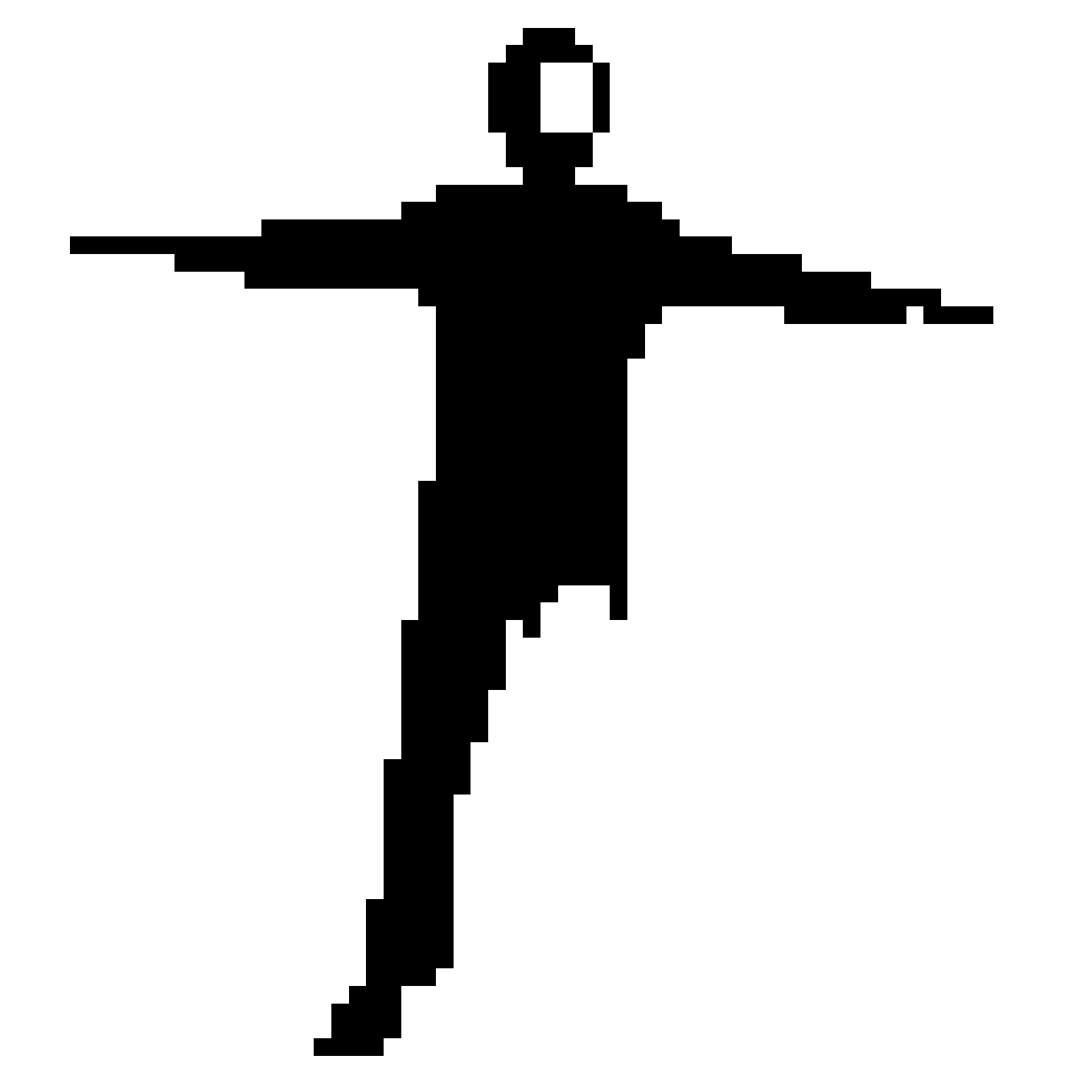}&
\includegraphics[height=3.5cm]{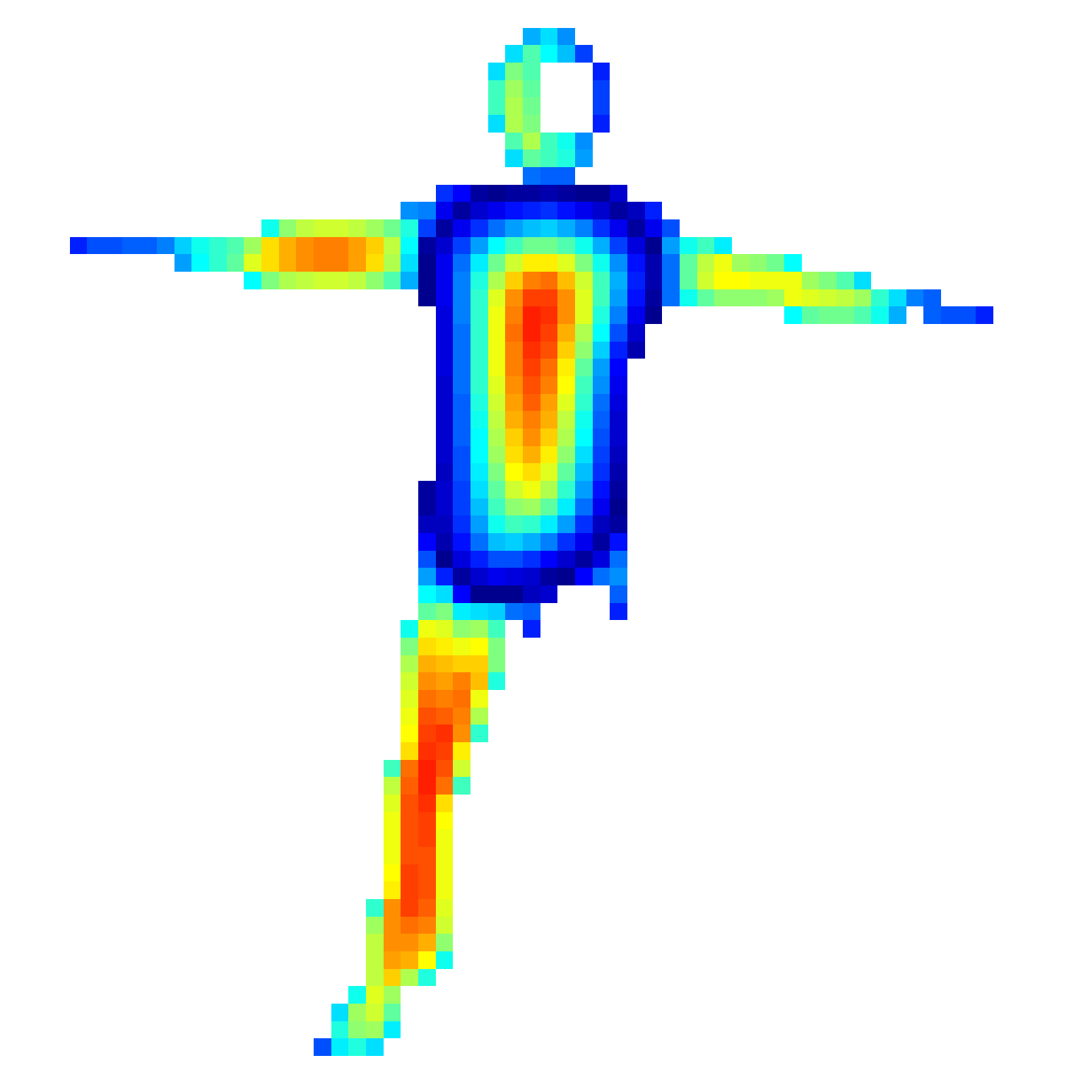}&
\includegraphics[height=3.5cm]{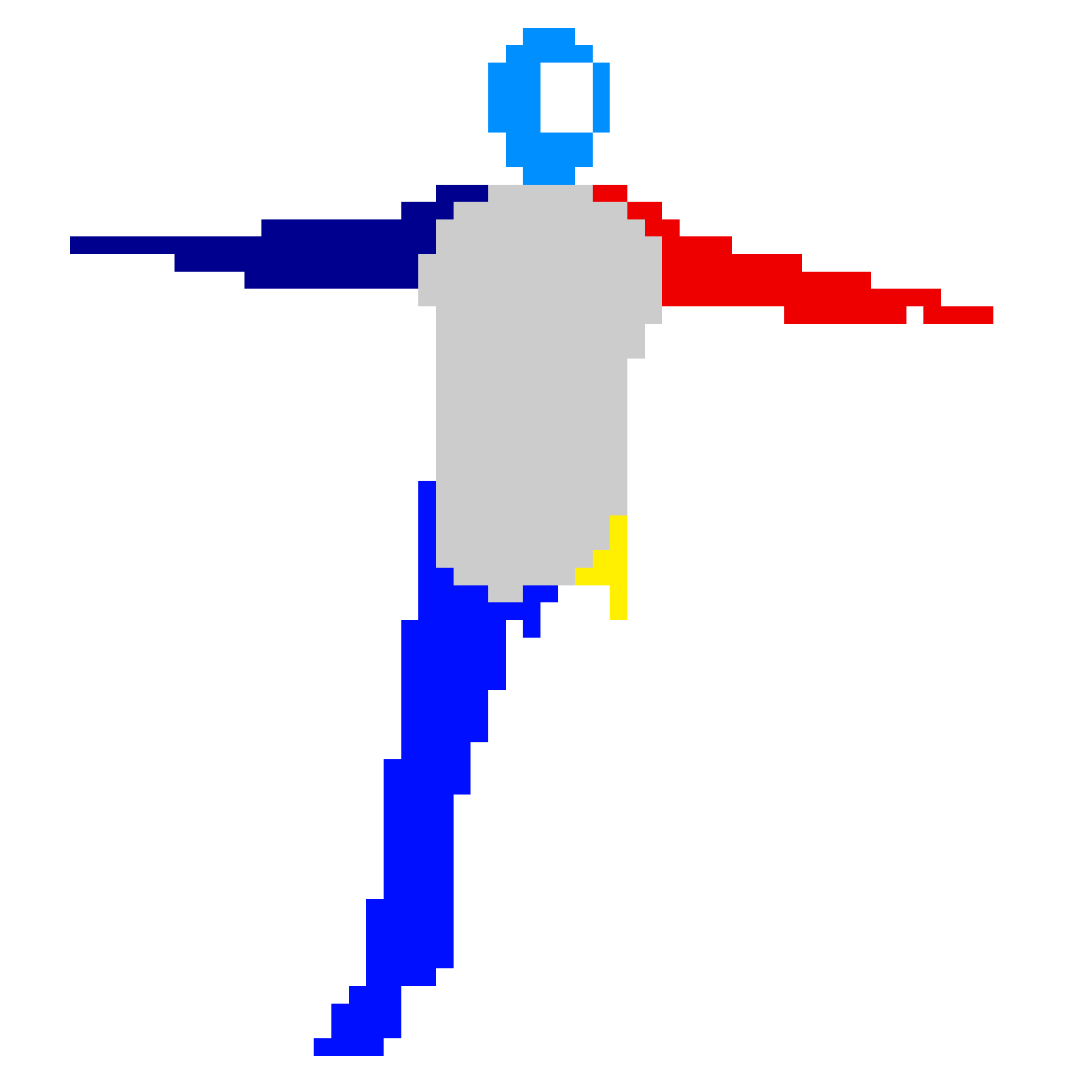}\\
\end{tabular}
\caption {The method is robust with respect to occlusion, missing
data, extra objects, and holes. (a) Input silhouettes. (b) Absolute
value of $w$. (c) Parts extracted by applying Matlab's {\sl
watershed} command to $w$.} \label{fig:occlusion}
\vglue 5pt
\end{figure}

\begin{figure}[!ht]
\centering
\begin{tabular}{cc}
\includegraphics[height=6.5cm]{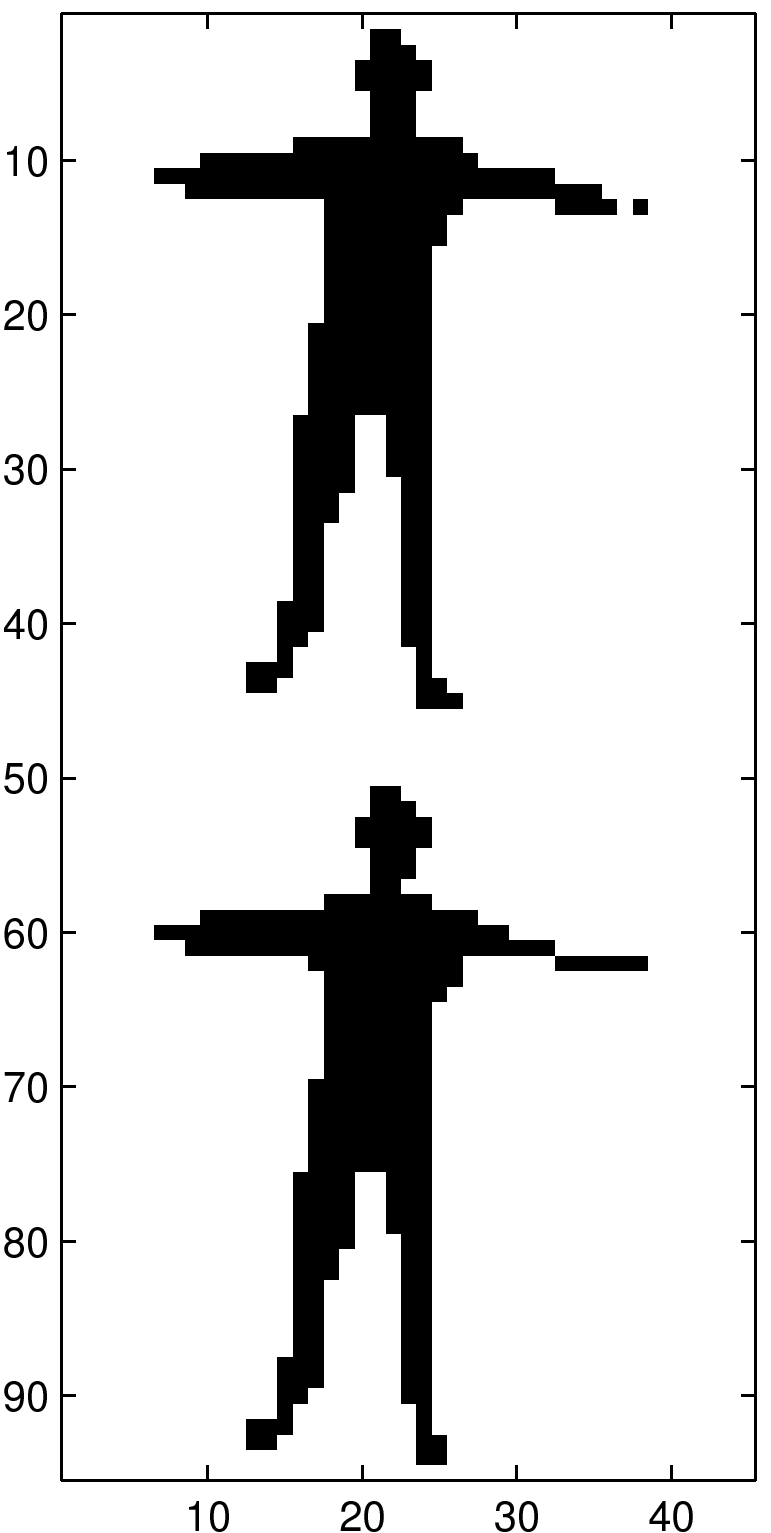}
\includegraphics[height=6.5cm]{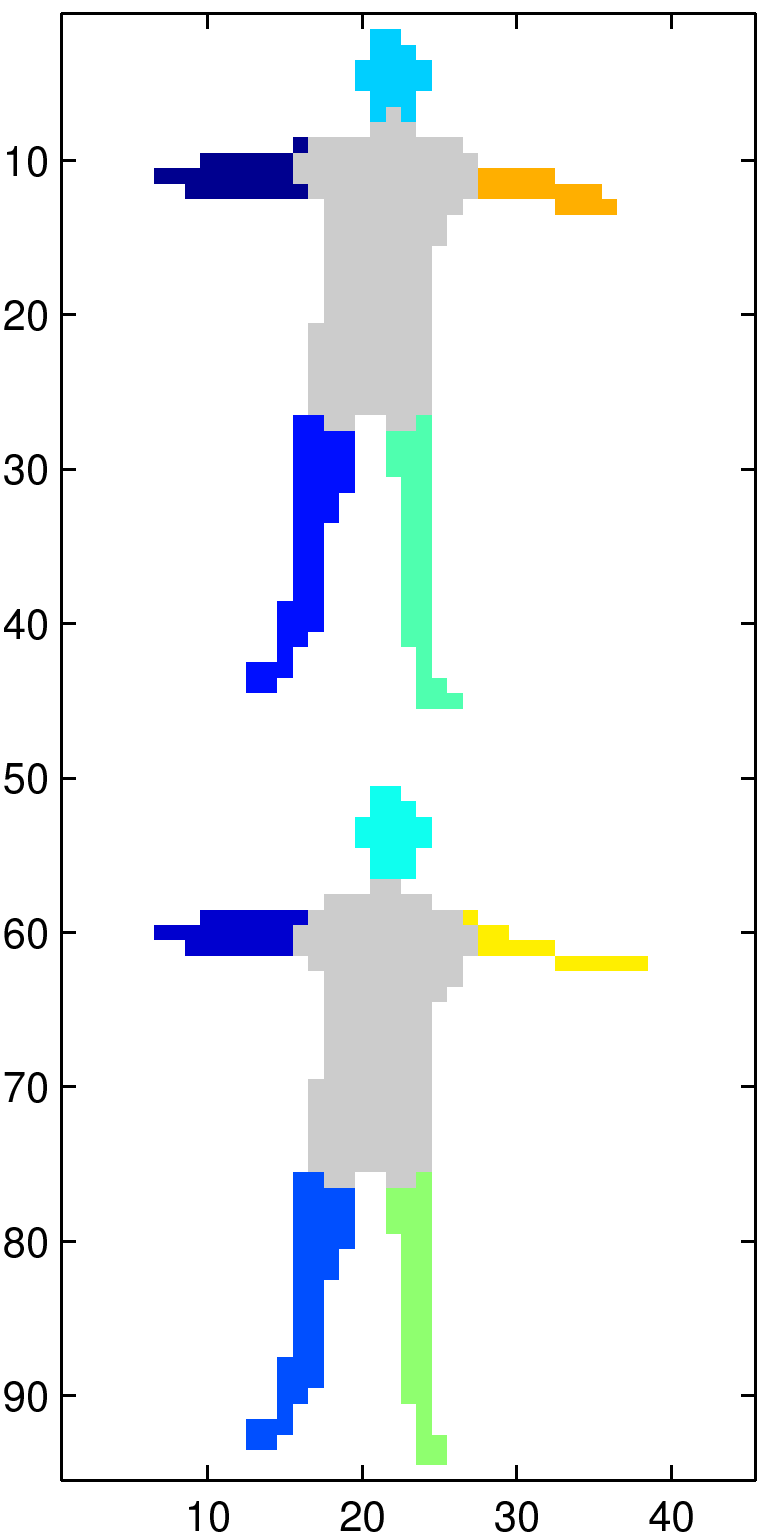}
\end{tabular}
\caption{A scene with two silhouettes. The method is applicable to
disconnected sets.} \label{fig:multiple}

\vglue 20pt

\centering
{\footnotesize\begin{tabular}{cc}
\includegraphics[height=4.0cm]{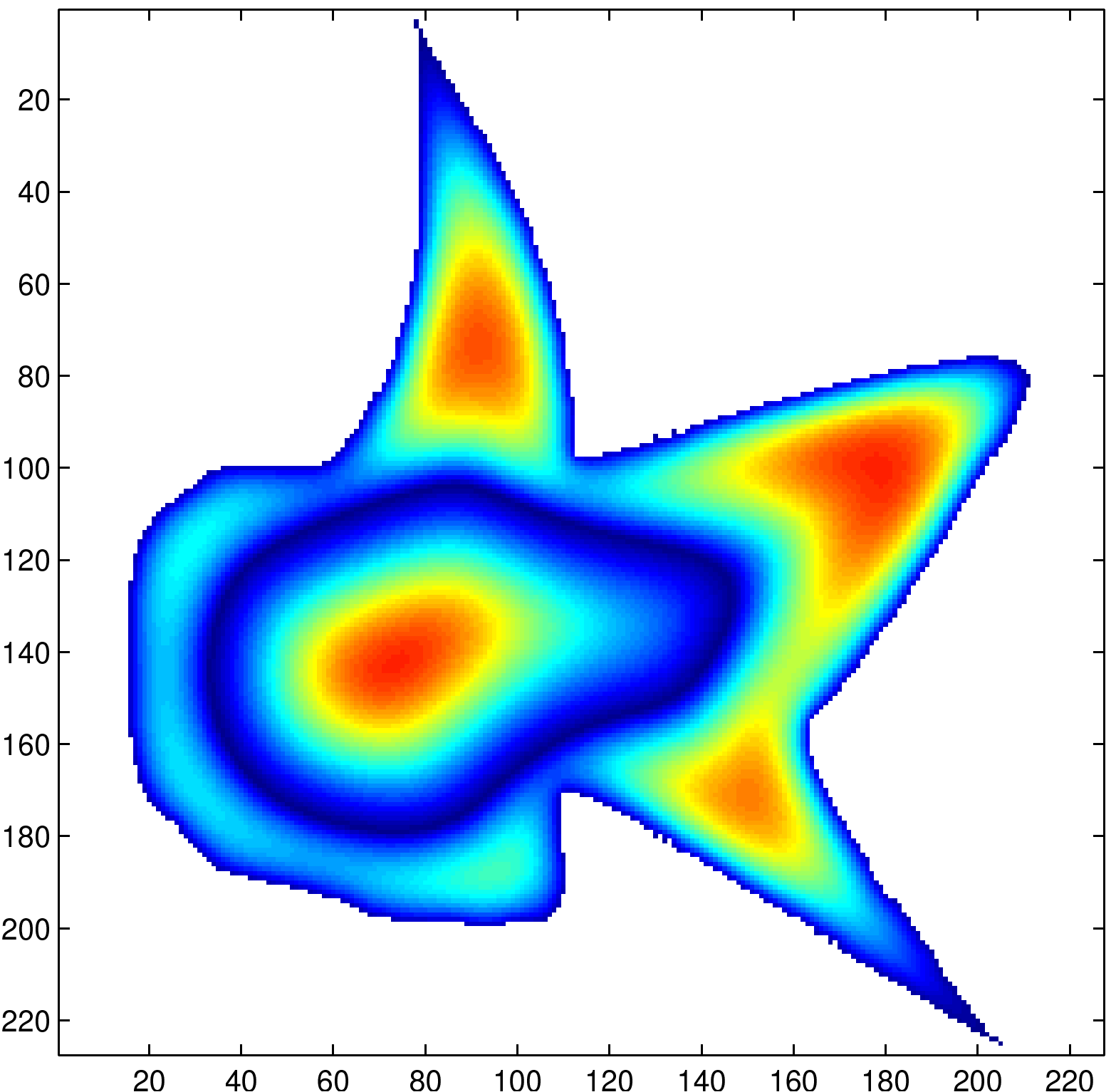}&
\includegraphics[height=4.0cm]{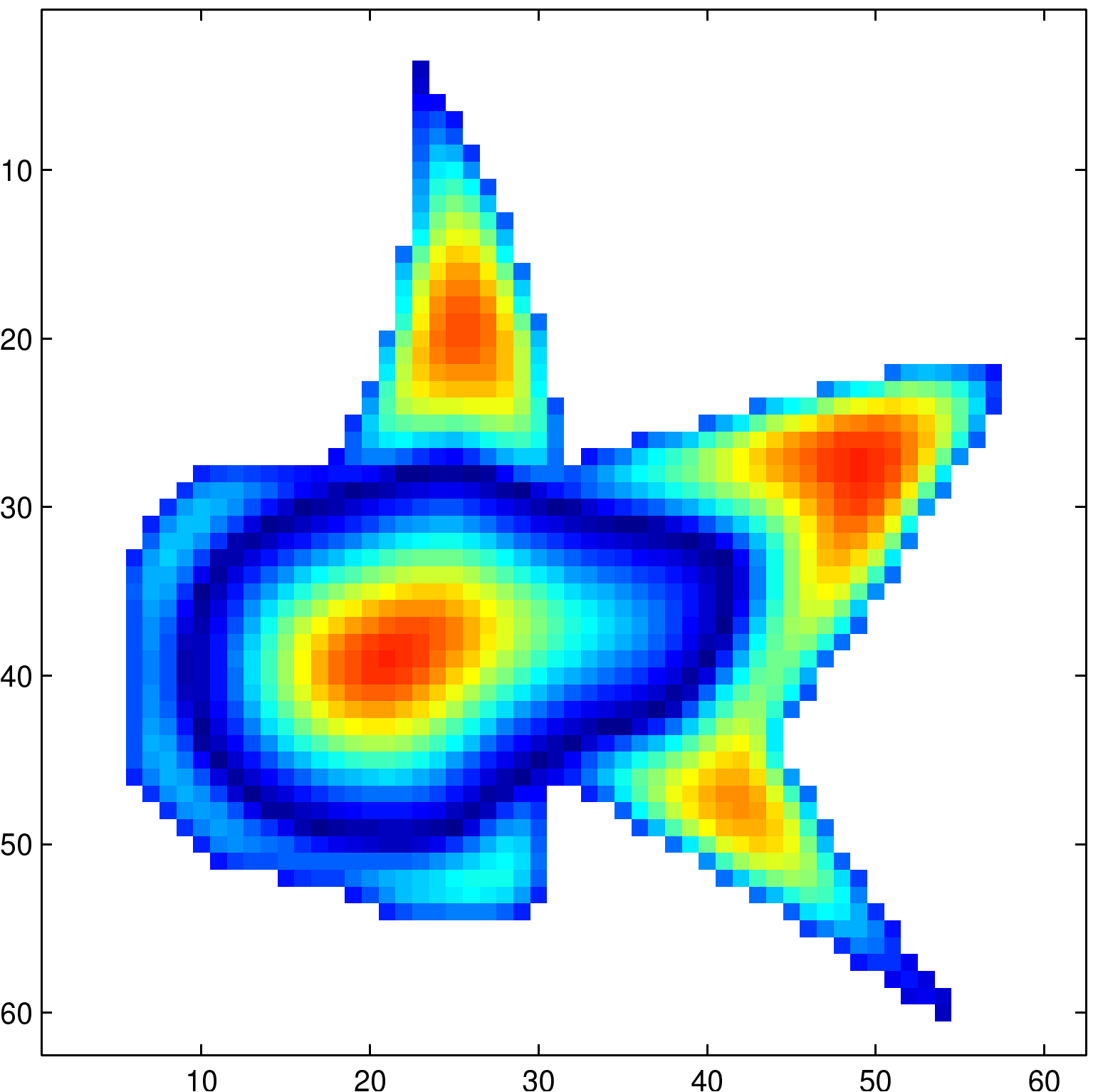}\\[4pt]
\includegraphics[height=4.0cm]{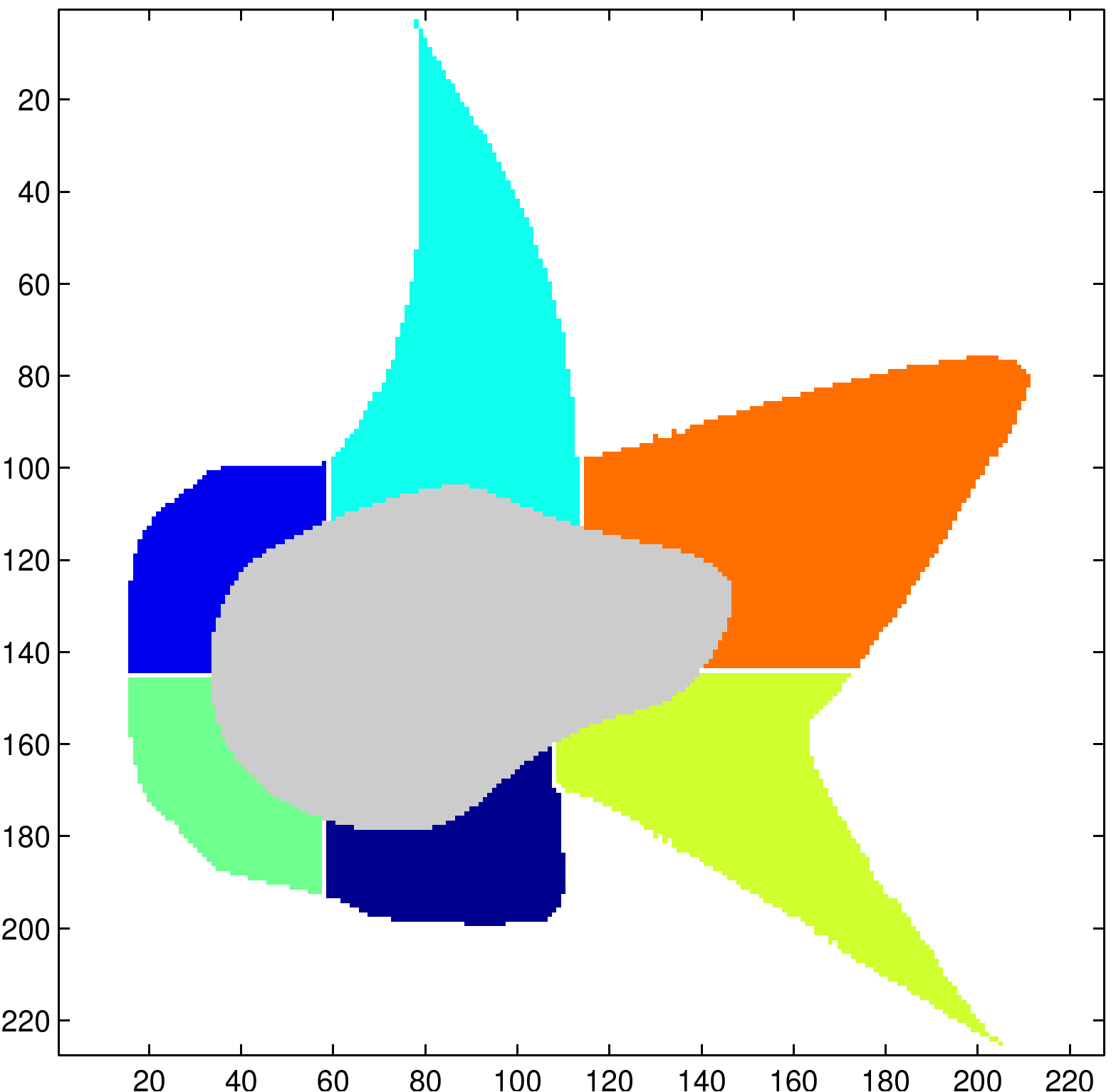} &
\includegraphics[height=4.0cm]{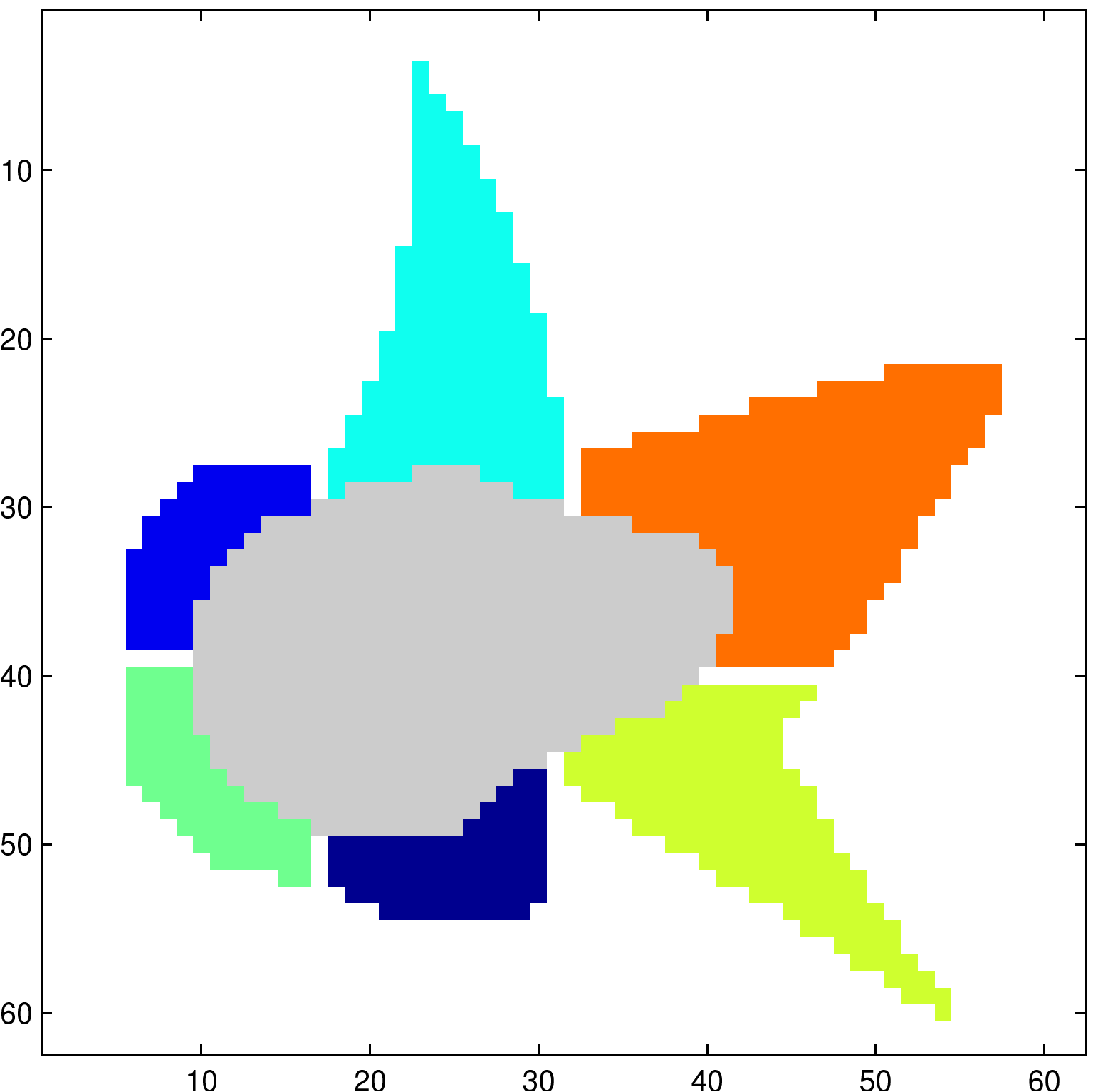}\\[2pt]
(a) & (b)
\end{tabular}}
\caption {The method is robust with respect to resolution changes.
An artificial shape~\cite{Aslan05} on (a)~$220 \times 220$,  (b) $60
\times 60$ lattices.} \label{fig:resolution}
\end{figure}

\begin{figure}[!b]
\centering
\begin{tabular}{cccc}
\includegraphics[height=2.4cm]{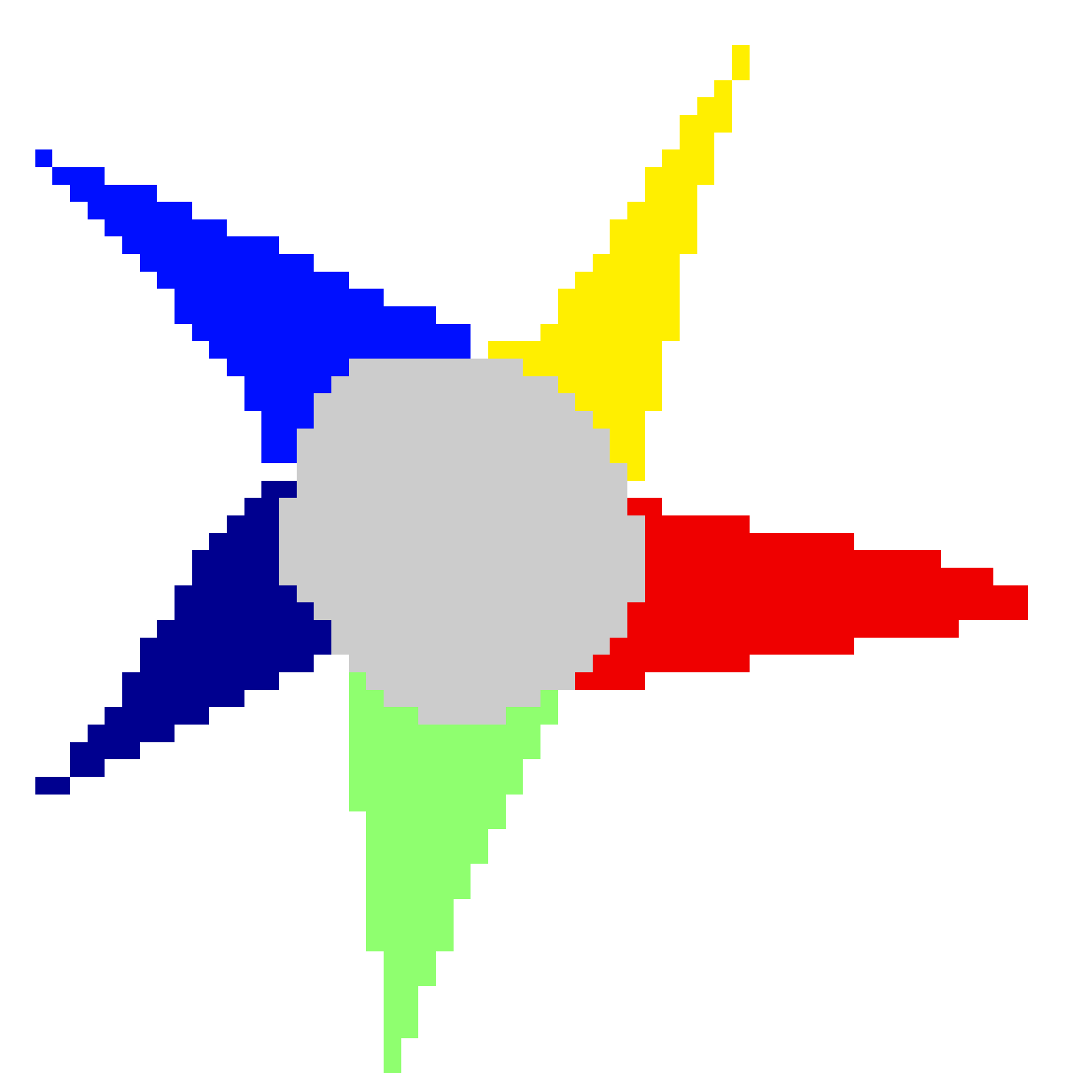} &
\includegraphics[height=2.4cm]{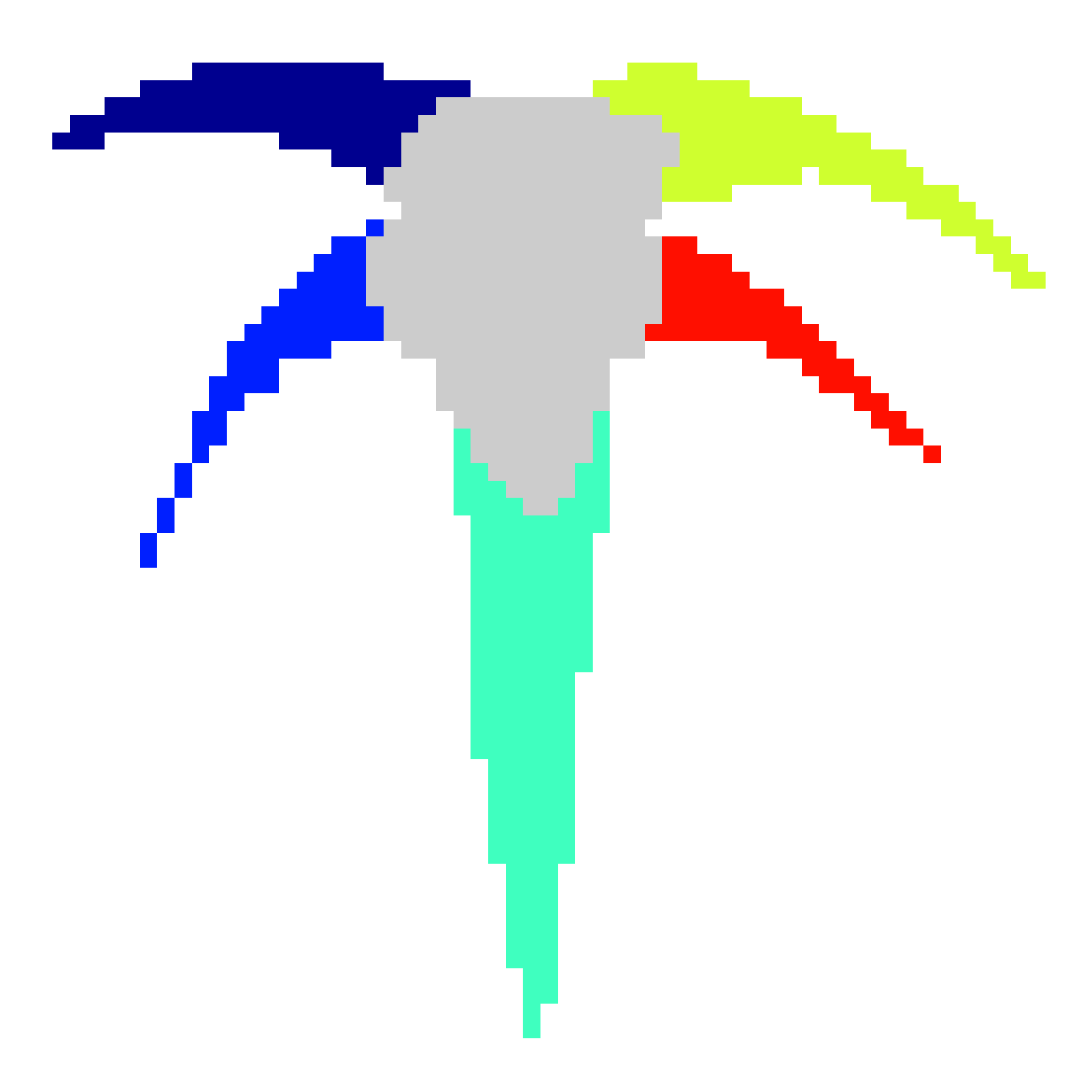}&
\includegraphics[height=2.4cm]{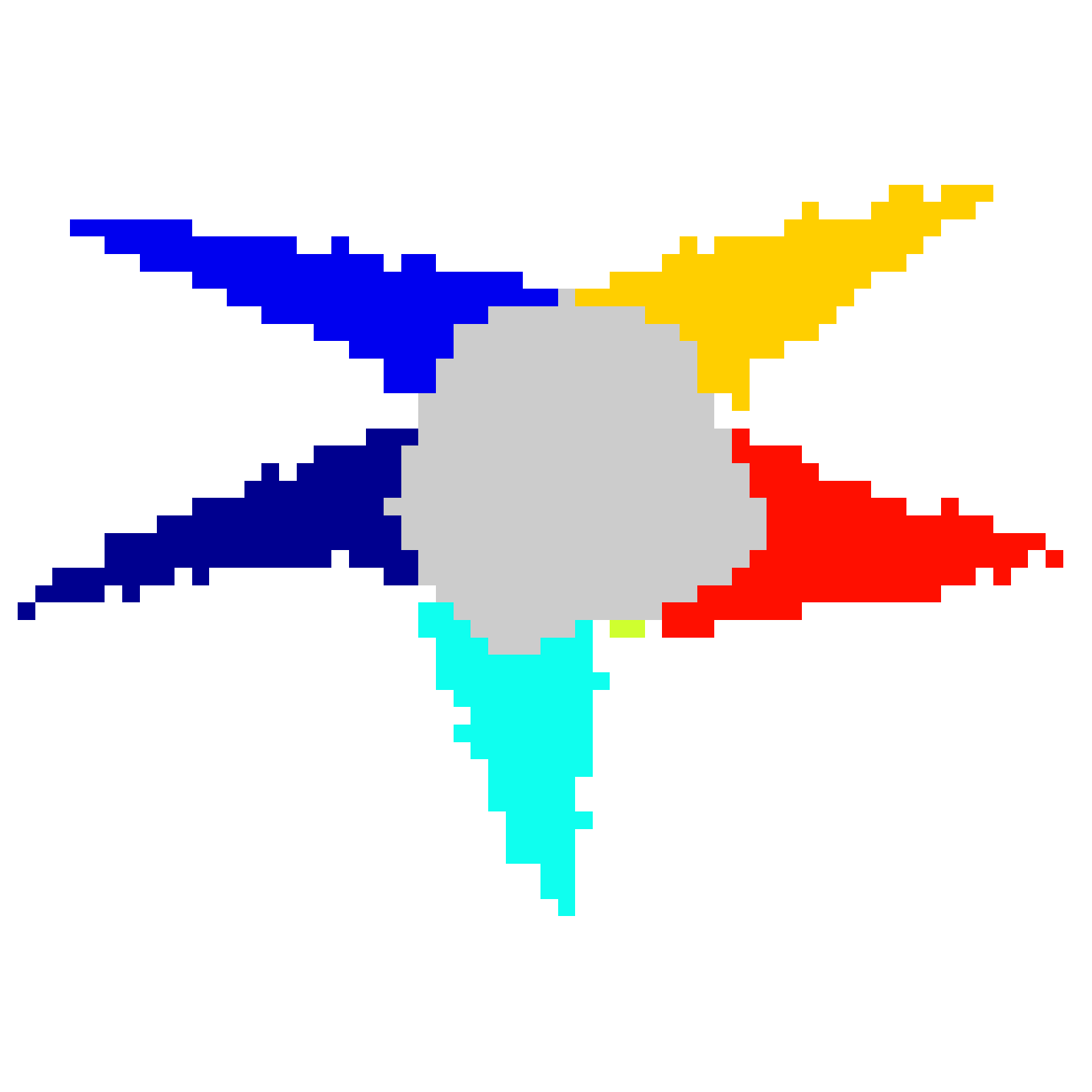}&
\includegraphics[height=2.4cm]{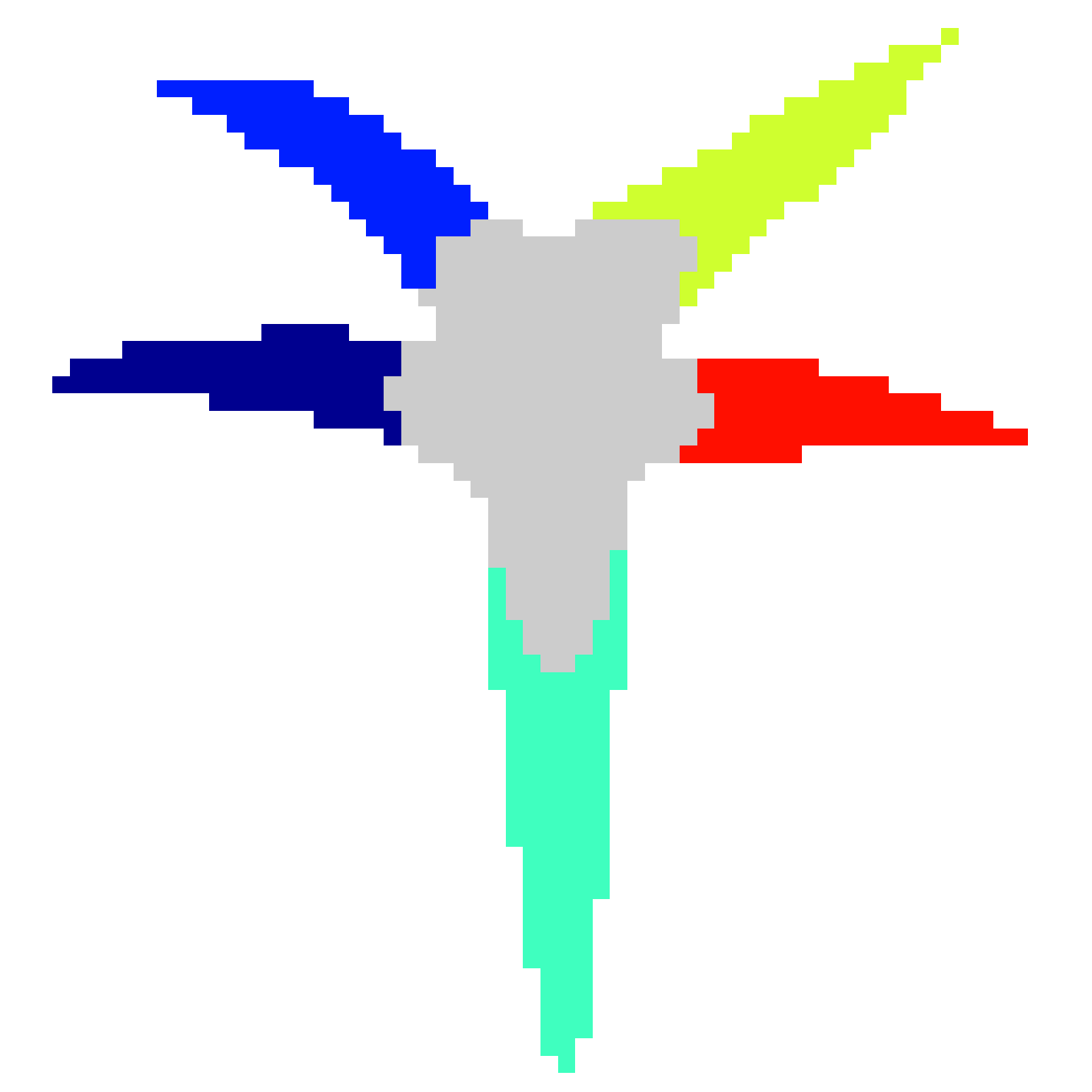}
\\
\includegraphics[height=2.4cm]{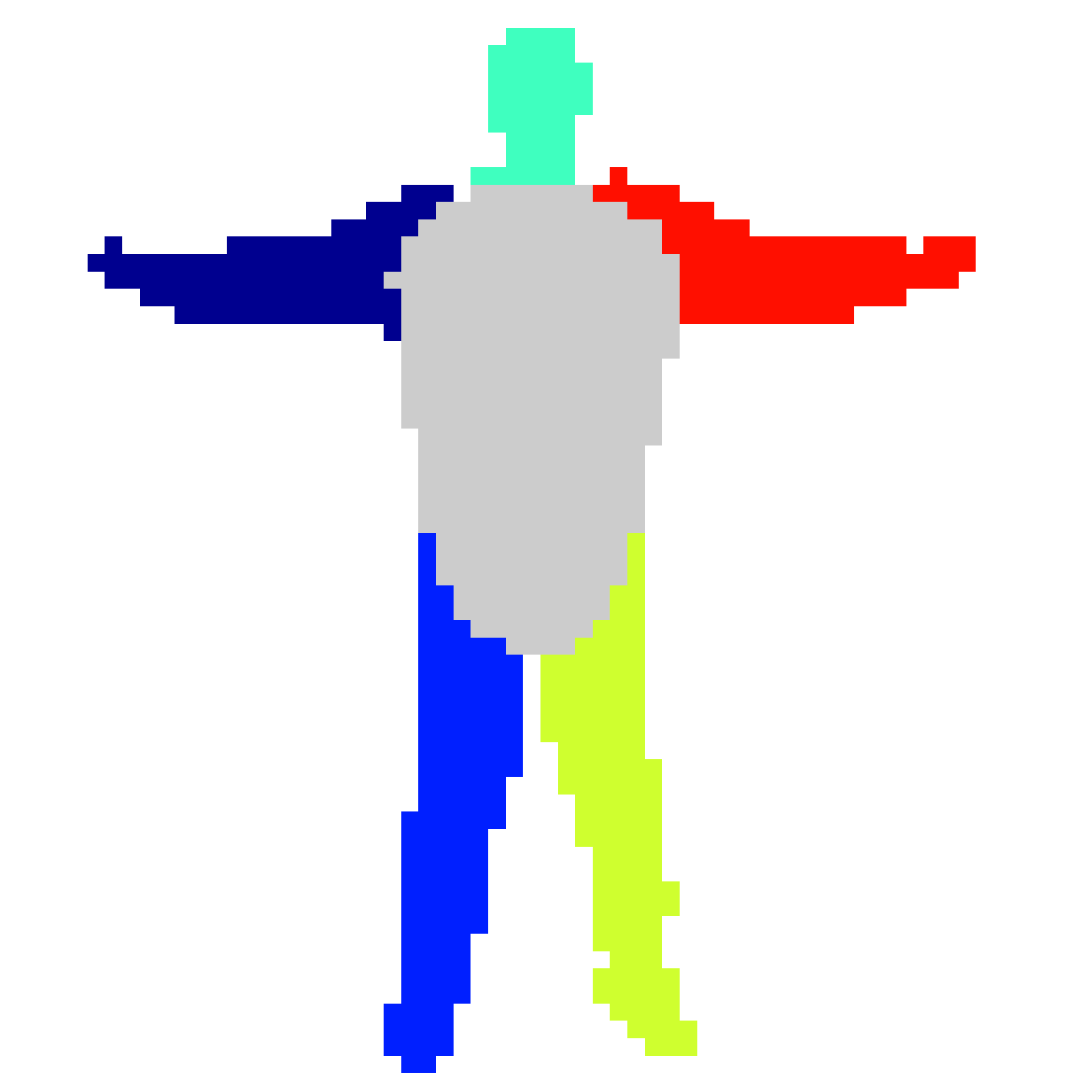}&
\includegraphics[height=2.4cm]{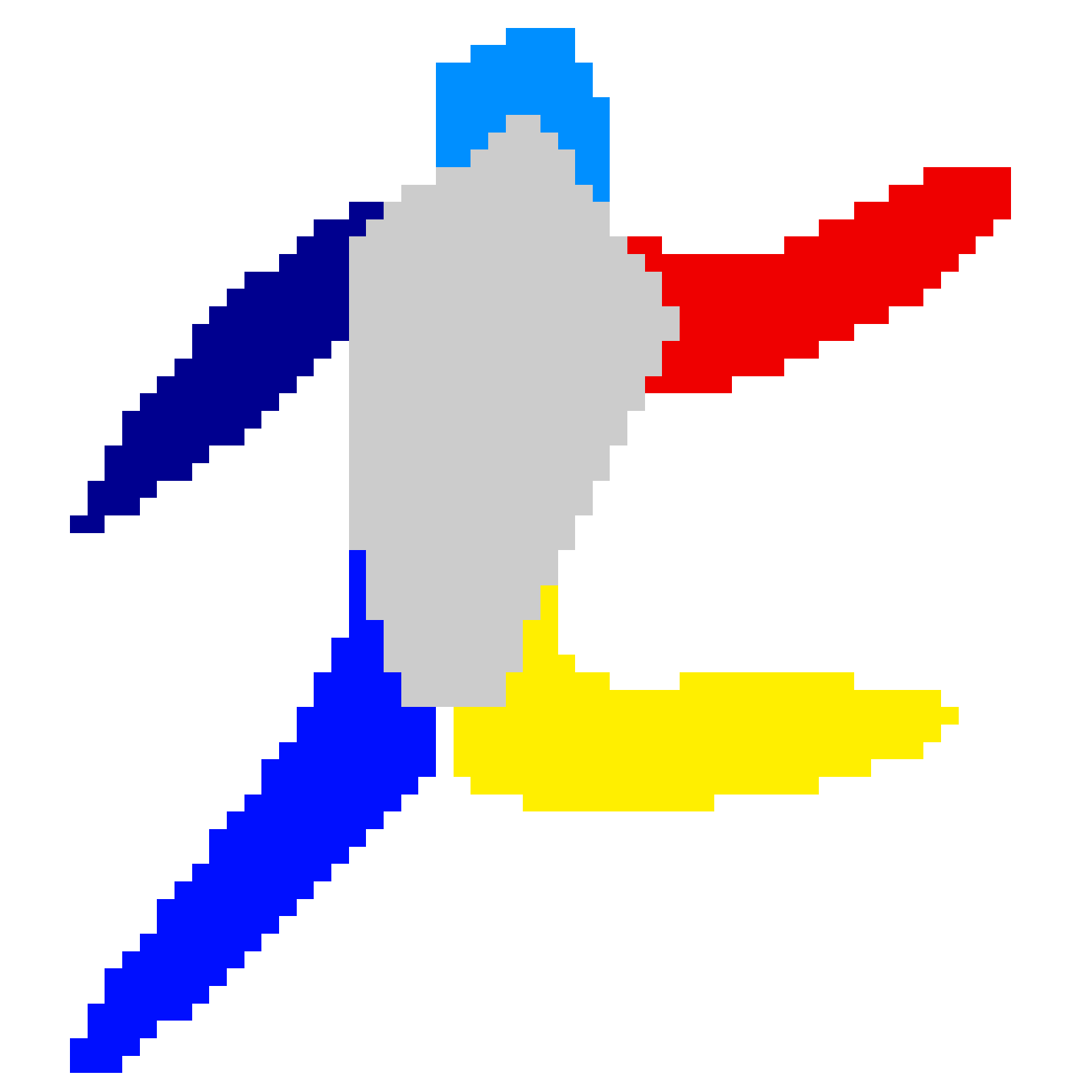}&
\includegraphics[height=2.4cm]{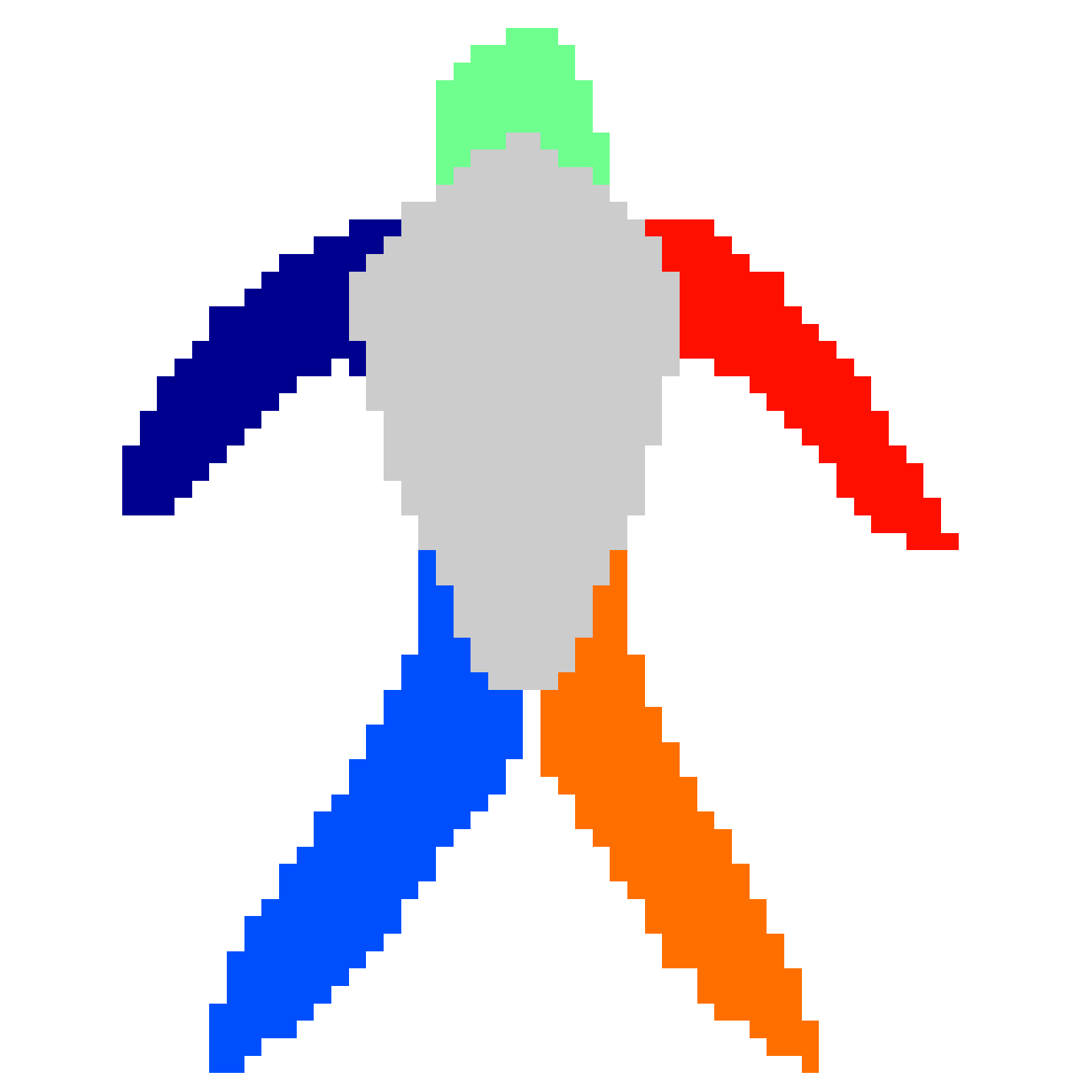}&
\includegraphics[height=2.4cm]{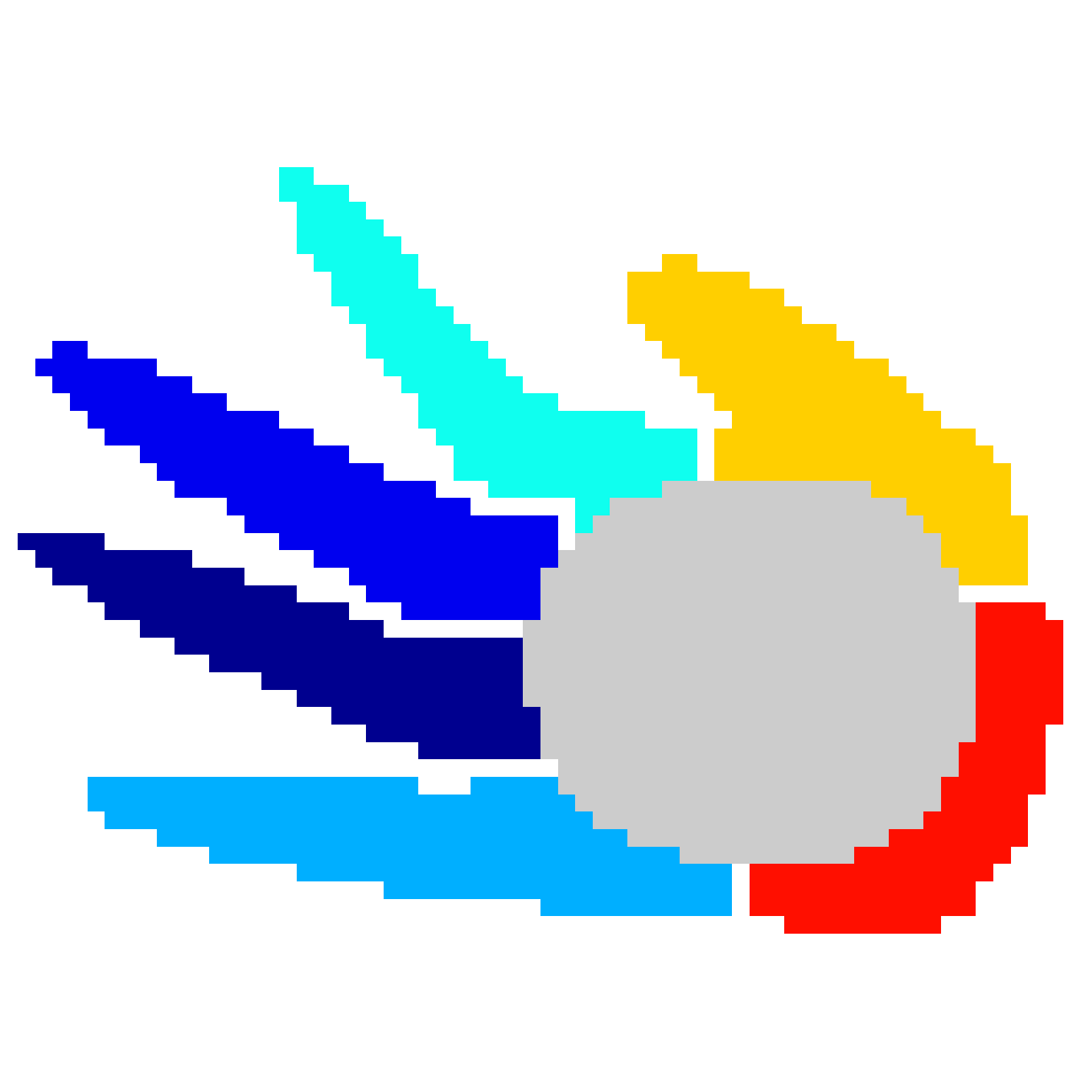}\\
\includegraphics[height=2.4cm]{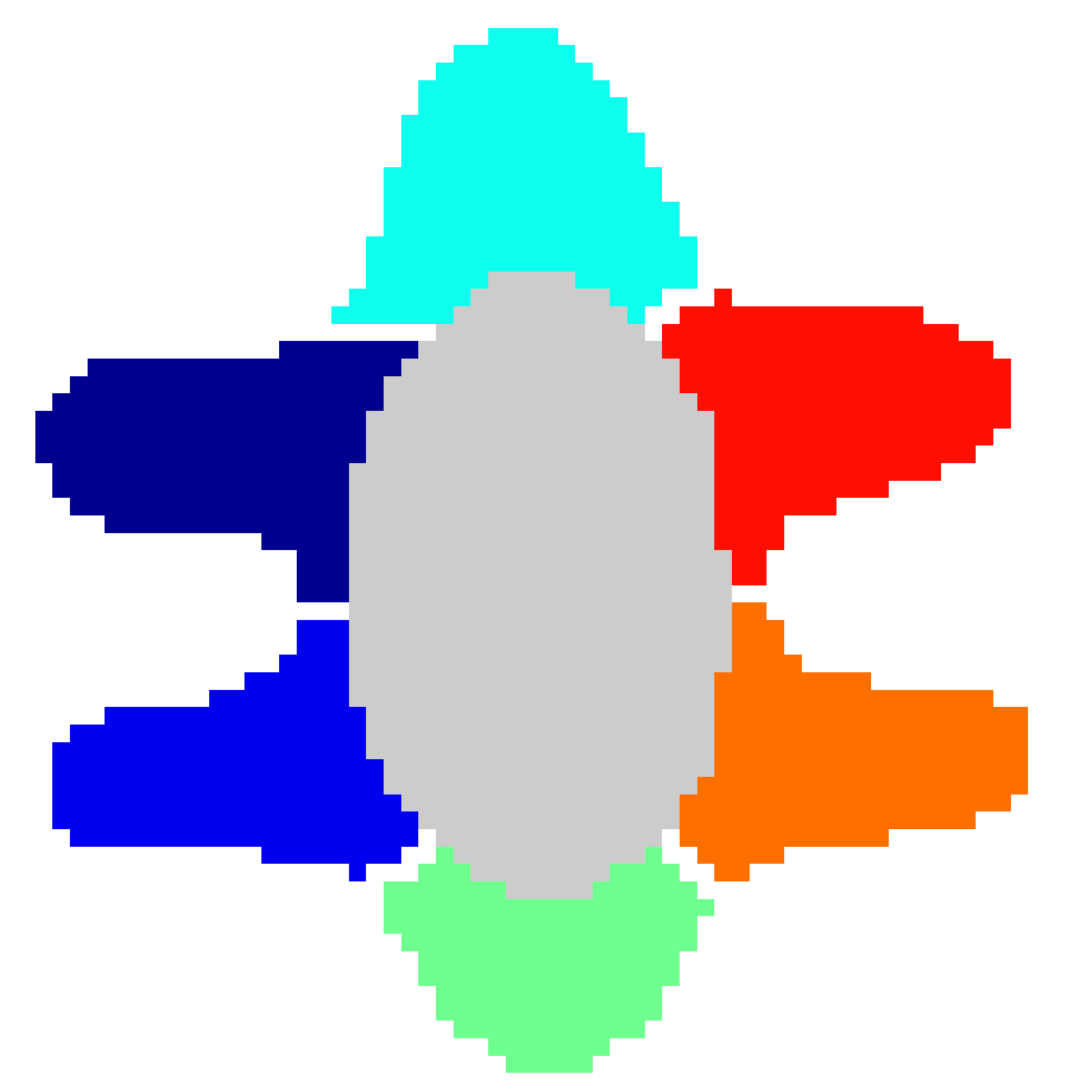}&
\includegraphics[height=2.4cm]{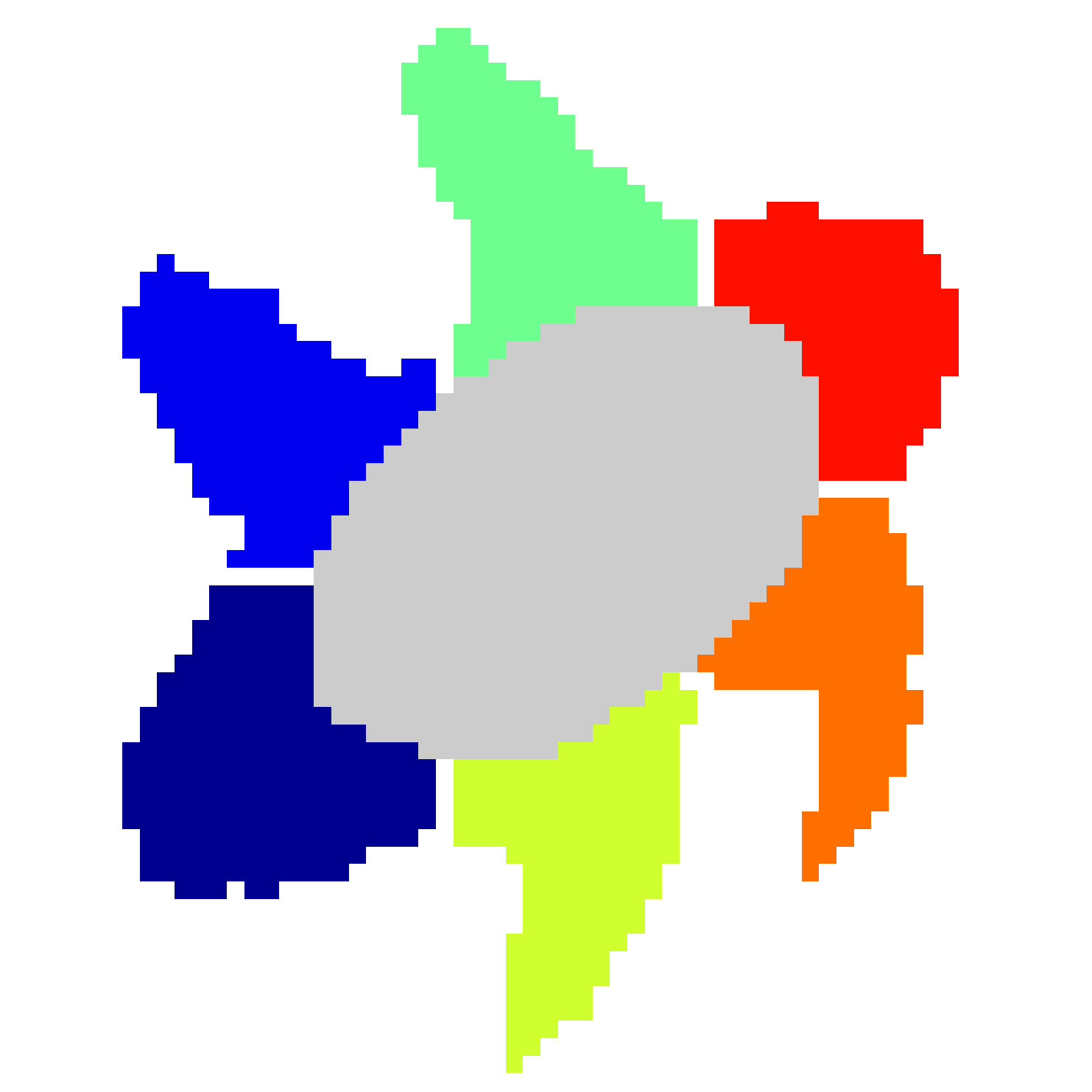}&
\includegraphics[height=2.4cm]{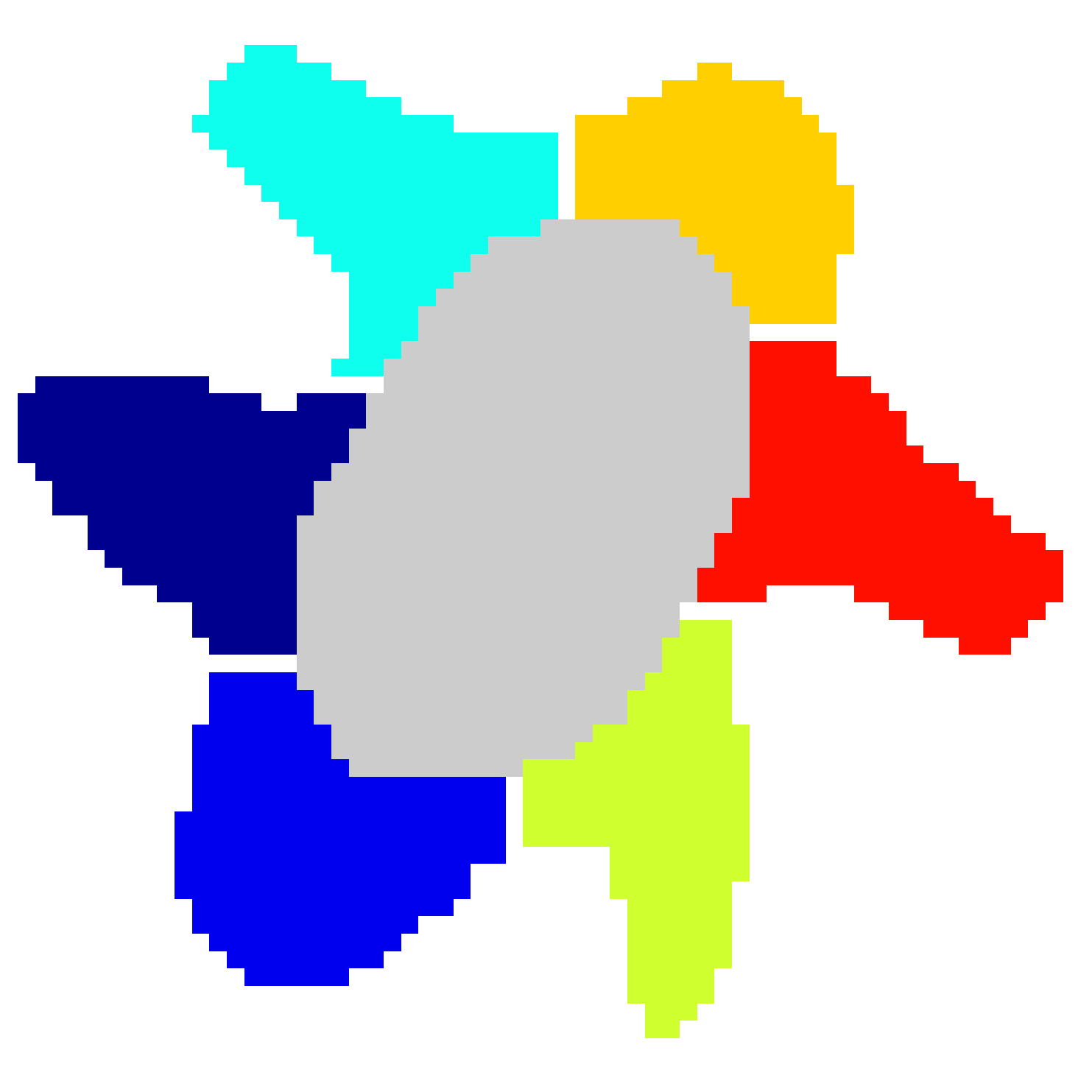}&
\includegraphics[height=2.4cm]{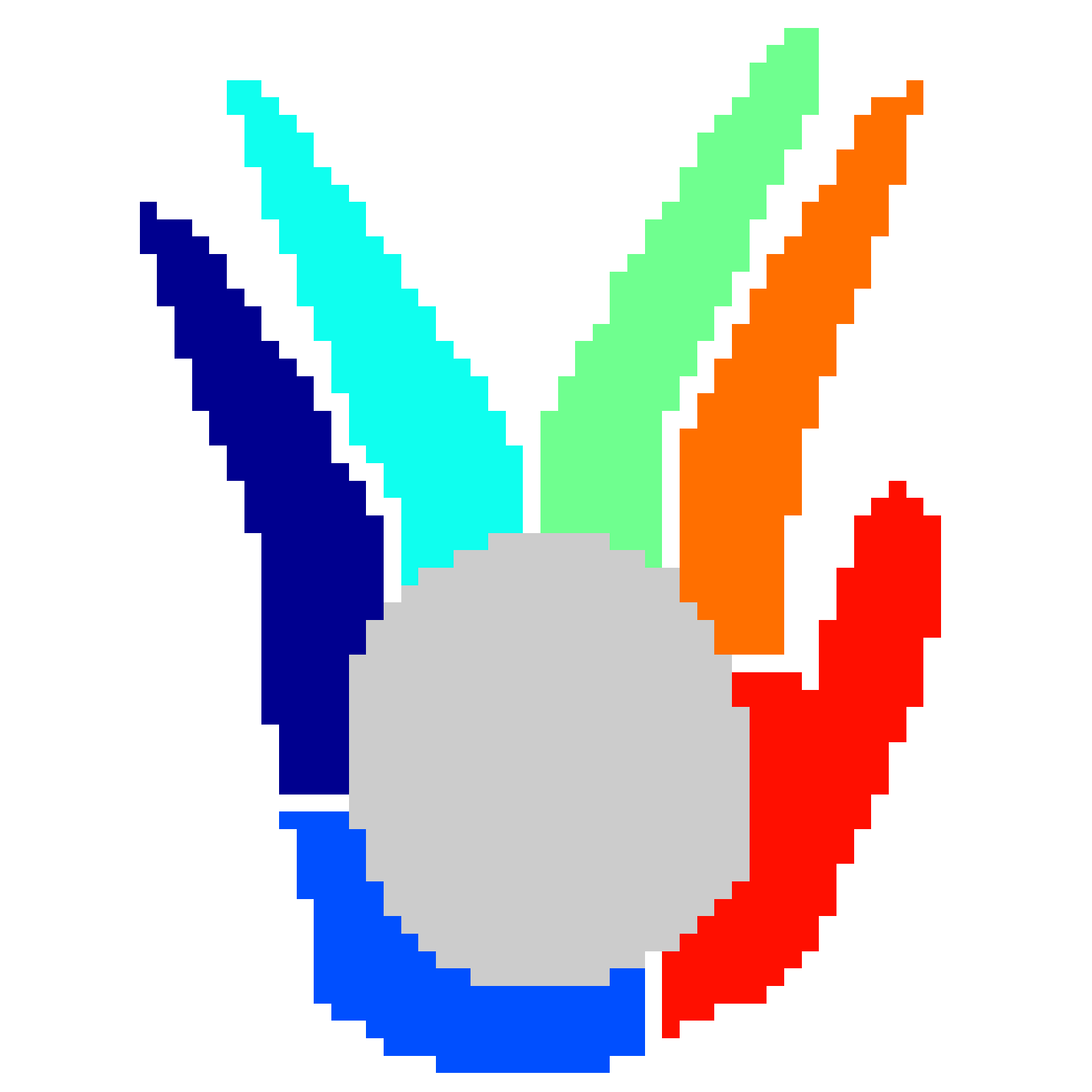}\\
\includegraphics[height=2.4cm]{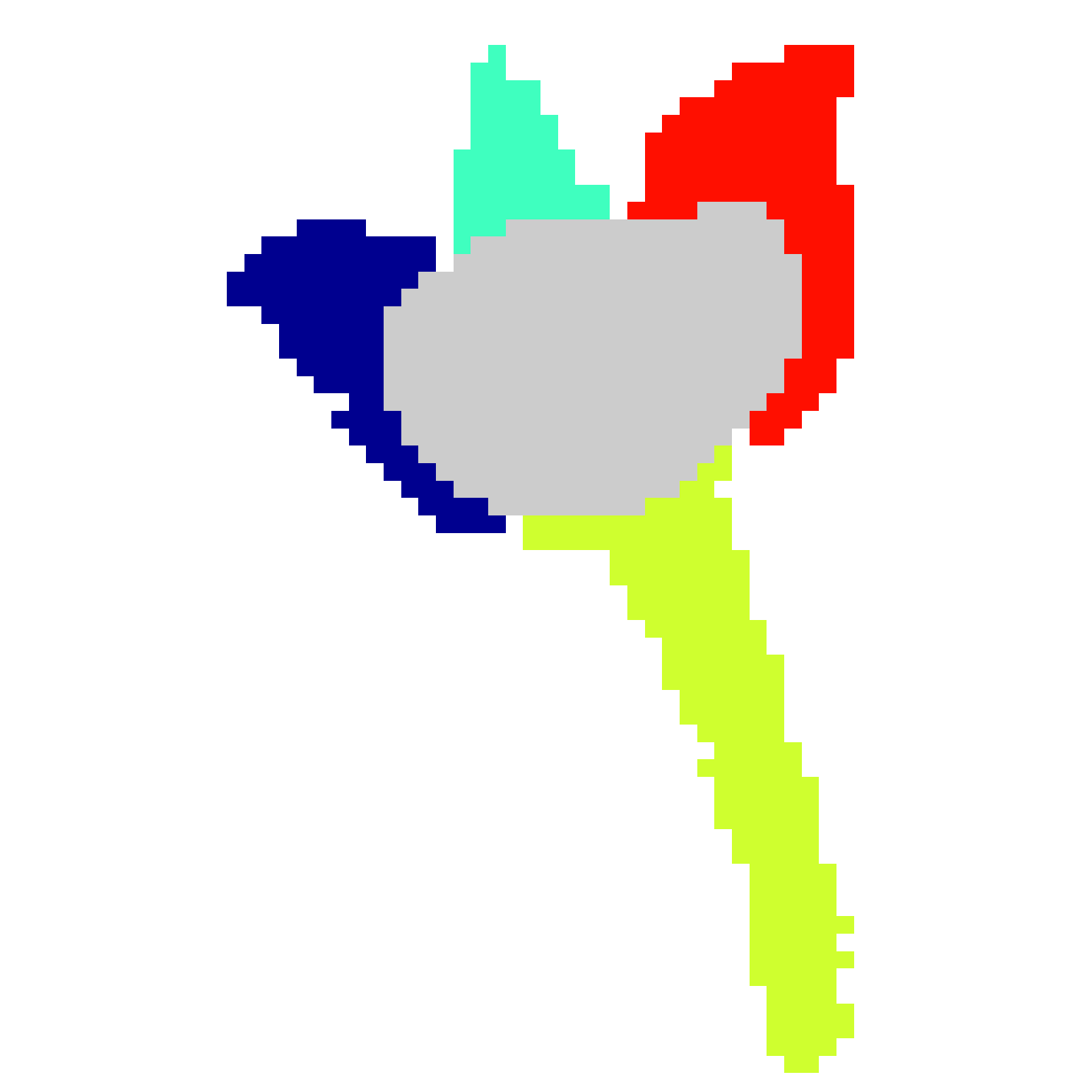}&
\includegraphics[height=2.4cm]{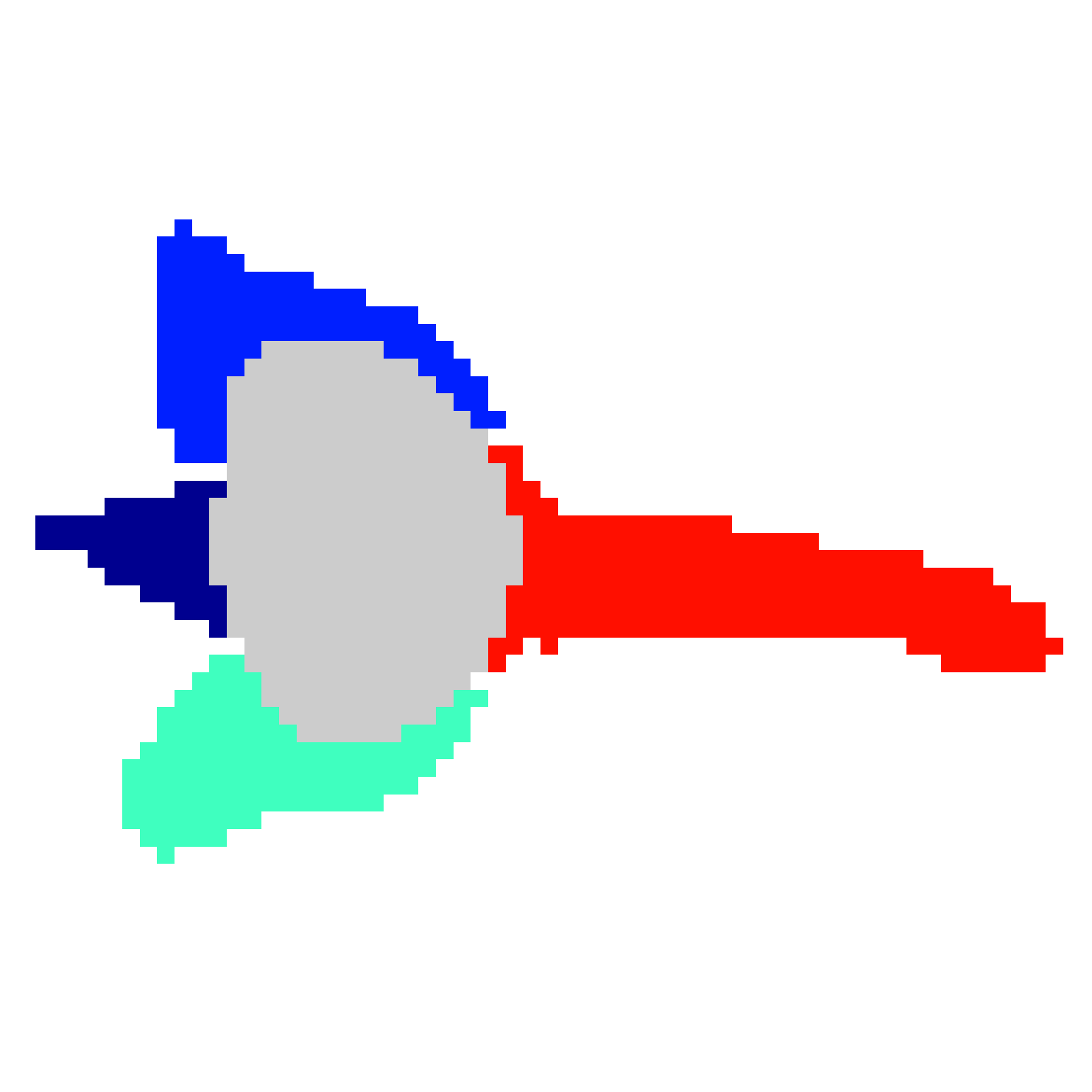}&
\includegraphics[height=2.4cm]{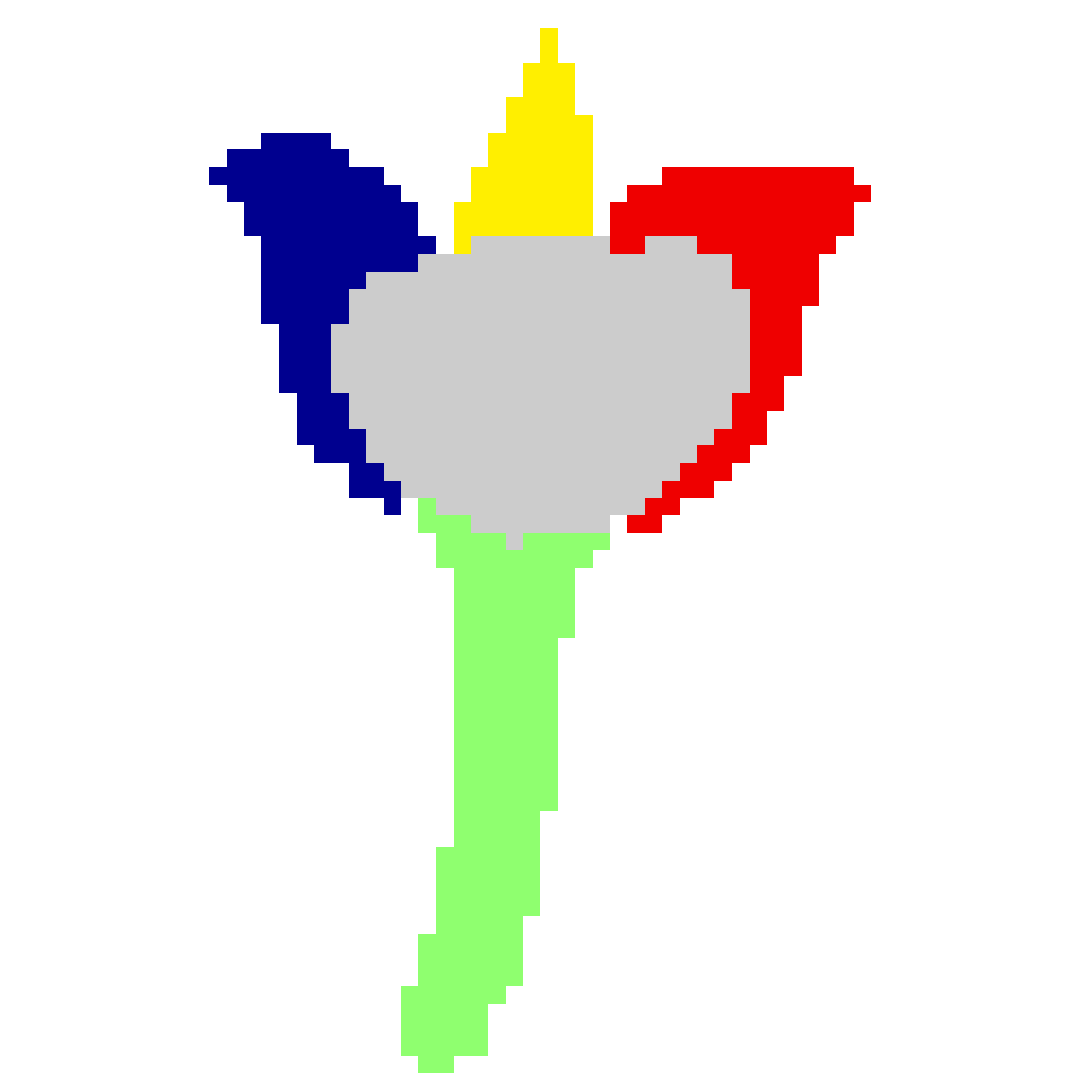}&
\includegraphics[height=2.4cm]{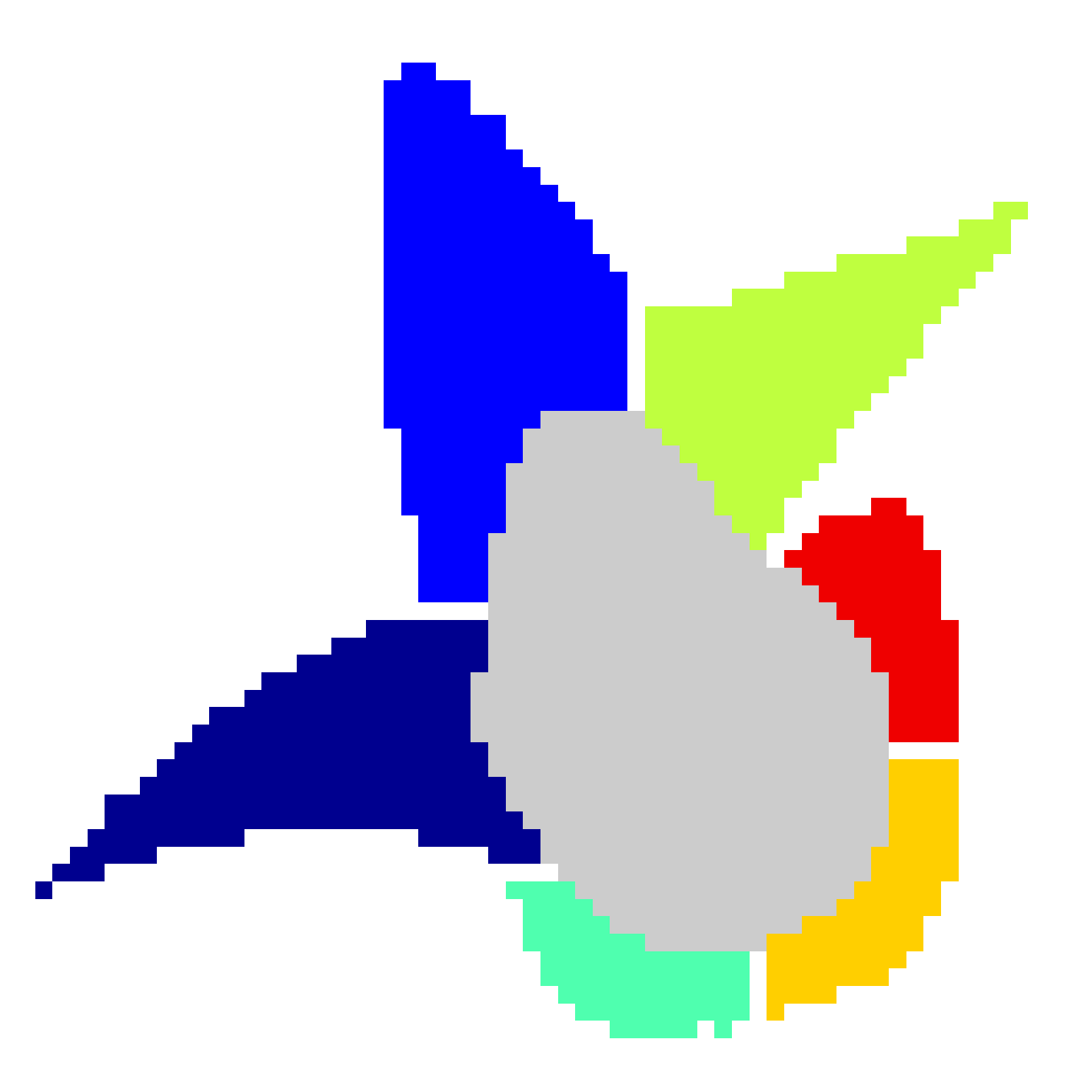}
\end{tabular}
\caption {Sample decompositions. Similar
shapes are partitioned similarly; and the parts are
compatible with intuition. } \label{fig:sample}
\end{figure}

\index{watershed}
In Figs.~(\ref{fig:occlusion}-\ref{fig:sample}),  some decomposition
results are provided. These results are obtained by applying
Matlab's {\sl watershed} command to $\Omega^-$. (This is equivalent
to considering a growth starting from each local minima of
$\omega$.)

In Fig.~\ref{fig:occlusion}, the first column depicts the given
shape. The second column depicts the normalized absolute value of
$\omega$. The third column depicts the parts. Bright colors are used
for parts on which $\omega$ is negative; and gray is used for the
part on which $\omega$ is positive.

\index{shape!parts!semantically meaningful}
\index{shape!with holes}
\index{shape!occlusion}
The silhouette shown in the first row is a sampled down version  of a
human silhouette~\cite{Gorelick06} from
its  original resolution of $414 \times 459$ to $60 \times 60$.  The
silhouettes on the remaining rows are drawn by the author to
introduce holes, missing portions, occlusions. In each case, semantically meaningful parts (the torso,
the legs, the arms, the head) are captured. Even though the
formulation includes a global term, the local changes, no matter how
significant they are, do not affect the detected parts.

In Fig.~\ref{fig:multiple}, the applicability of the method when the
input consists of disconnected sets (multiple objects in a scene) is
demonstrated.

\vfill\pagebreak

\vfill\pagebreak

\begin{figure}[ht]
\centering
\begin{tabular}{ccc}
\includegraphics[height=3.2cm]{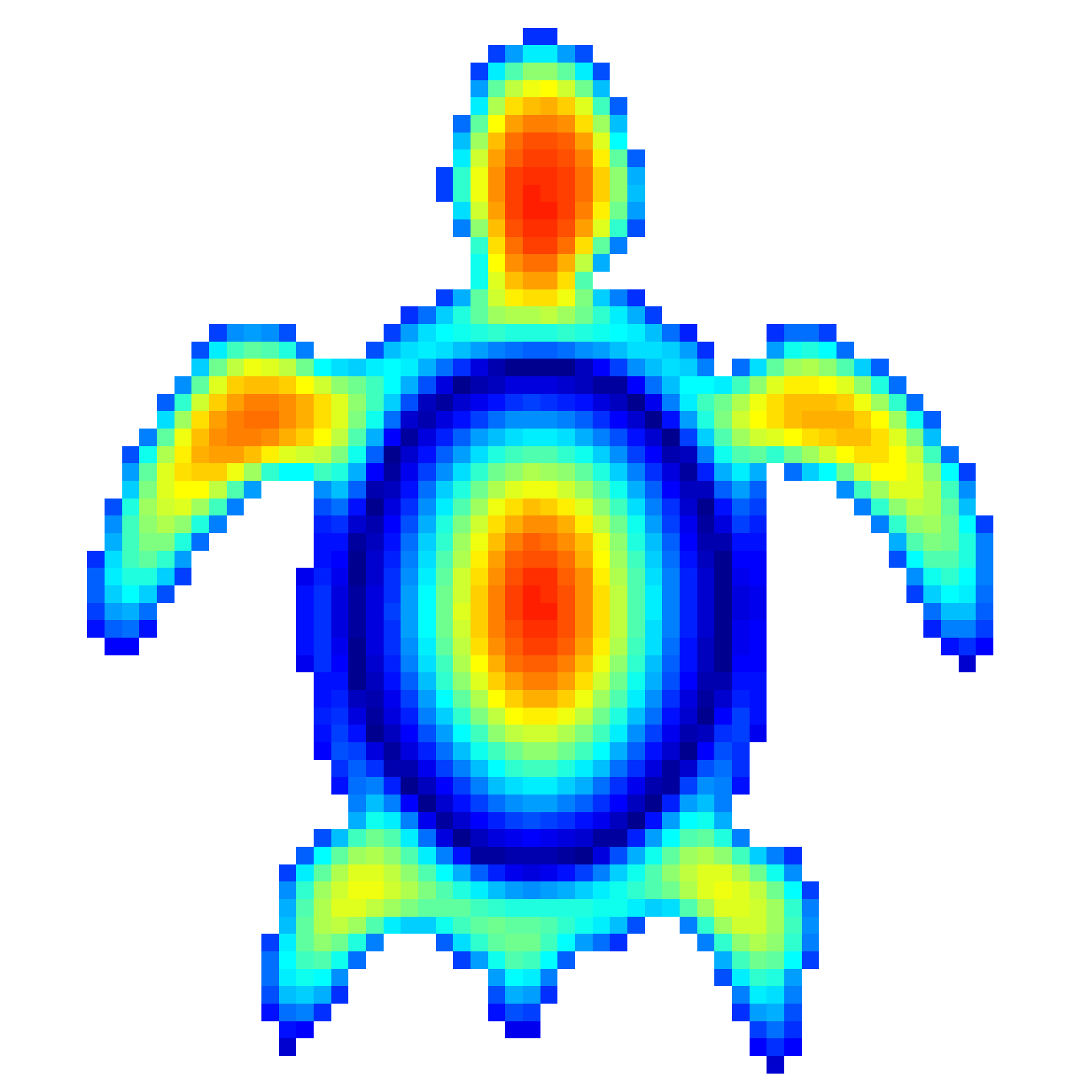}&
\includegraphics[height=3.2cm]{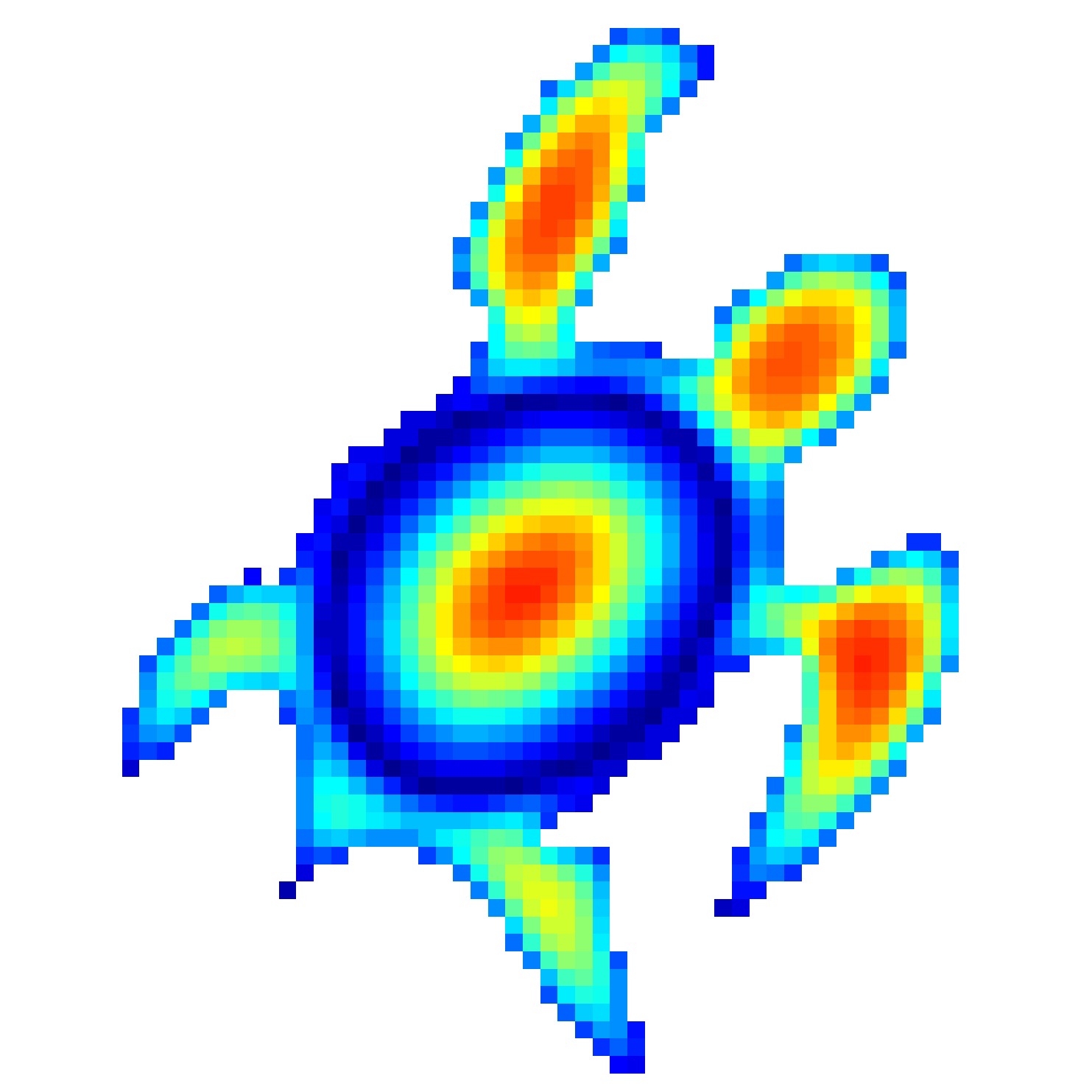}&
\includegraphics[height=3.2cm]{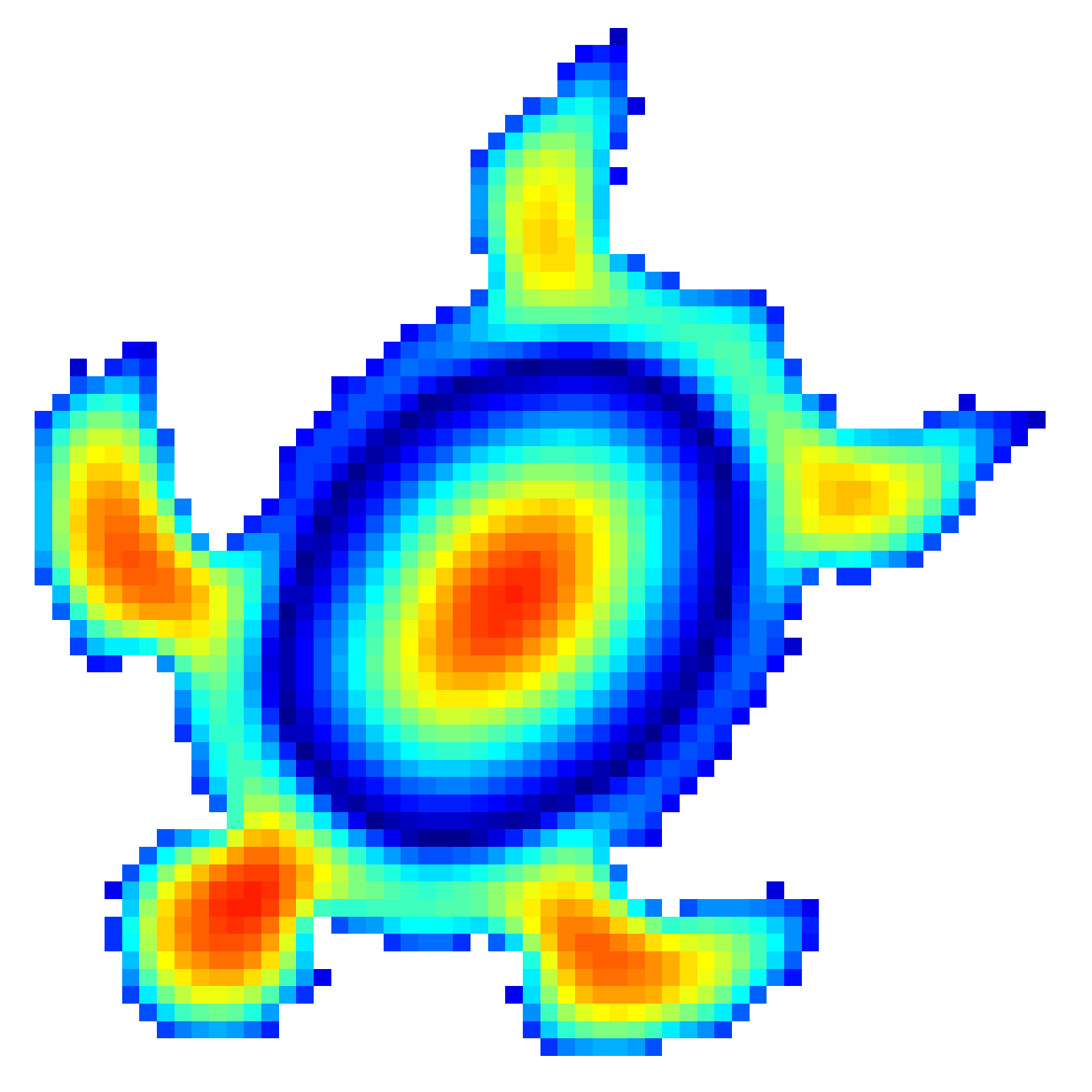}
\\
\includegraphics[height=3.2cm]{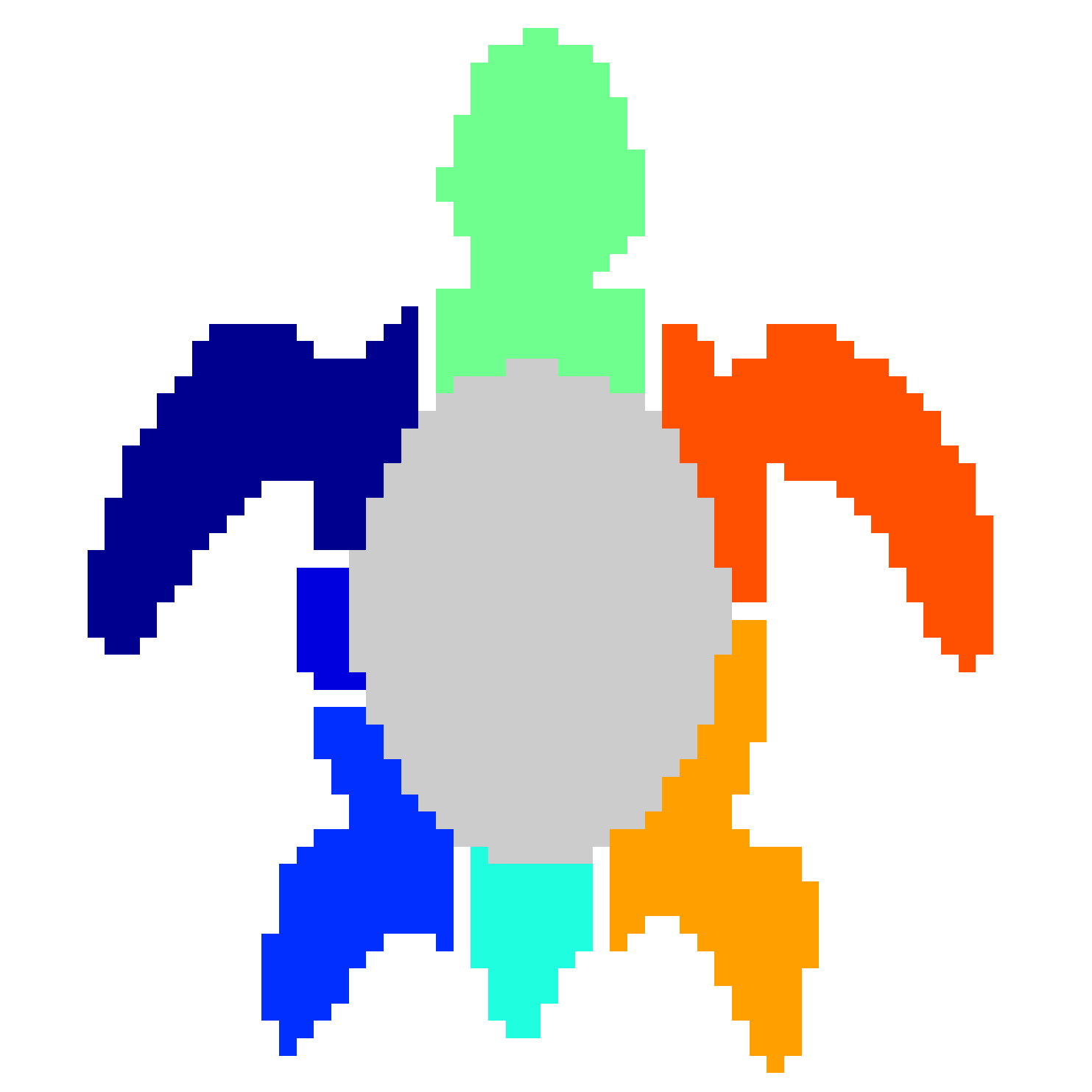}&
\includegraphics[height=3.2cm]{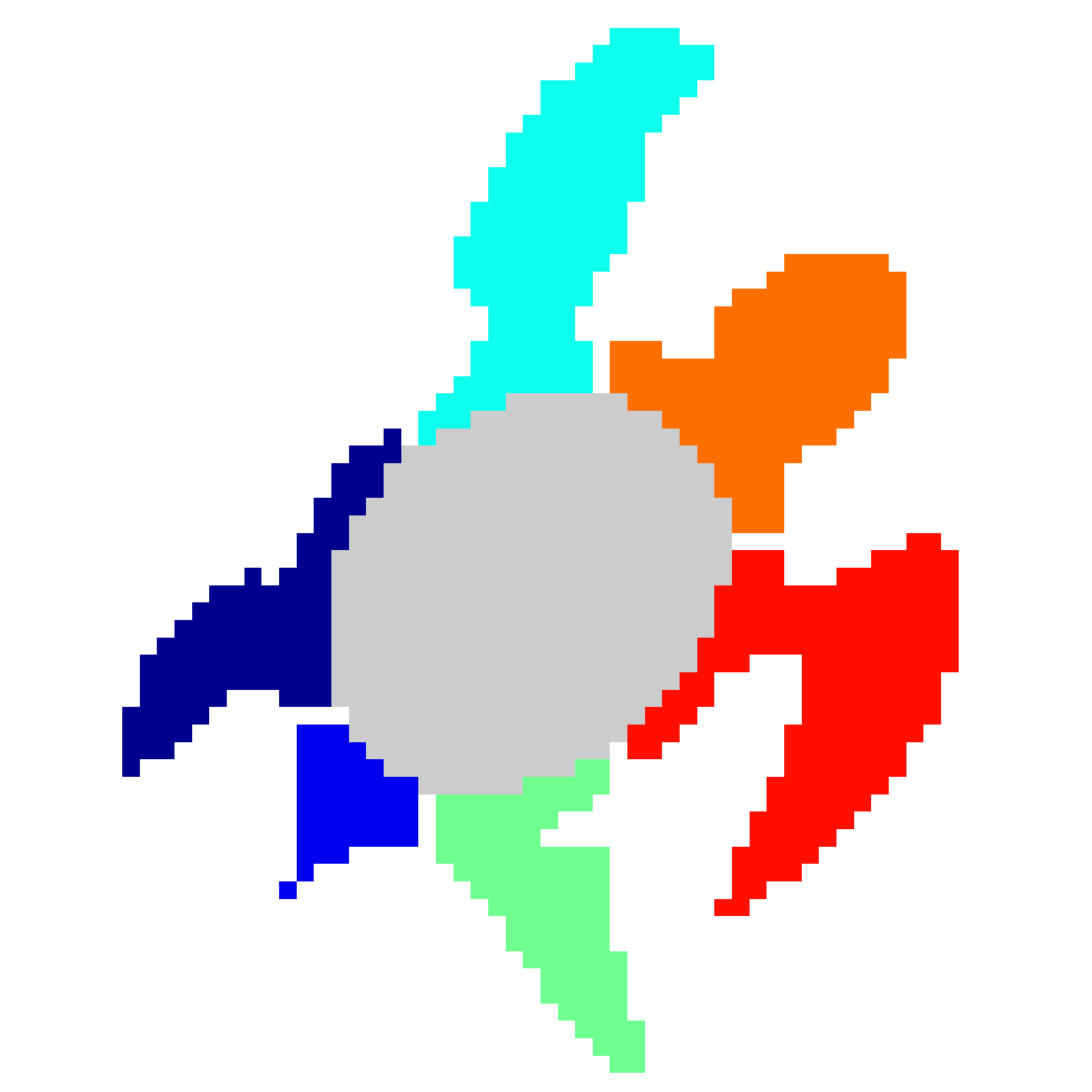}&
\includegraphics[height=3.2cm]{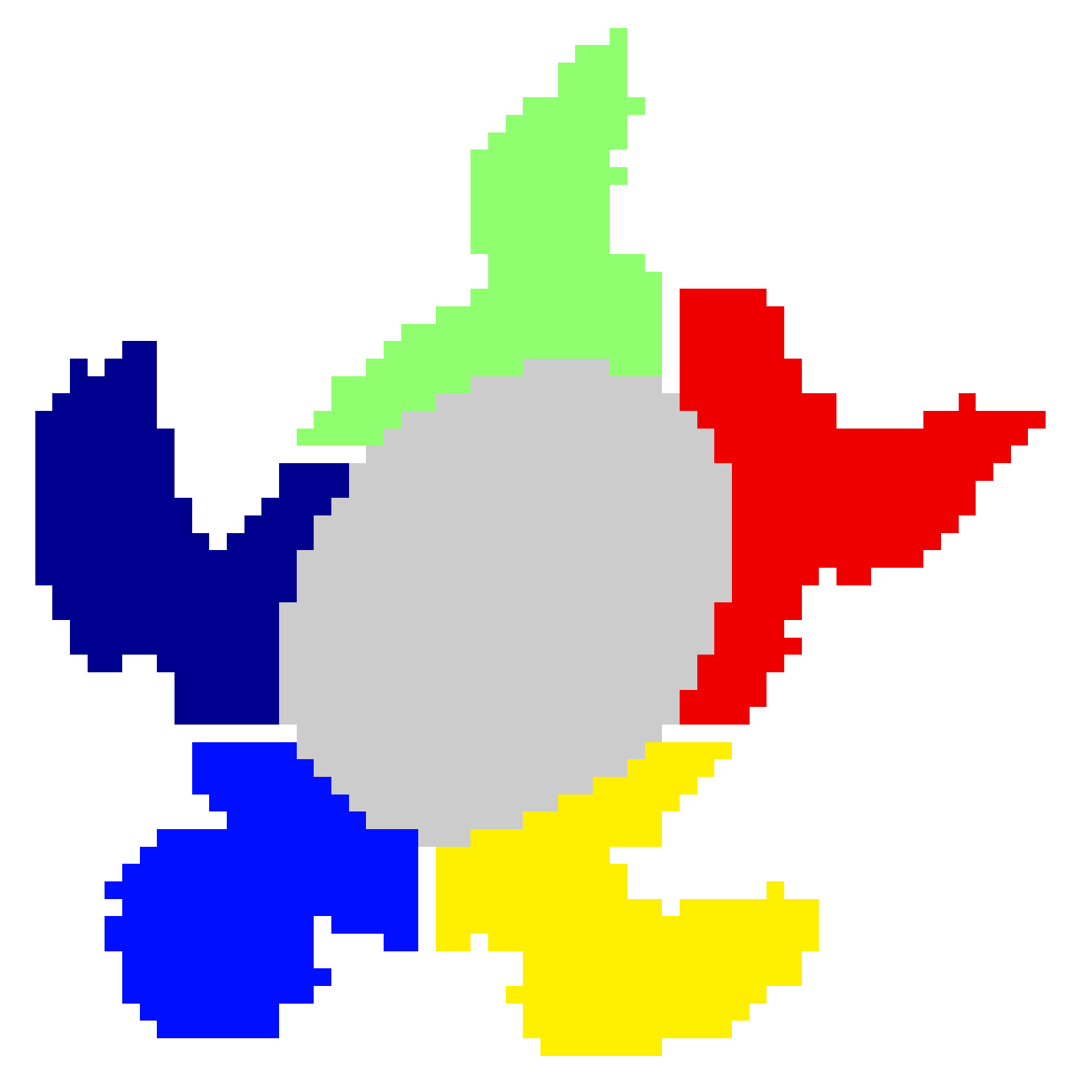}\\
\end{tabular}
\caption{Un-intuitive parts. See text. }
\label{fig:turtle}

\vglue 20pt

\centering
{\footnotesize
\begin{tabular}{cccc}
\includegraphics[height=2.5cm]{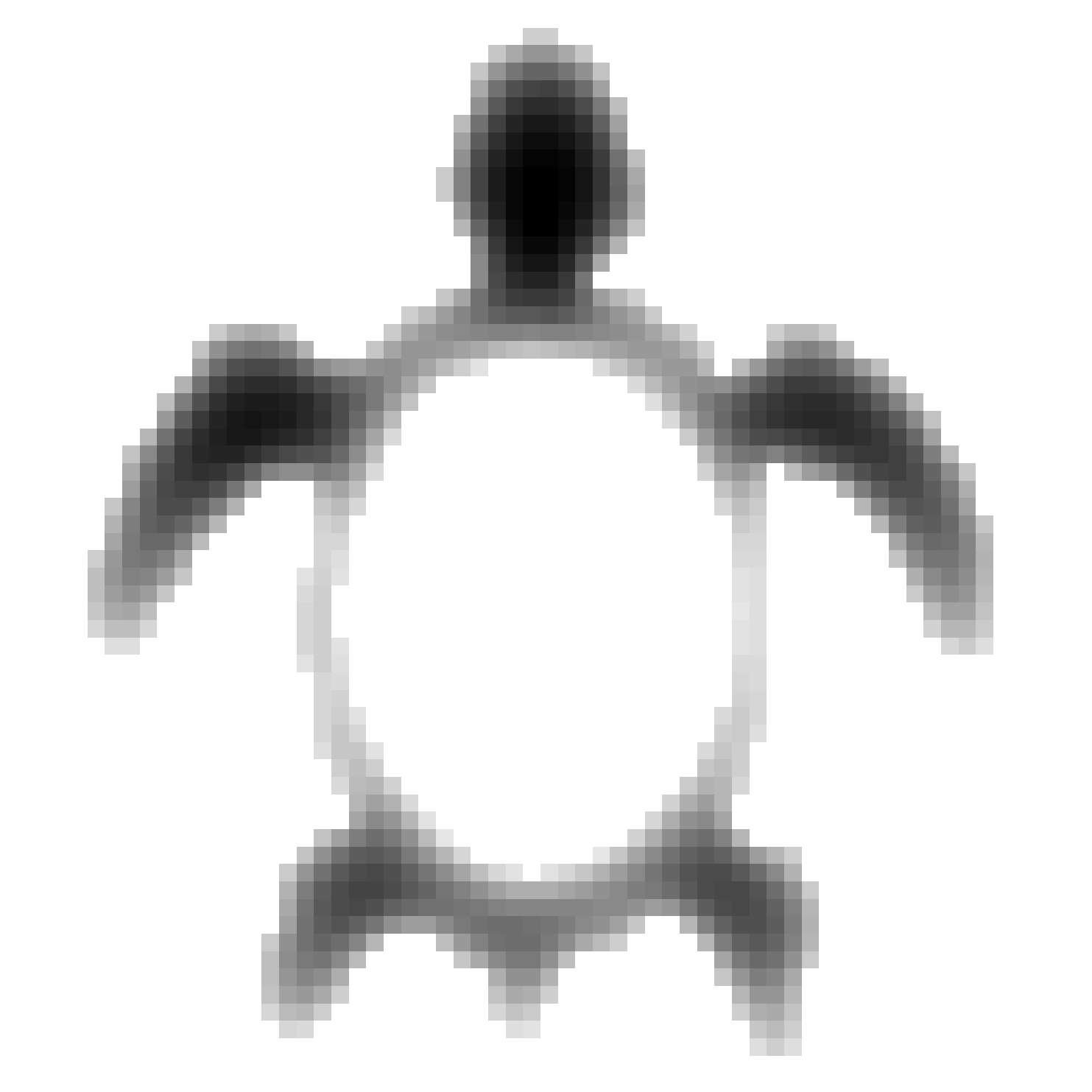}&
\includegraphics[height=2.5cm]{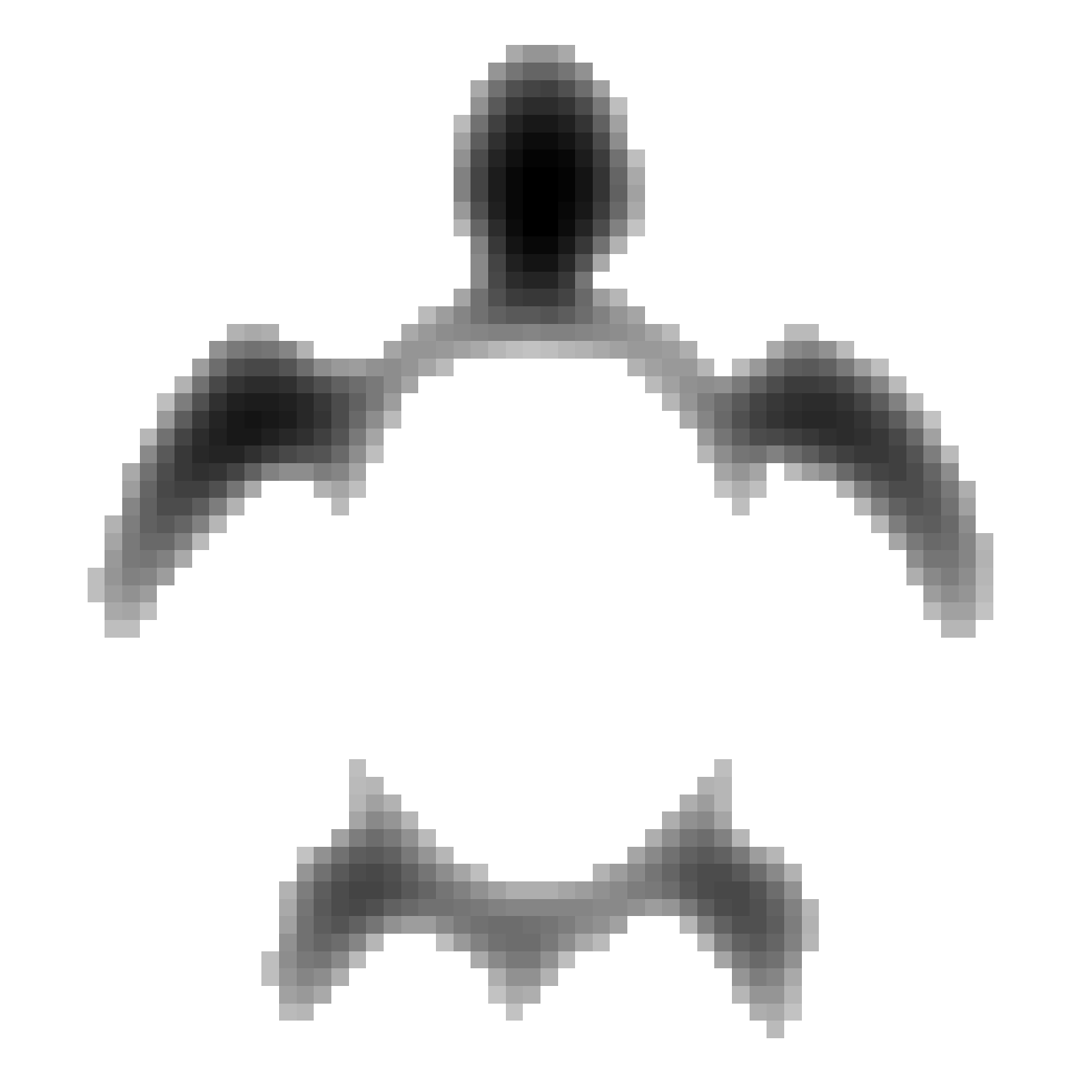}&
\includegraphics[height=2.5cm]{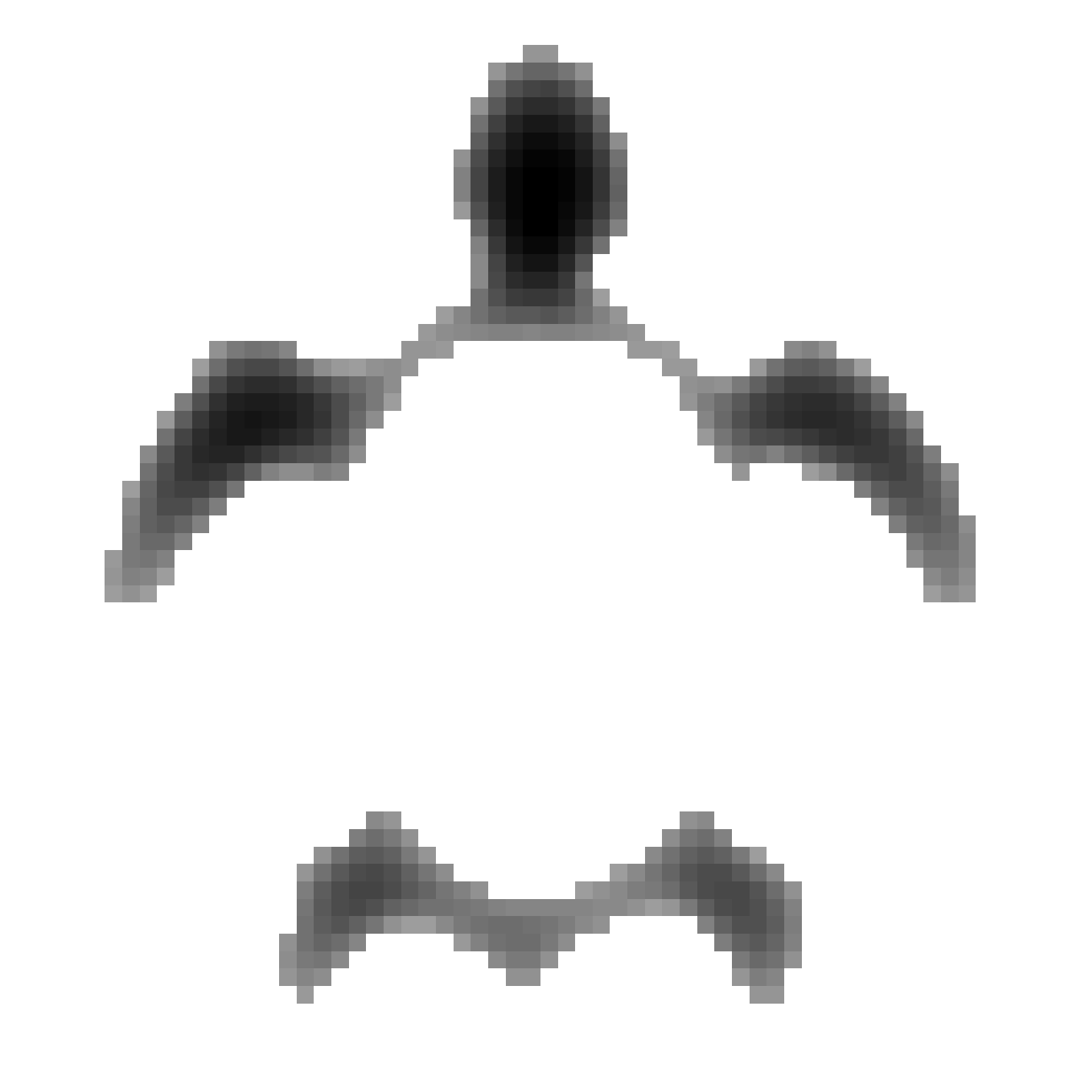}&
\includegraphics[height=2.5cm]{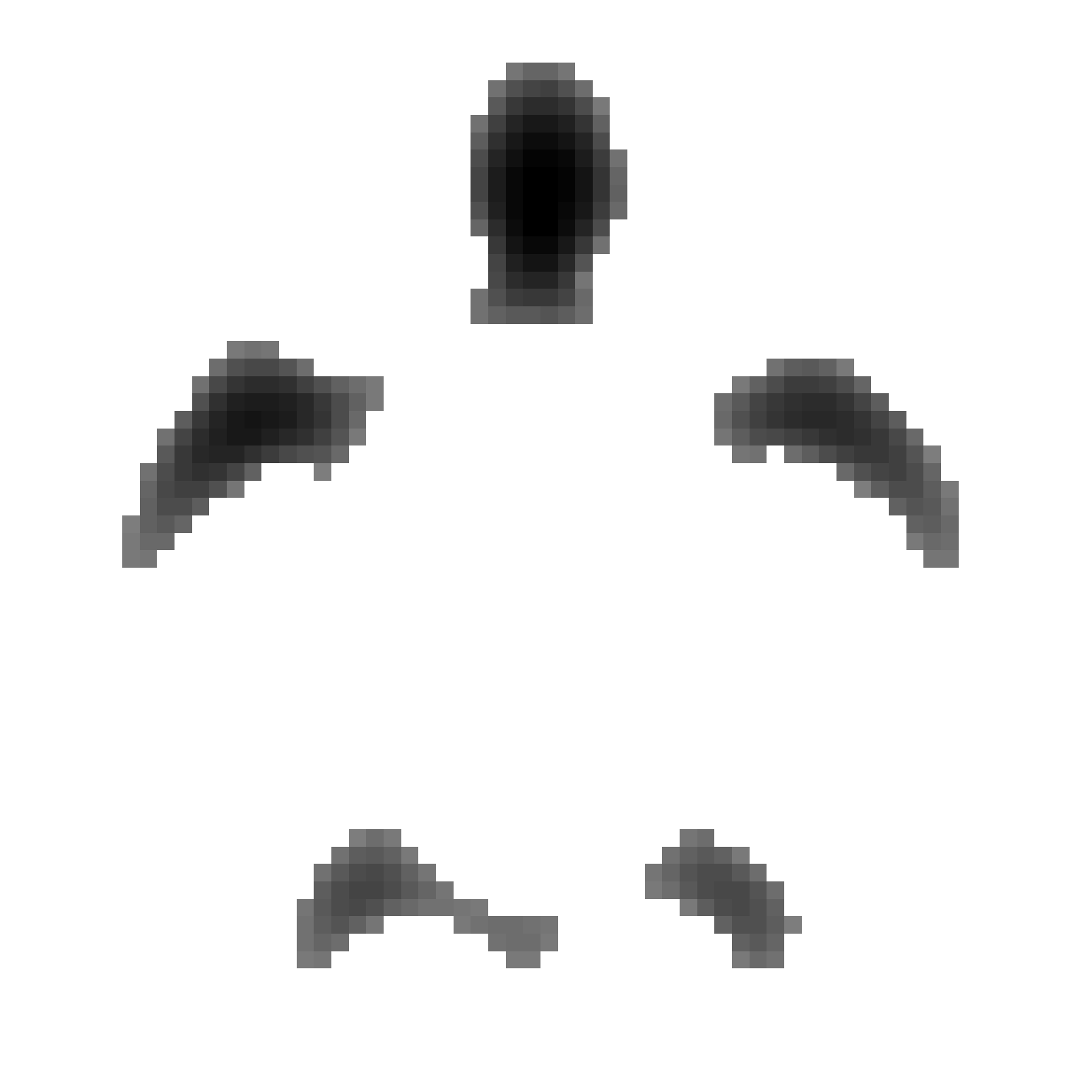}\\
(a)& (b) & (c)& (d)\\
\includegraphics[height=2.5cm]{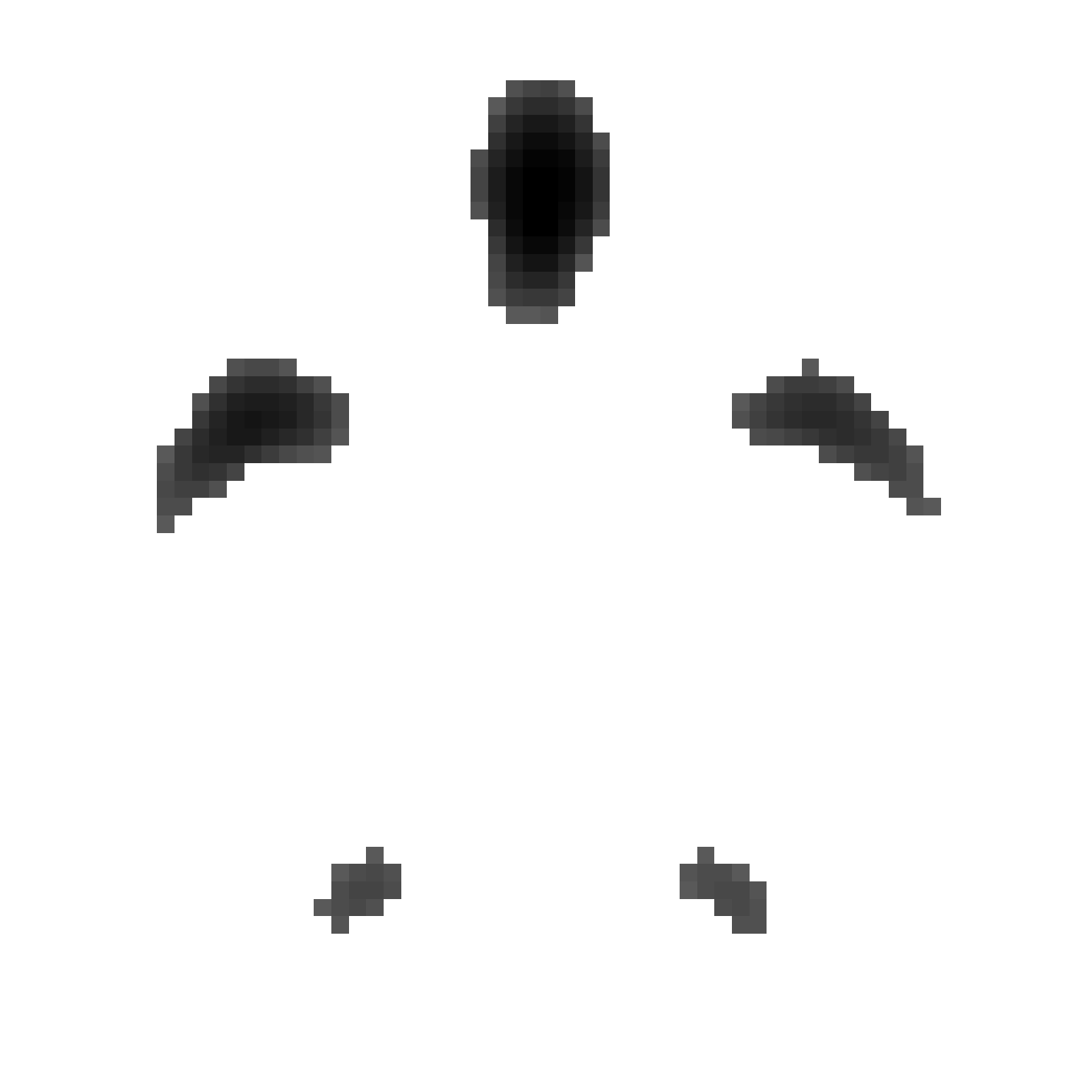}&
\includegraphics[height=2.5cm]{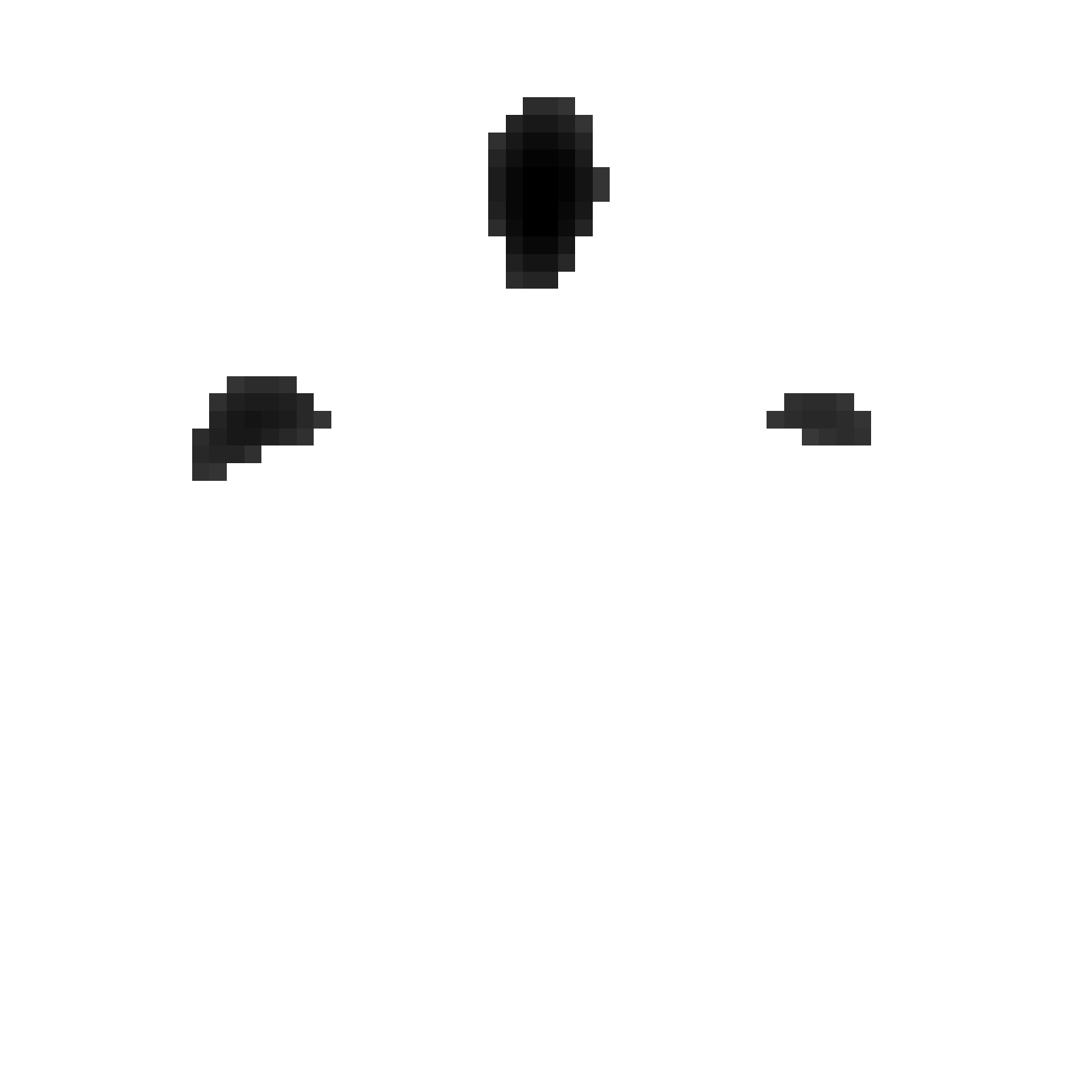}&
\includegraphics[height=2.5cm]{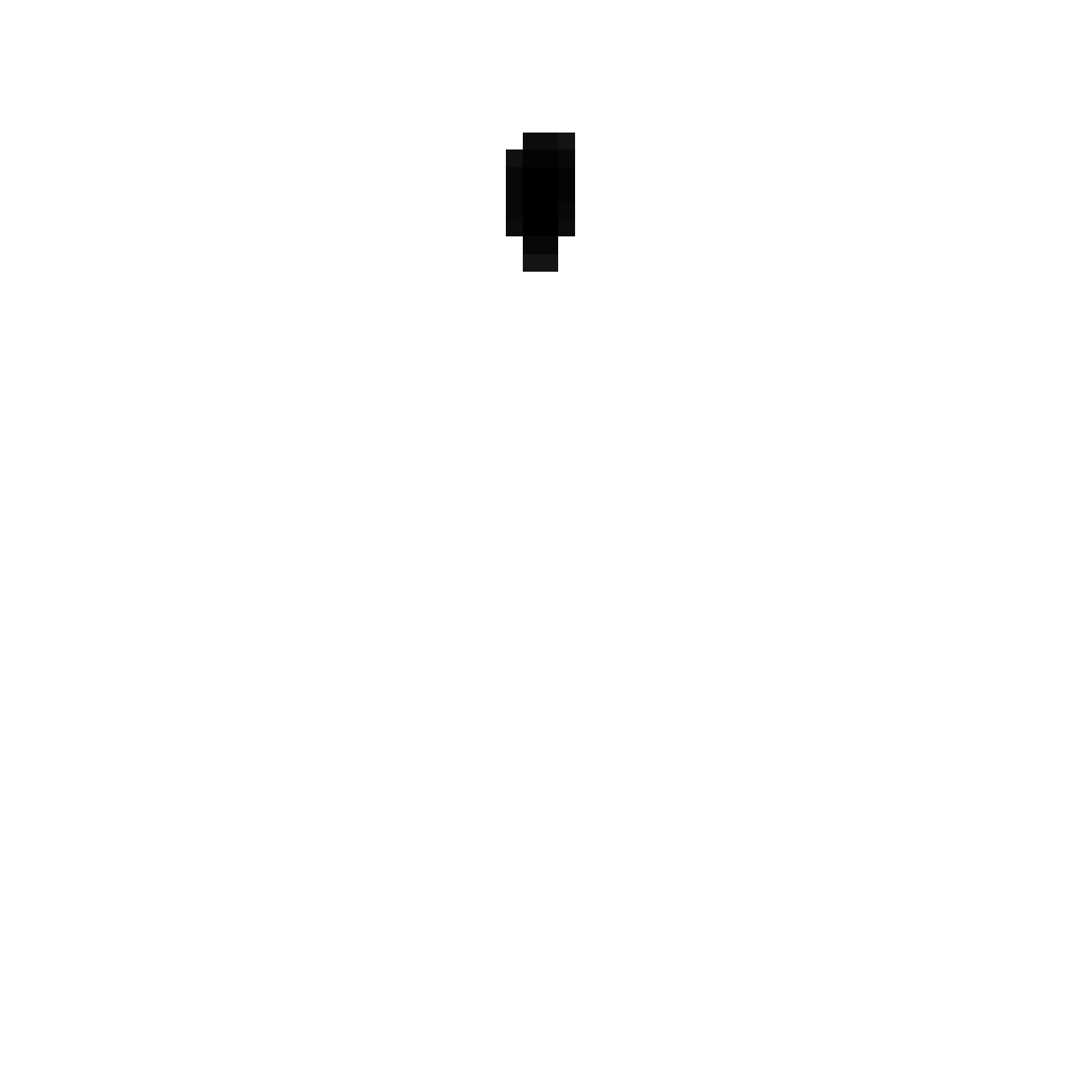}&
\includegraphics[height=2.5cm]{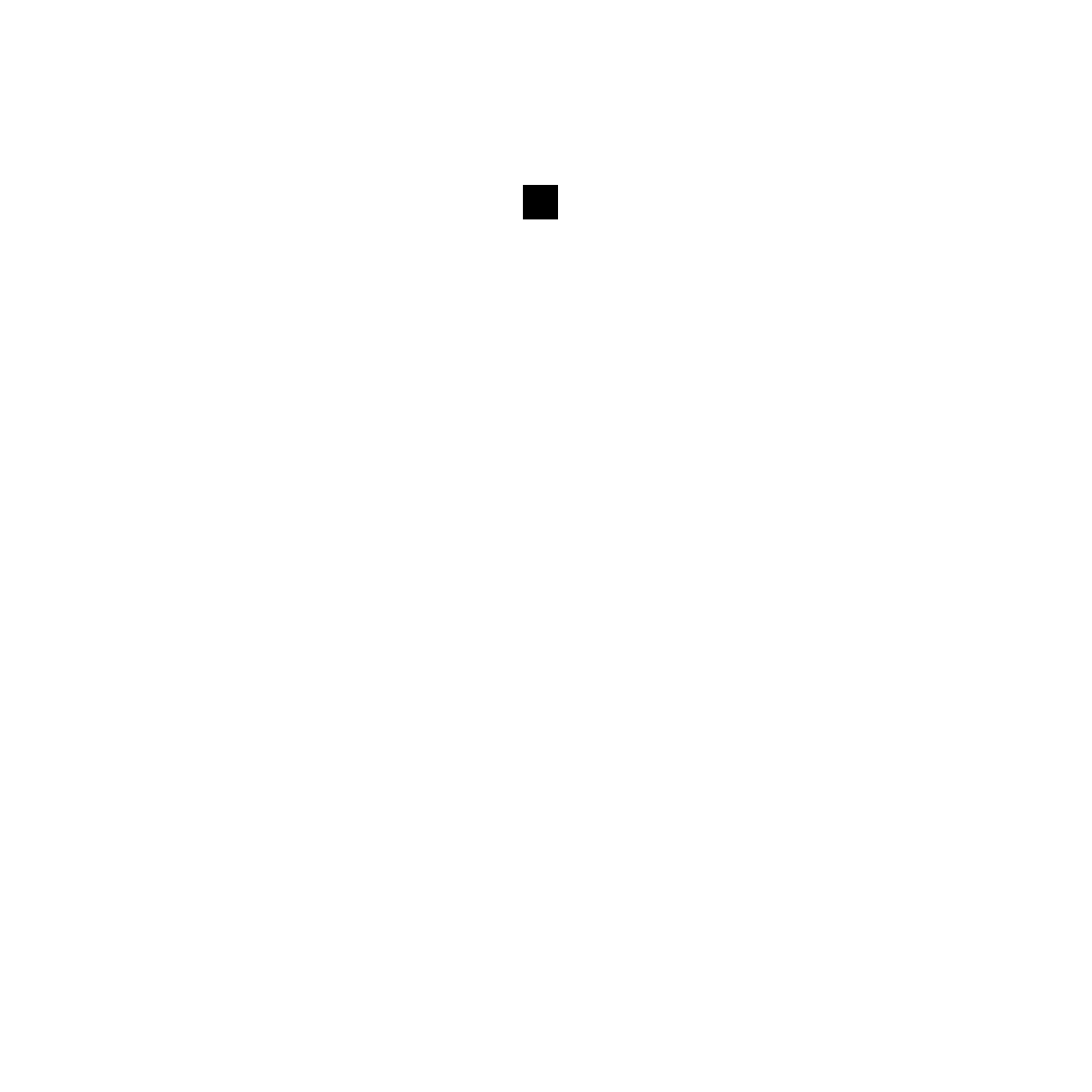}\\
(e)& (f) & (g)& (h)
\end{tabular}}
\caption{Saliency of a part.  In each figure, the restriction of
$\omega$ to locations, where its value is less than a given
threshold, is depicted. The thresold is increased gradually.
 (a)-(h) $\omega <-4,-9,-14,-19,-24,-29,-34,-37$, respectively. }
\label{fig:turtle_history}
\end{figure}

In Fig.~\ref{fig:resolution}, the robustness of the
method with respect to changes in resolution is demonstrated. An artificial shape
from Aslan~\cite{Aslan05,Aslantez,Aslan08} is used in its original resolution in (a), and in a
reduced resolution in (b).

In Fig.~\ref{fig:sample}, the  decomposition
results for a variety of shapes are provided. The decompositions are consistent; similar
shapes are partitioned similarly and the captured parts are
compatible with our intuition.

In some cases, as in Fig.~\ref{fig:turtle}, the decomposition
process starting from each and every local minima  ({\sl i.e.}
ignoring saliency) may create un-intuitive parts. Normalized
absolute value of $\omega$ for three sample turtle shapes are
depicted in the first row. In all of the three cases, one can easily
spot five local maxima in the peripheral part, and one local maxima
in the central part. However, the peripheral structure of the first
turtle shape is decomposed into seven  pieces. For the second
turtle, there are six pieces corresponding to the head, the two
legs, the two arms, and the tail. For the third turtle, there are
five pieces corresponding to the head, the two legs, the two arms.
In the third turtle, due to smoother transition between the two
legs, the tail part is missed. This produces inconsistencies among
silhouettes from the same category. However, inconsistent parts are
not as salient as the consistent parts. A clarifying illustration is
given  in Fig.~\ref{fig:turtle_history} with the help of the first
turtle which is decomposed into eight pieces including the torso. A
growth process is simulated starting from
Fig.~\ref{fig:turtle_history}~(h) and ending at
Fig.~\ref{fig:turtle_history}~(a).
 In each sub figure, the restriction of $\omega$ to the
locations, where its value is less than a given threshold  is
depicted. The threshold is increased gradually.
At the first threshold level in (h), only the head appears.
At the
third threshold level in (f), the arms appear. The head still continues
as an individual blob. At the fourth threshold in (e) level the two legs
appear. The five pieces remain separate.
At the next threshold level in (d) there are still five pieces.
However, notice that in somewhere between (e)
and (d), the tail piece comes to existence and then gets combined with the
rightmost leg.
\index{saliency}
It is appropriate to  say that the saliency of the tail is quite low compared
to the saliency of the other five pieces, due to its short life span as an individual entity.
At the next threshold level in (c), the two legs combine to form a single
piece corresponding to the lower body; and the two arms and the head combine to form
the upper body.
The separation of the peripheral structure into upper and lower body divides
the elliptical gross structure from the minor axis.
Finally in
(a),  the upper and lower parts combine to form a single peripheral
structure.

\begin{figure}[b]
\centering
{\footnotesize\begin{tabular}{cccc}
\includegraphics[height=2.45cm]{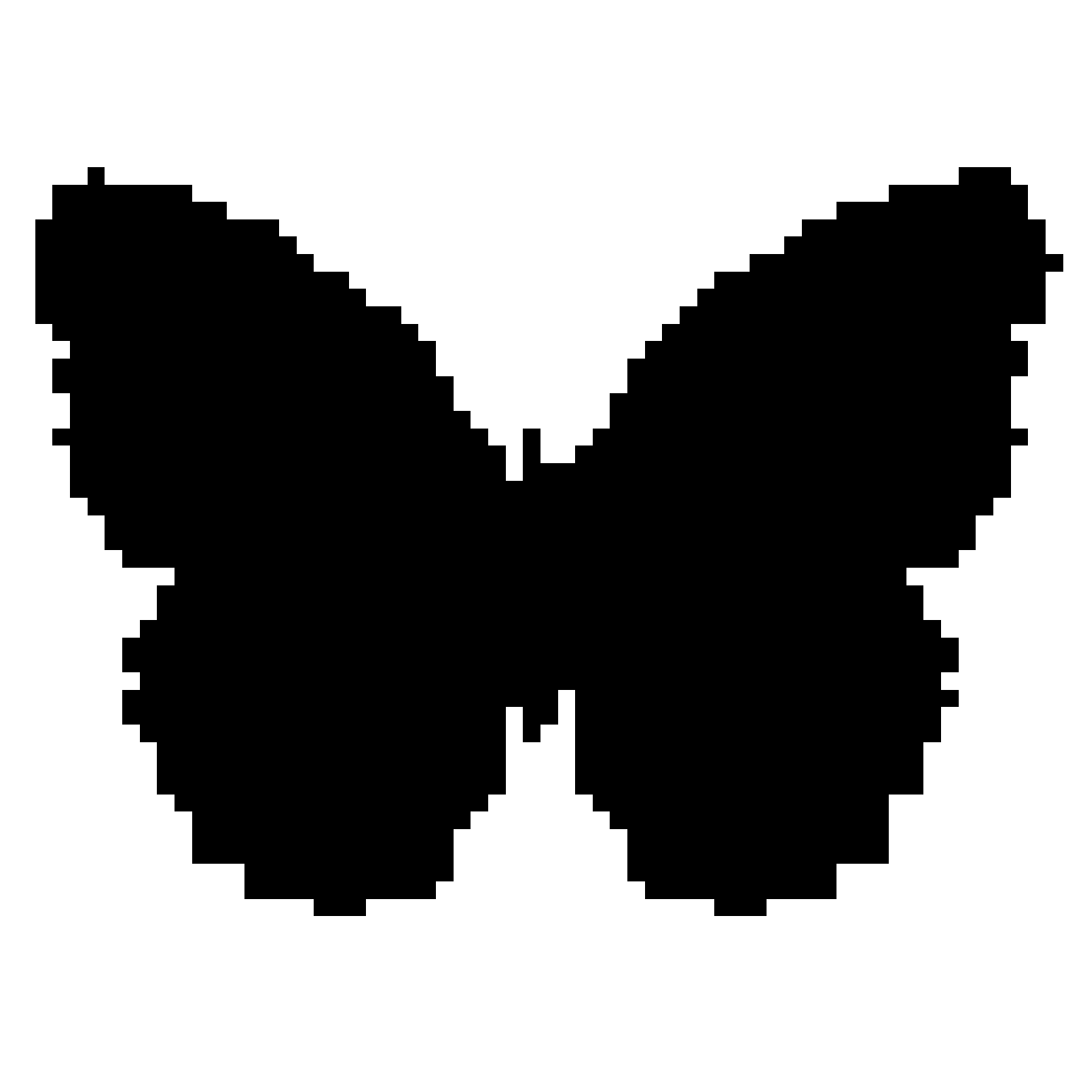}&
\includegraphics[height=2.45cm]{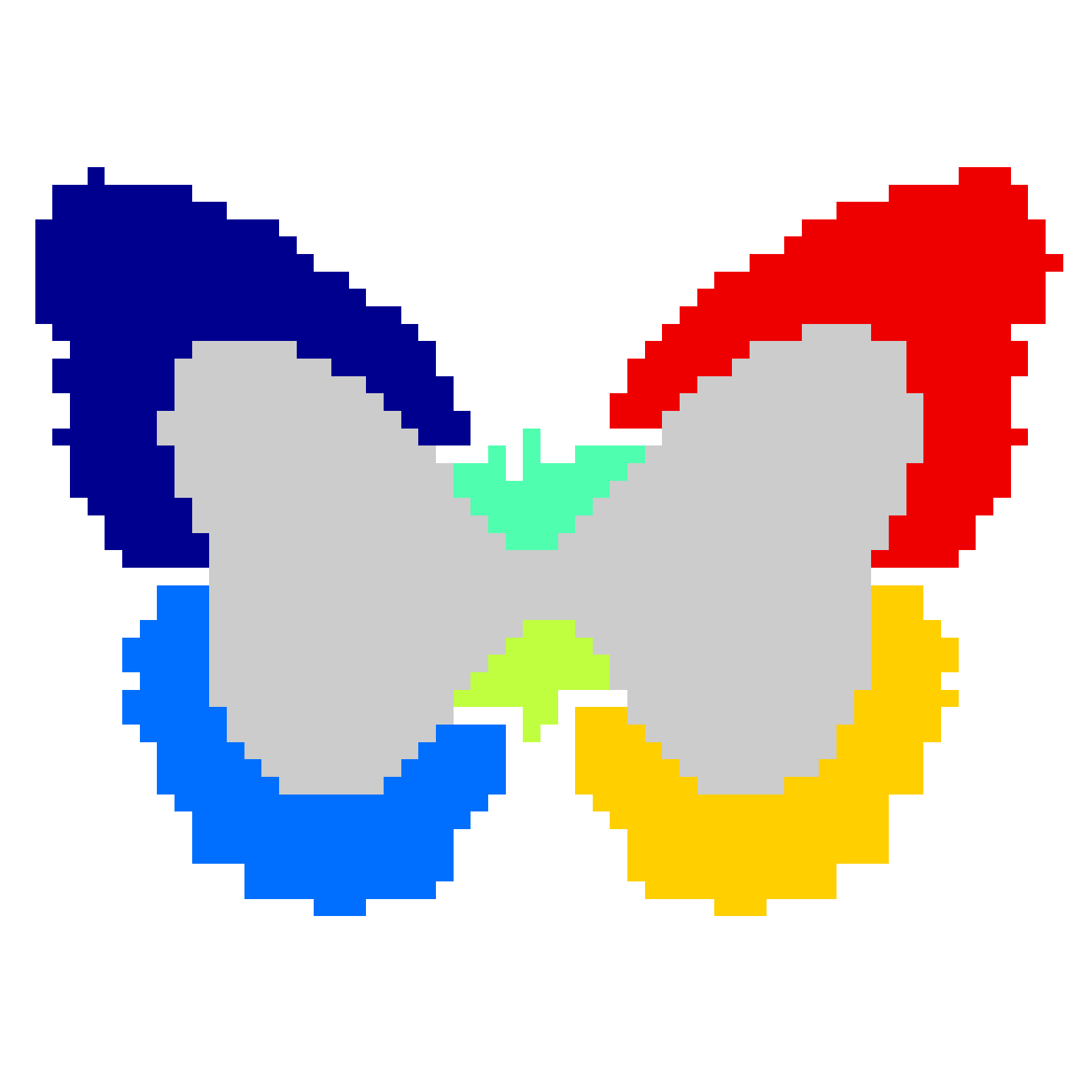}&
\includegraphics[height=2.45cm]{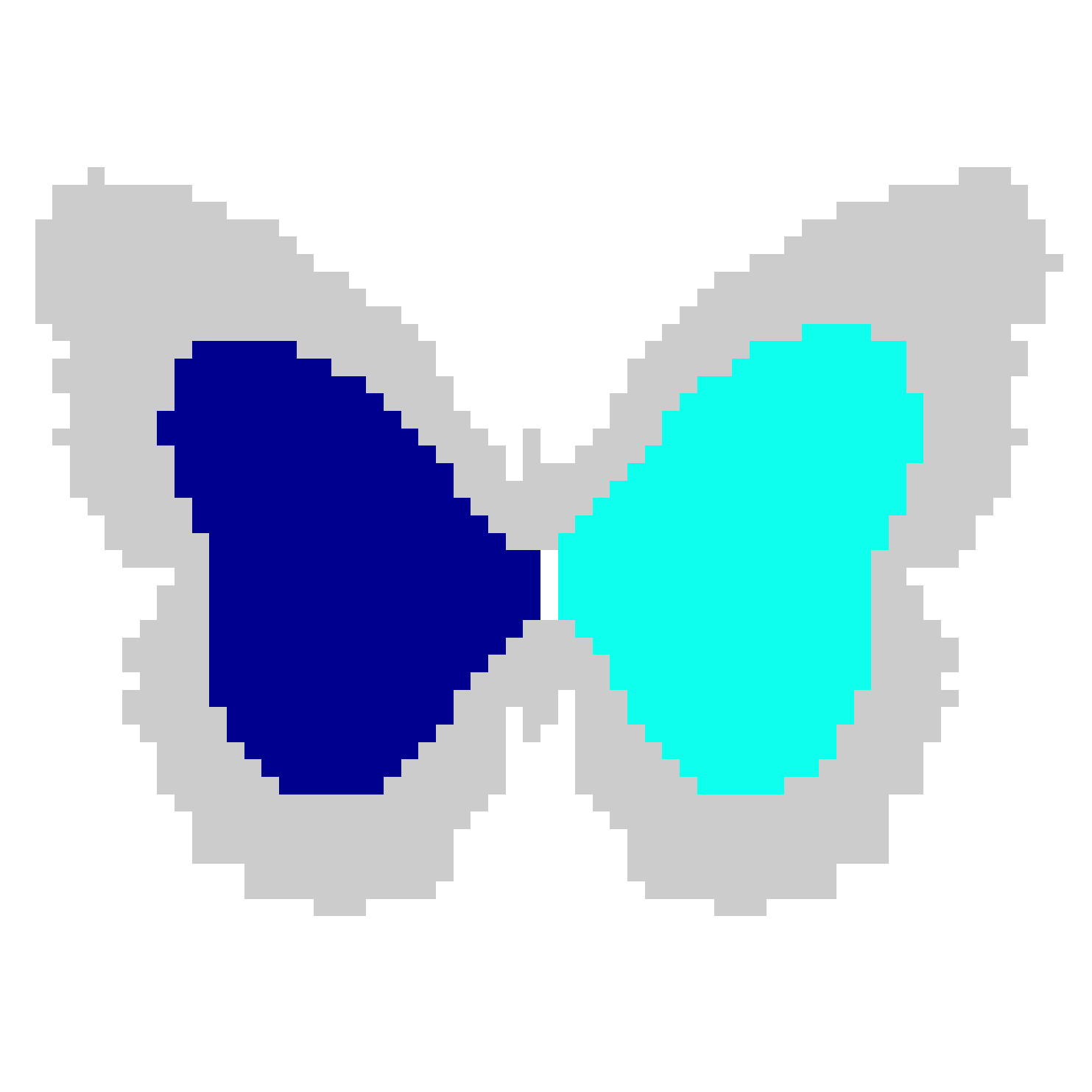}&
\includegraphics[height=2.45cm]{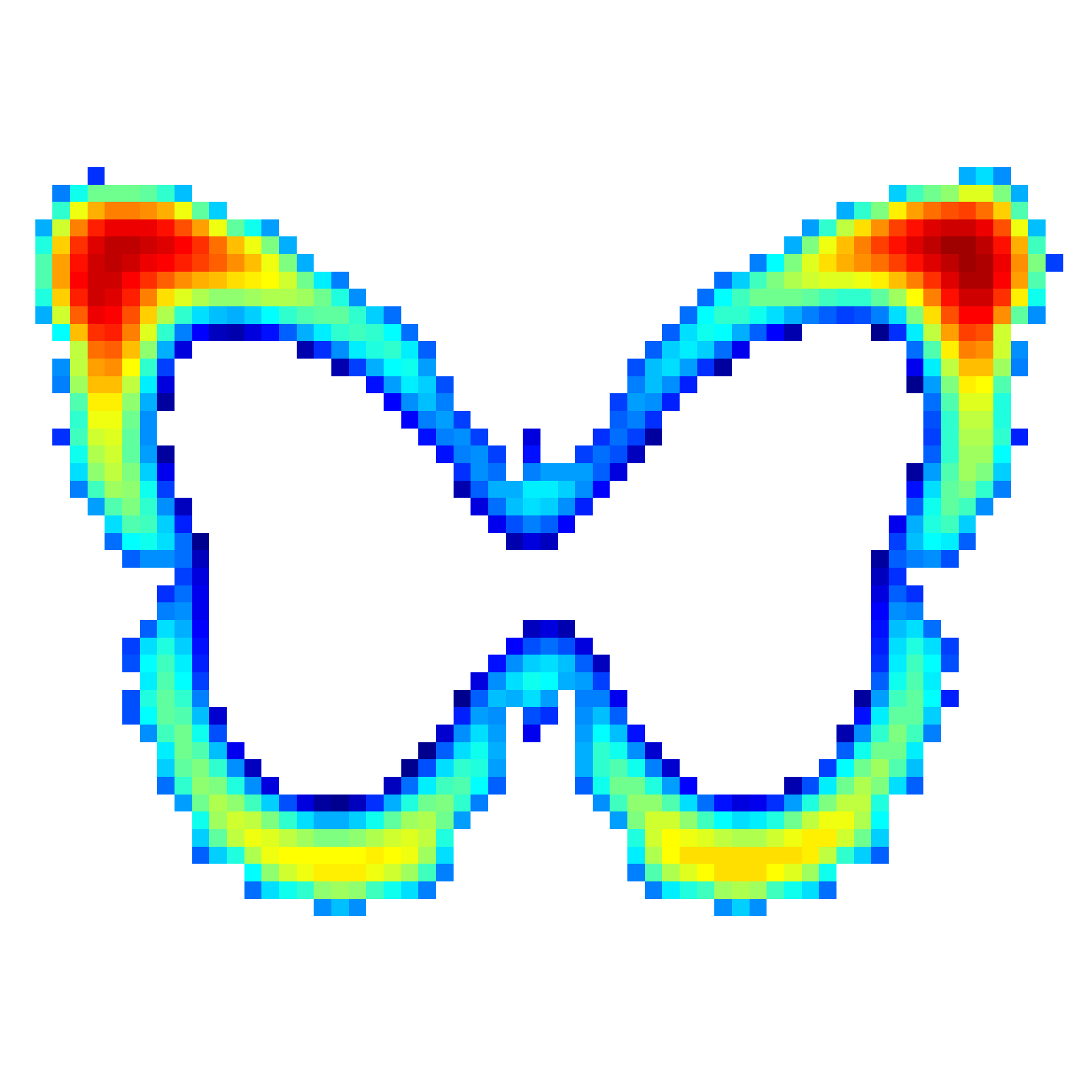}\\
(a) & (b) & (c) & (d)
\end{tabular}}
\caption{A shape whose gross structure is composed of  two blobs.
(a) A butterfly shape on a $60 \times 60$ lattice. (b-c) The parts.
(d) The restriction of $\omega$ to areas where the values are
negative.} \label{fig:kelebek10}

\vglue 18pt

\centering
{\footnotesize\begin{tabular}{cccc}
\includegraphics[height=2.45cm]{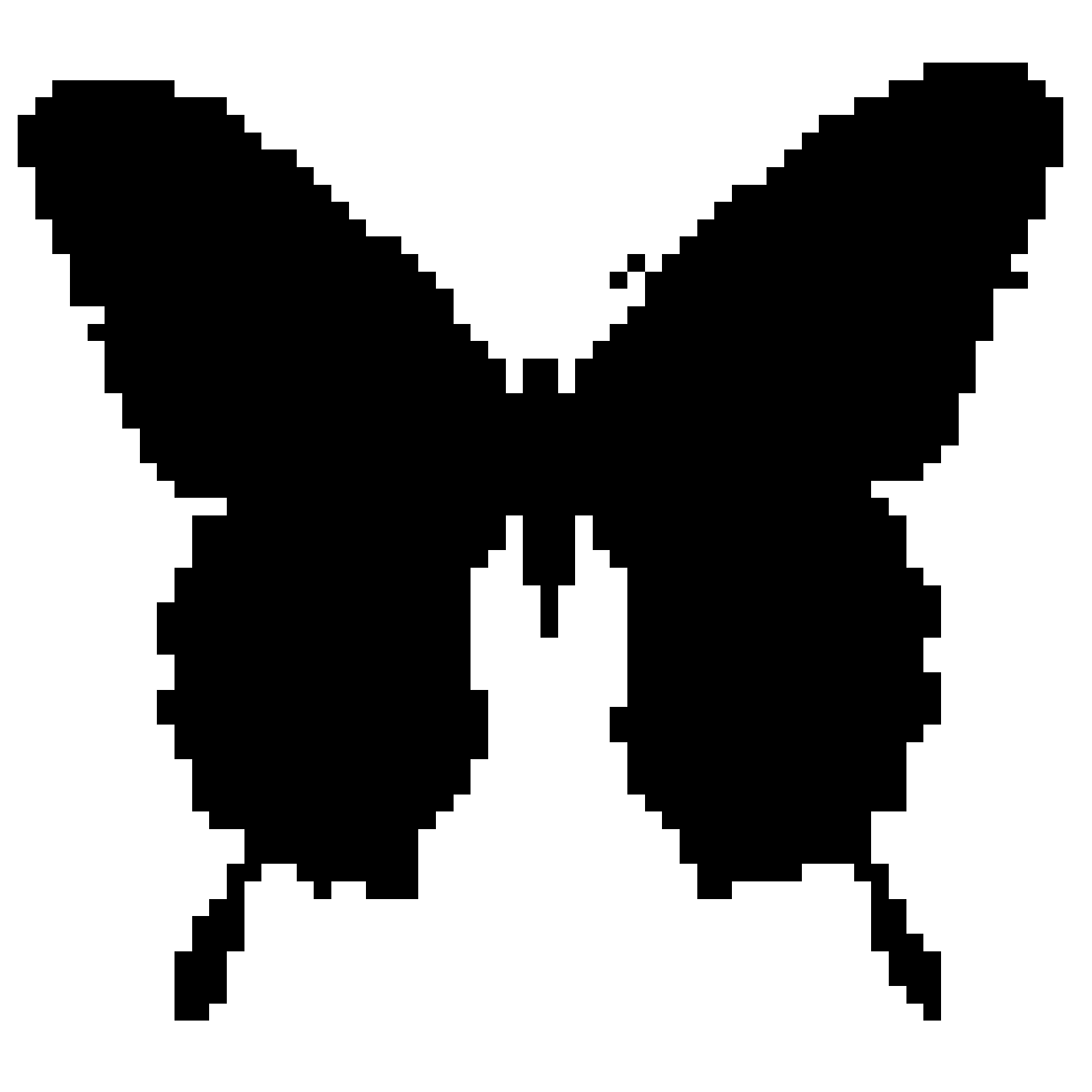}&
\includegraphics[height=2.45cm]{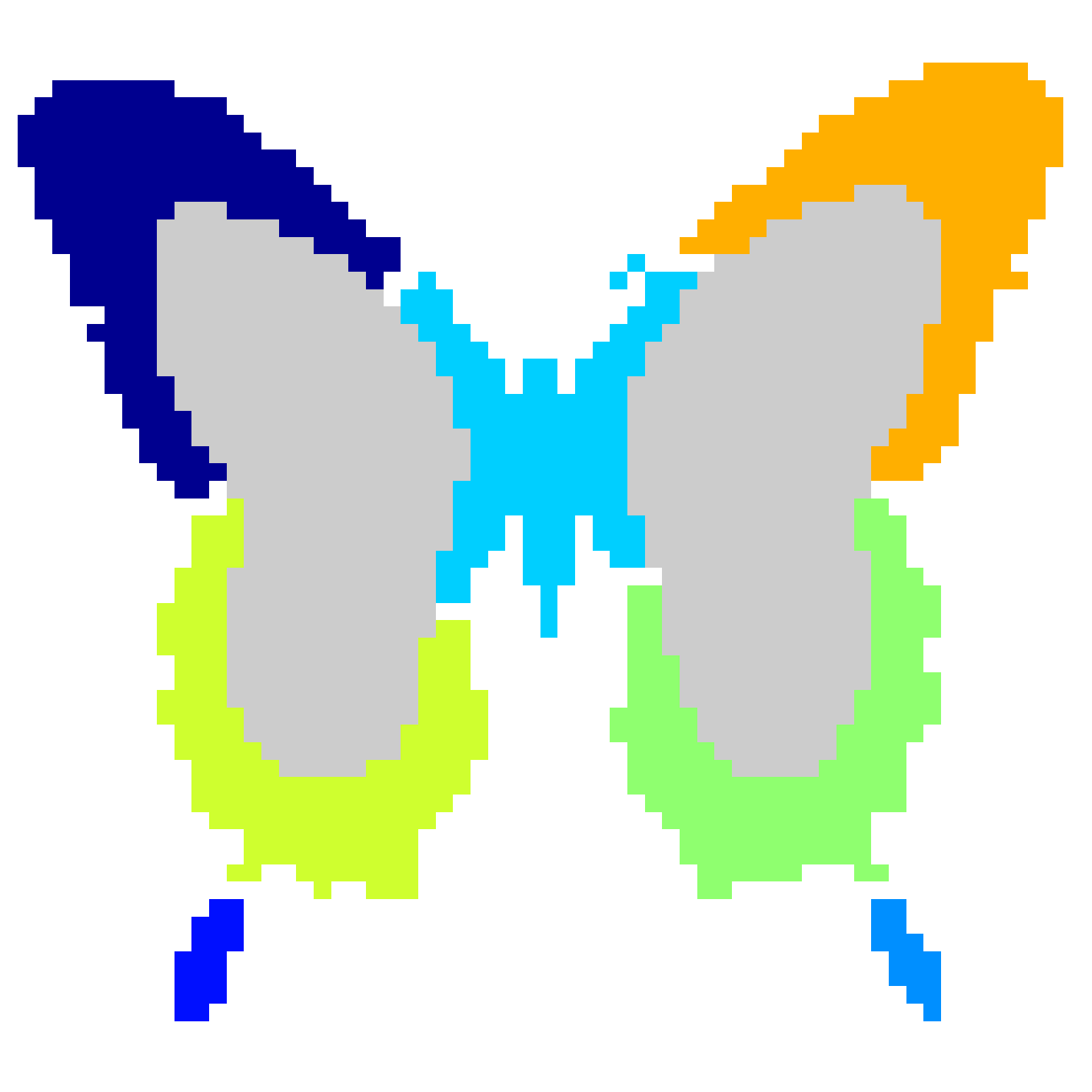}&
\includegraphics[height=2.45cm]{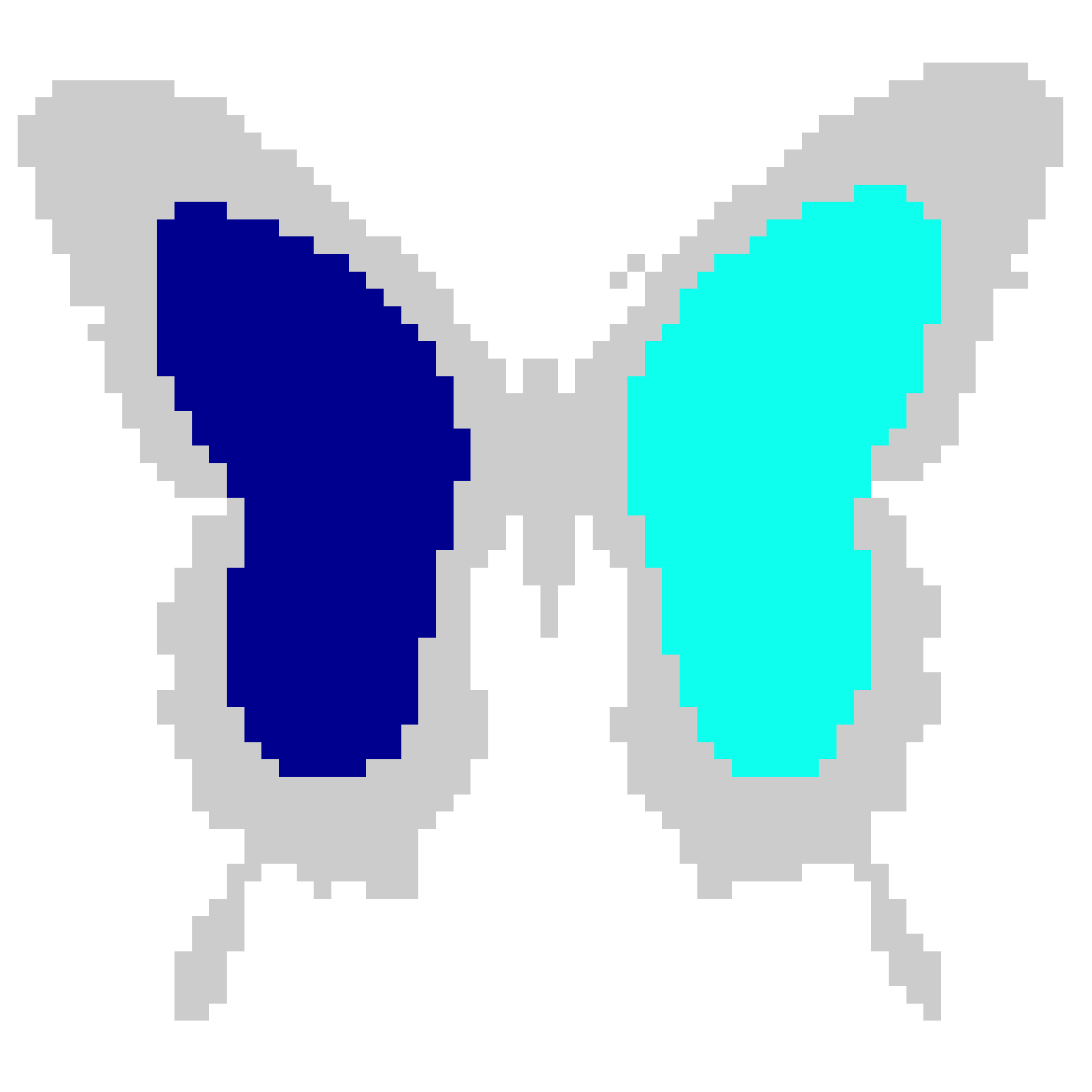}&
\includegraphics[height=2.45cm]{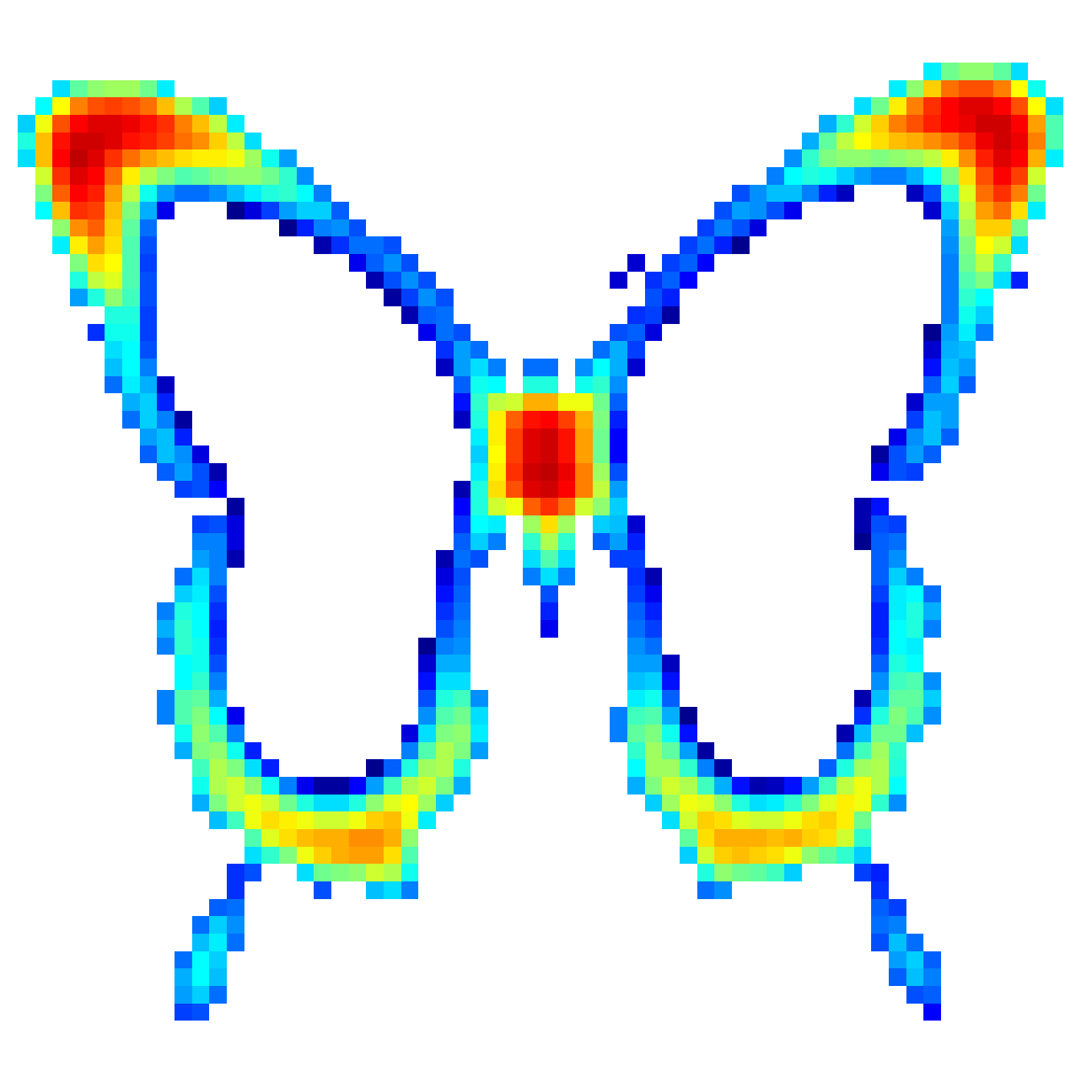}\\
(a) & (b) & (c) & (d)
\end{tabular}}
\caption{When the neck gets thinner, the gross structure may split
into two disjoint sets.} \label{fig:kelebek8}
\end{figure}

In all of the previous examples, the gross structure is shown in
gray, as a single part. There may be shapes such that the gross
structure is composed of multiple parts, i.e. shapes with strong
necks. The parts of the butterfly  shape in Fig.~\ref{fig:kelebek10}
(a) are shown in (b) and (c). In (b), the parts in the peripheral
structure are shown in bright colors; the gross structure is in
gray. In (c) the parts in the gross structure are shown in bright
colors; the peripheral structure is in gray. For reference, the
restriction of $\omega$ to the peripheral structure, i.e. where it is
negative, is shown in (d). As the neck which connects the two lobes
of the butterfly gets thinner, the gross structure may be split into
two disjoint sets. (See Fig.~\ref{fig:kelebek8}.)

\begin{figure}[t]
\centering
{\footnotesize \begin{tabular}{ccc}
\includegraphics[height=2.9cm]{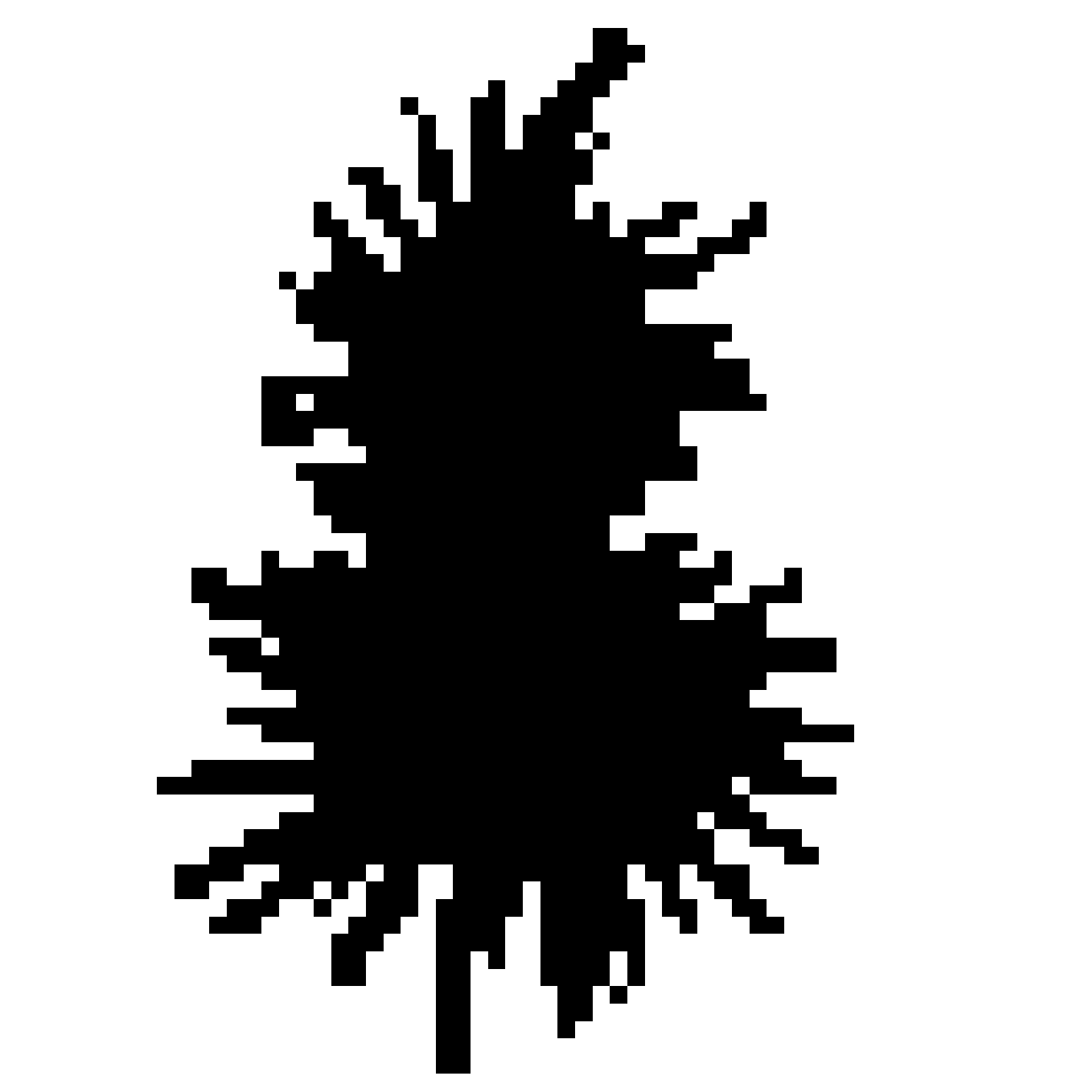}&
\includegraphics[height=2.9cm]{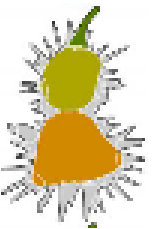}&
\includegraphics[height=2.9cm]{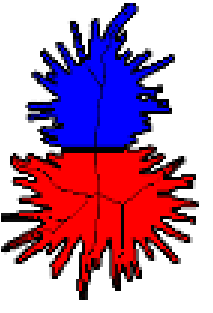}\\
(a)&(b) &(c)\\[6pt]
\includegraphics[height=2.9cm]{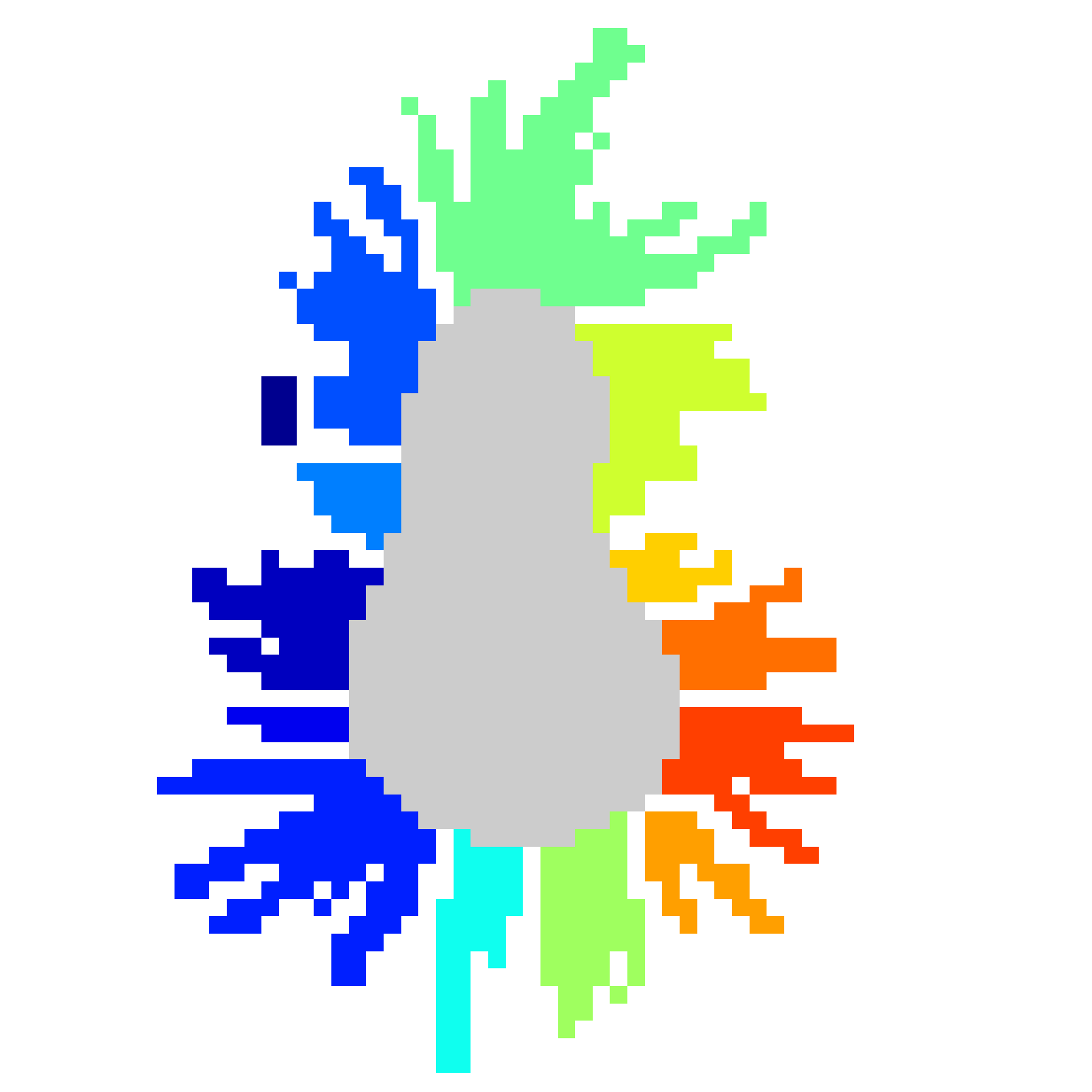}&
\includegraphics[height=2.9cm]{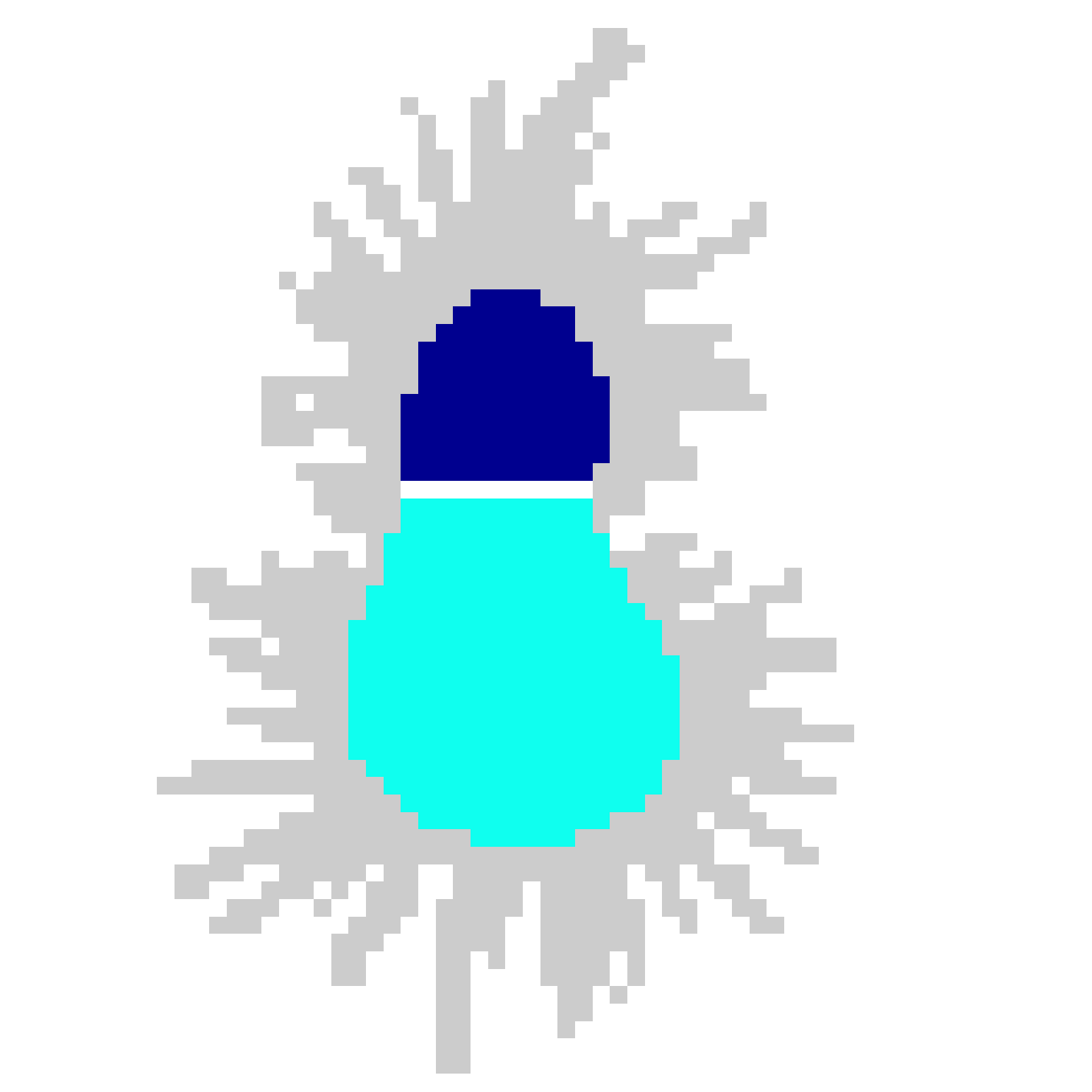}&
\includegraphics[height=2.9cm]{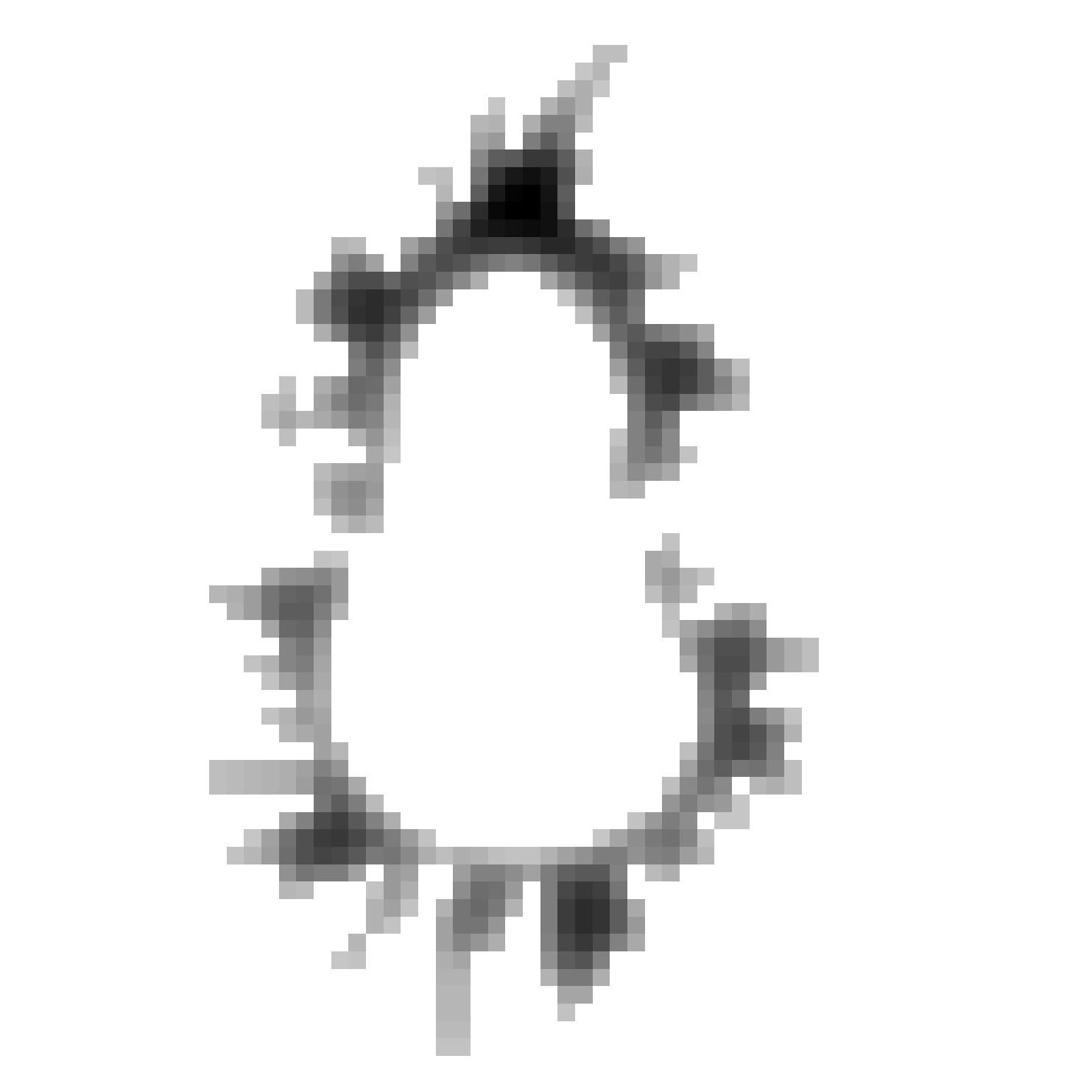}\\
(d)& (e) & (f)
\end{tabular}}
\caption{A noisy shape composed of two blobs. (a) The prickly pear
on a $80 \times 80$ lattice.  (b-c) The decompositions presented
in~\cite{Mi07} and \cite{Zeng07}, respectively. (d-e) The
decomposition by the new method. (f) The restriction of $\omega$ to where
$\omega <-3$. See text. [(b-c) taken from the original sources~\cite{Mi07} and \cite{Zeng07}]}
\label{fig:prickly}
\end{figure}

\subsection{Comparison to Recent Decomposition Methods by Mi and DeCarlo~\cite{Mi07} and Zeng
{\sl et al.}~\cite{Zeng07}}

In two recent papers,  Mi and DeCarlo~\cite{Mi07}, and Zeng {\sl
et al.}~\cite{Zeng07} present decomposition methods which exploit
both the  skeleton and the boundary curvature information. Neither
of the methods consider a fully global context. \index{context} It is worth
comparing the three methods using an illustrative example: {\sl the
prickly pear}, which is shown in Fig.~\ref{fig:prickly}(a).
\index{shape!prickly}

The decompositions obtained in Mi and DeCarlo~\cite{Mi07}  and Zeng
{\sl et al.}~\cite{Zeng07}  using local symmetry axis and contour
curvature are shown in (b) and (c), respectively. The new
decomposition (using a reduced $80 \times 80$ resolution) is shown
in (d) and (e). All of the three  methods find two blobs. Similar to
the new decomposition, the decomposition by Mi and
DeCarlo~\cite{Mi07} separates  the boundary texture (shown in gray
in (b)) from the main structure. On the other hand, the
decomposition by Zeng {\sl et al.}~\cite{Zeng07} does not separate
boundary texture from the main structure leading to the
interpretation of the shape as  two prickly balls glued together.

\index{shape!parsing}
\index{context}
\index{medial axis}
As experimental studies on human subjects
demonstrate~\cite{Renninger03}, multiple (and mutually exclusive)
parses of a given shape are possible. Thus, in a purely bottom-up
process without  considering a specific application or a context,
one can not decide which partitioning scheme is the best.

I remark that the advantages of the new method are purely from the
computational point of view. It does not involve any parameters or
thresholds.  It can work at very low resolutions as opposed to other
methods which involve the computation of curvature or local symmetry
axes, since their computation requires certain resolution.

\begin{figure}[ht]
\centering
{\footnotesize \begin{tabular}{ccc}
\includegraphics[height=2.9cm]{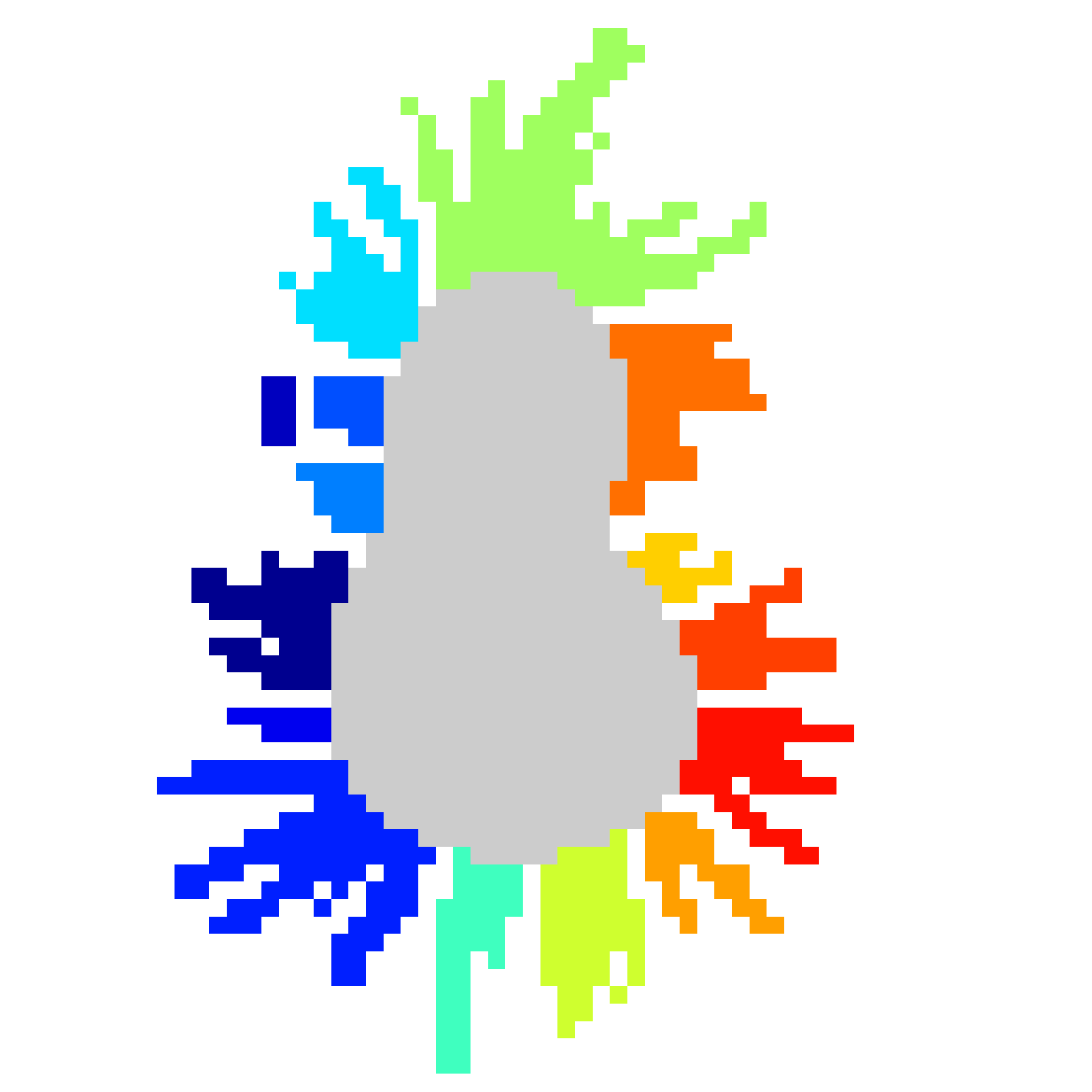}&
\includegraphics[height=2.9cm]{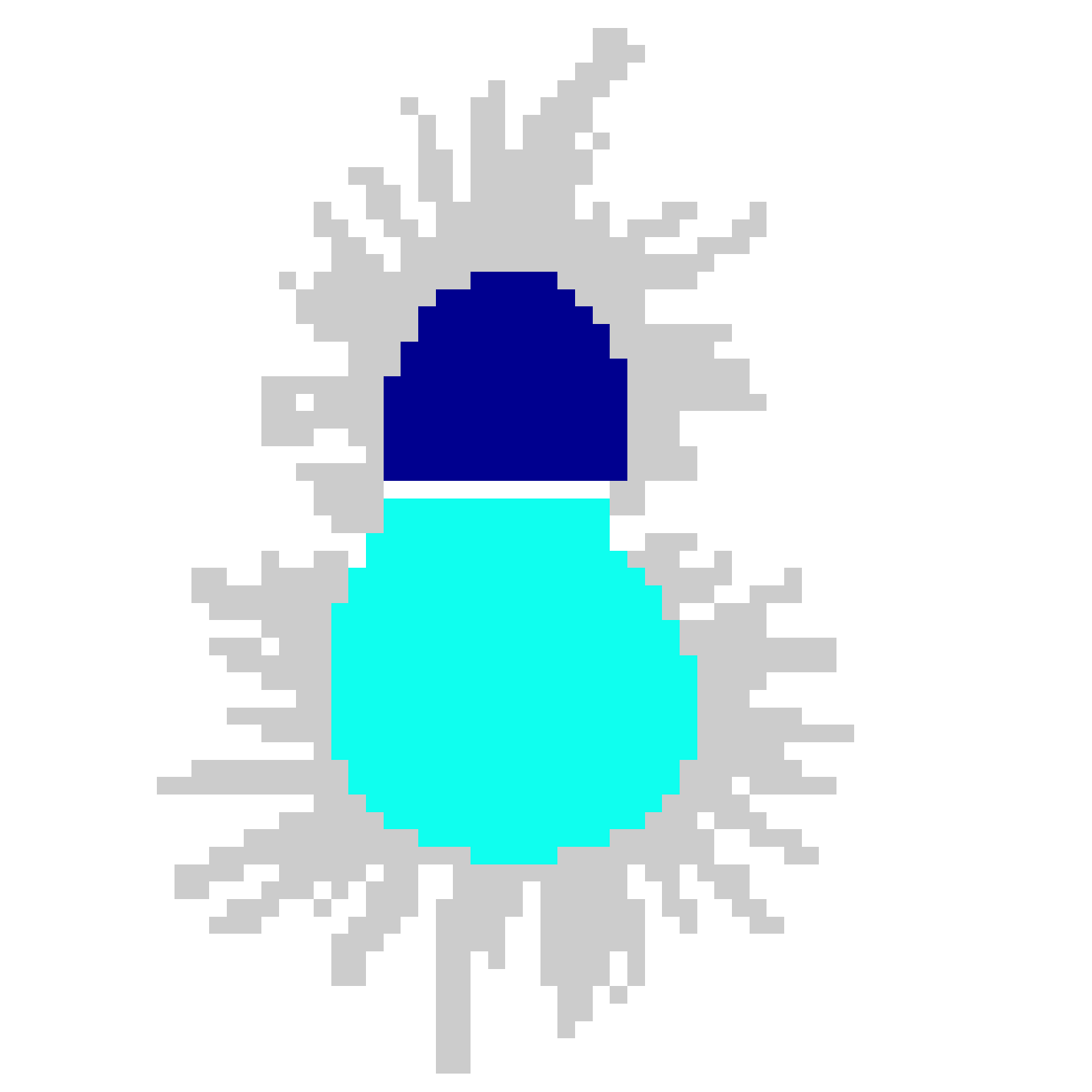}&
\includegraphics[height=2.9cm]{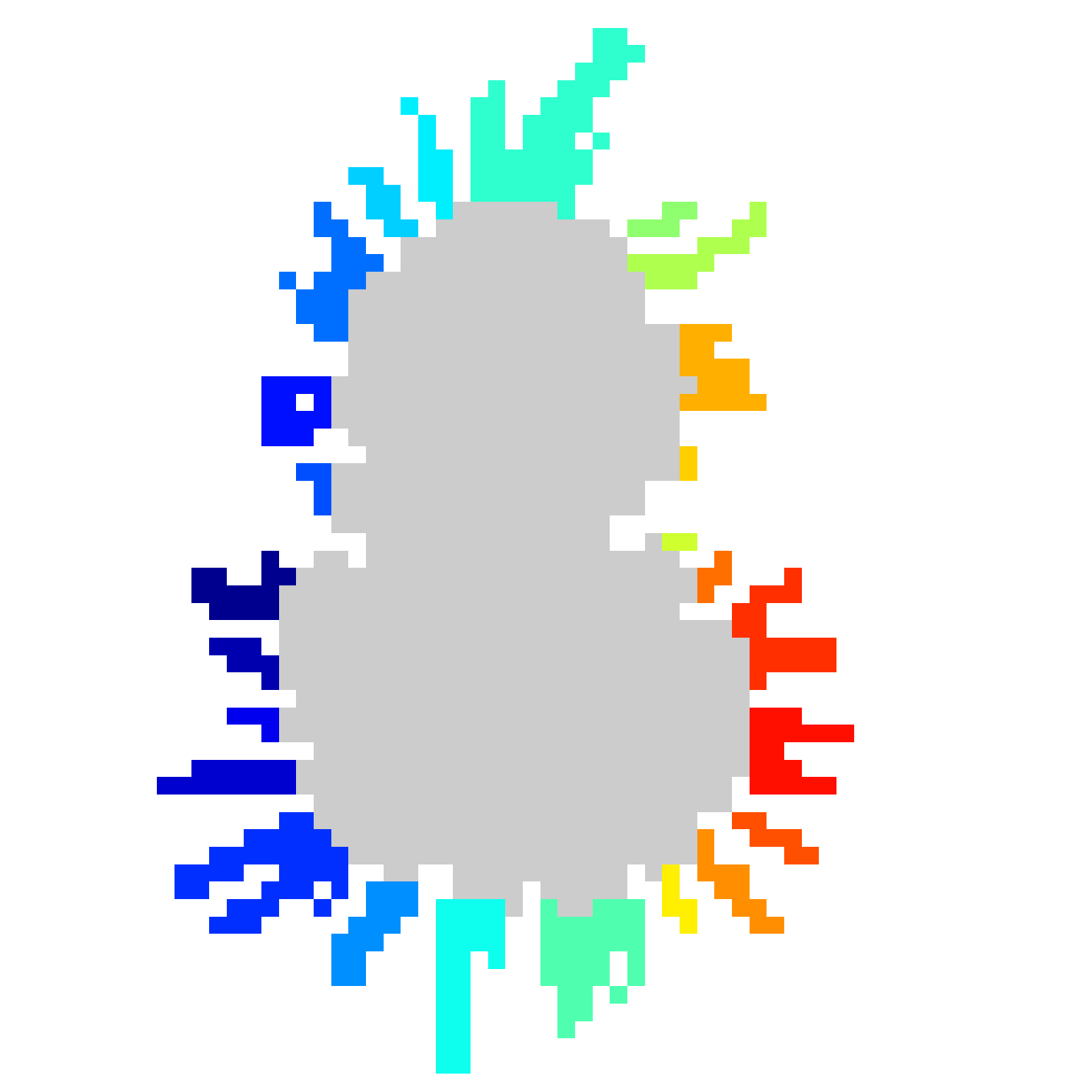}\\
(a)&(b) &(c)\\
\end{tabular}}
\caption{The effect of reducing the importance of the global term.
(a-b) $c=0.5$. (c) $c=0.125$. } \label{fig:prickly_param}

\vglue 18pt

\centering
\begin{tabular}{ccc}
\includegraphics[height=2.9cm]{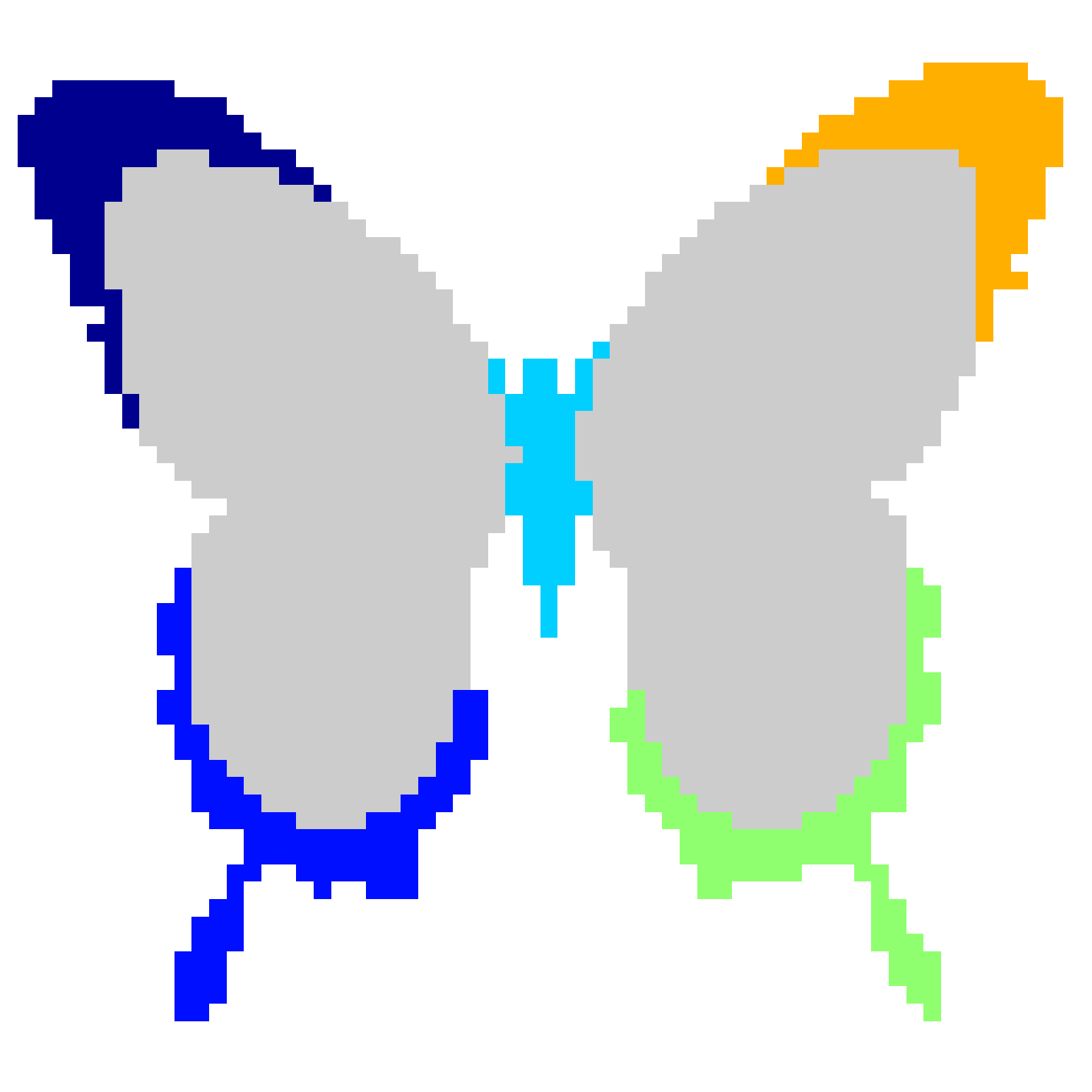}
\includegraphics[height=2.9cm]{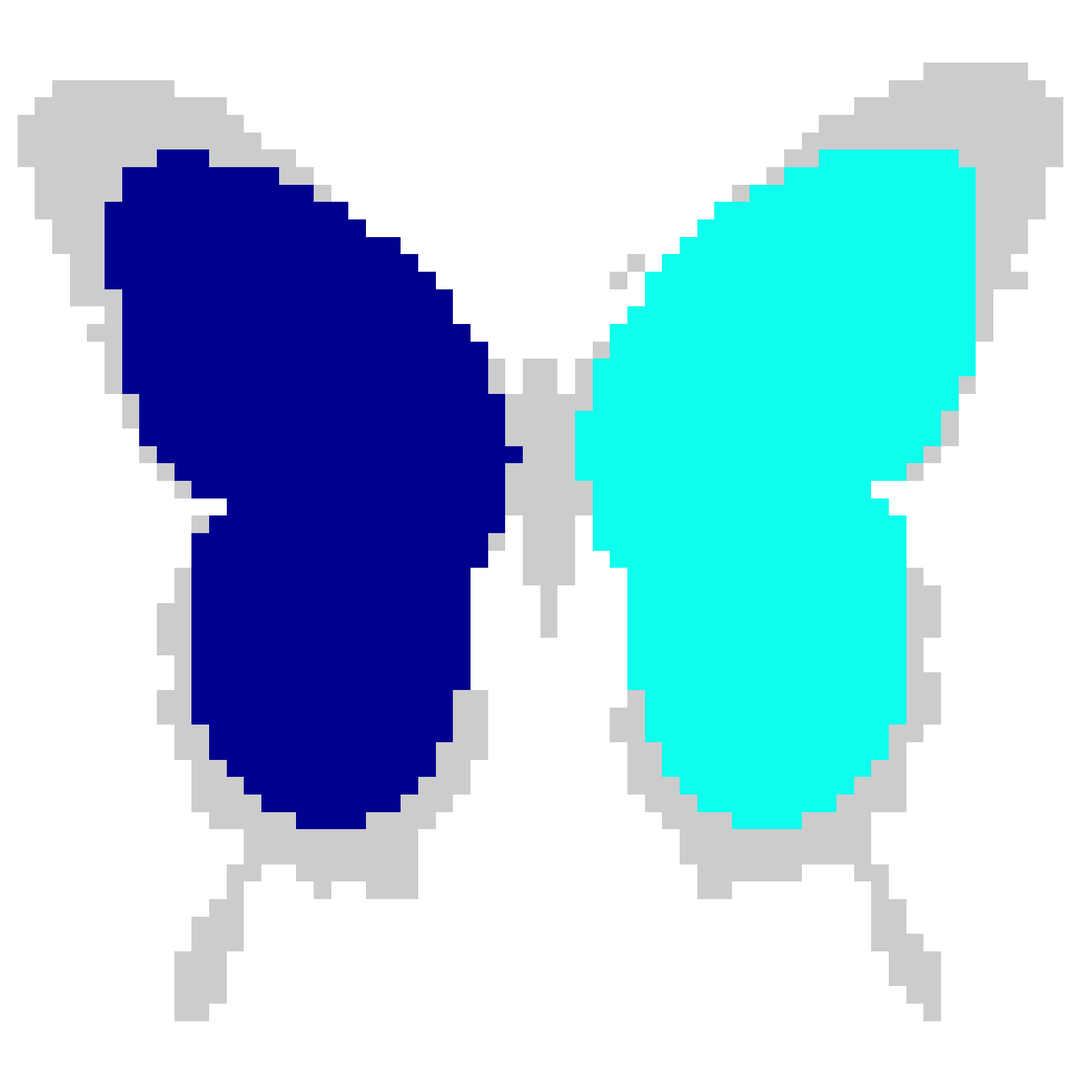}
\includegraphics[height=2.9cm]{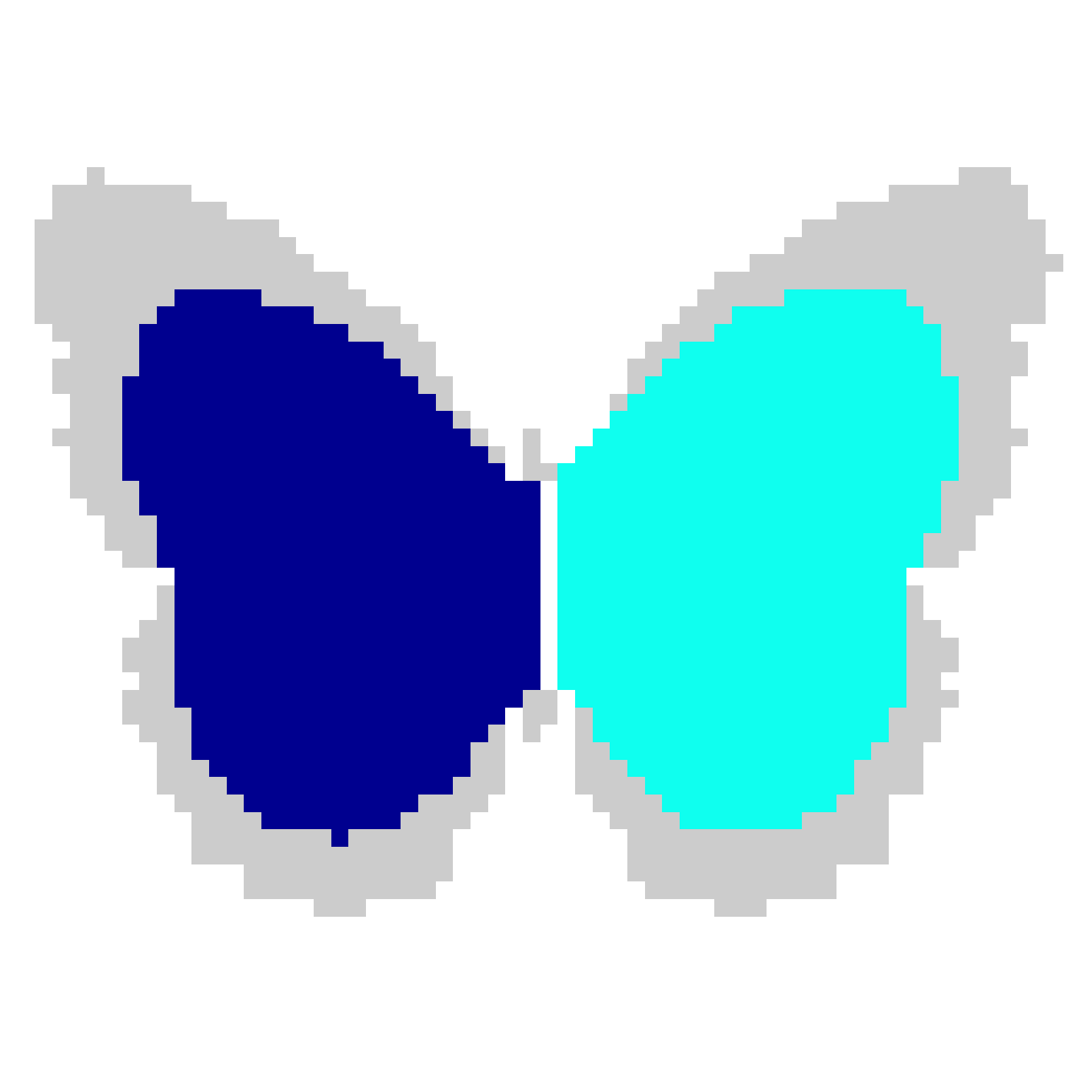}
\end{tabular}
\caption{The effect of reducing the importance of the global term.
$c=0.25$.} \label{fig:kelebek_param}
\end{figure}

\index{context}
Notice
that the restriction of the field $\omega$ to where the values are
less than a given threshold $-3$, in (f), indicates  the first
partitioning of the peripheral structure along the minor axis,
similar to the turtle case in Fig.~\ref{fig:turtle}.
  This indication  is consistent with the
result of the method of Zeng {\sl et al.}~\cite{Zeng07}  which
computes the partition line by sequentially eliminating the boundary
detail using Discrete Curve Evolution~\cite{Latecki99}.
\index{curve evolution!discrete}
\index{features!local} \index{features!global} \index{non-local} 
The new method is essentially a parameter free method
 with the assumption
that equal importance should be given to local and global terms.
  However, the other two
methods do not use global features (note that non-local is not
necessarily global). Thus, for a comparative evaluation purpose, it
is worth trying to reduce the effect of the global term by imagining
a constant weight $c < 1$ in front of $E_{Reg}^G$
in~\ref{eq:energy}. In Fig.~\ref{fig:prickly_param} (a-b), $c=0.5$.
One can notice the slight reduction of the peripheral region. The
global term is responsible for the balance between the negative
values and the positive values of $\omega$. As its importance
decreases, more pixels tend to attain positive values. In (c),
$c=0.125$. As $c$ decreases, the peripheral structure shrinks
further and the implied decomposition approaches to the one shown in
Fig.~\ref{fig:prickly} (c). In Fig.~\ref{fig:kelebek_param}, the
effect of reducing the importance of the global term is demonstrated
using butterfly shapes.

\begin{figure}[t]
\vglue -4pt
\centering
{\footnotesize\begin{tabular}{ccc}
\includegraphics[height=2.6cm]{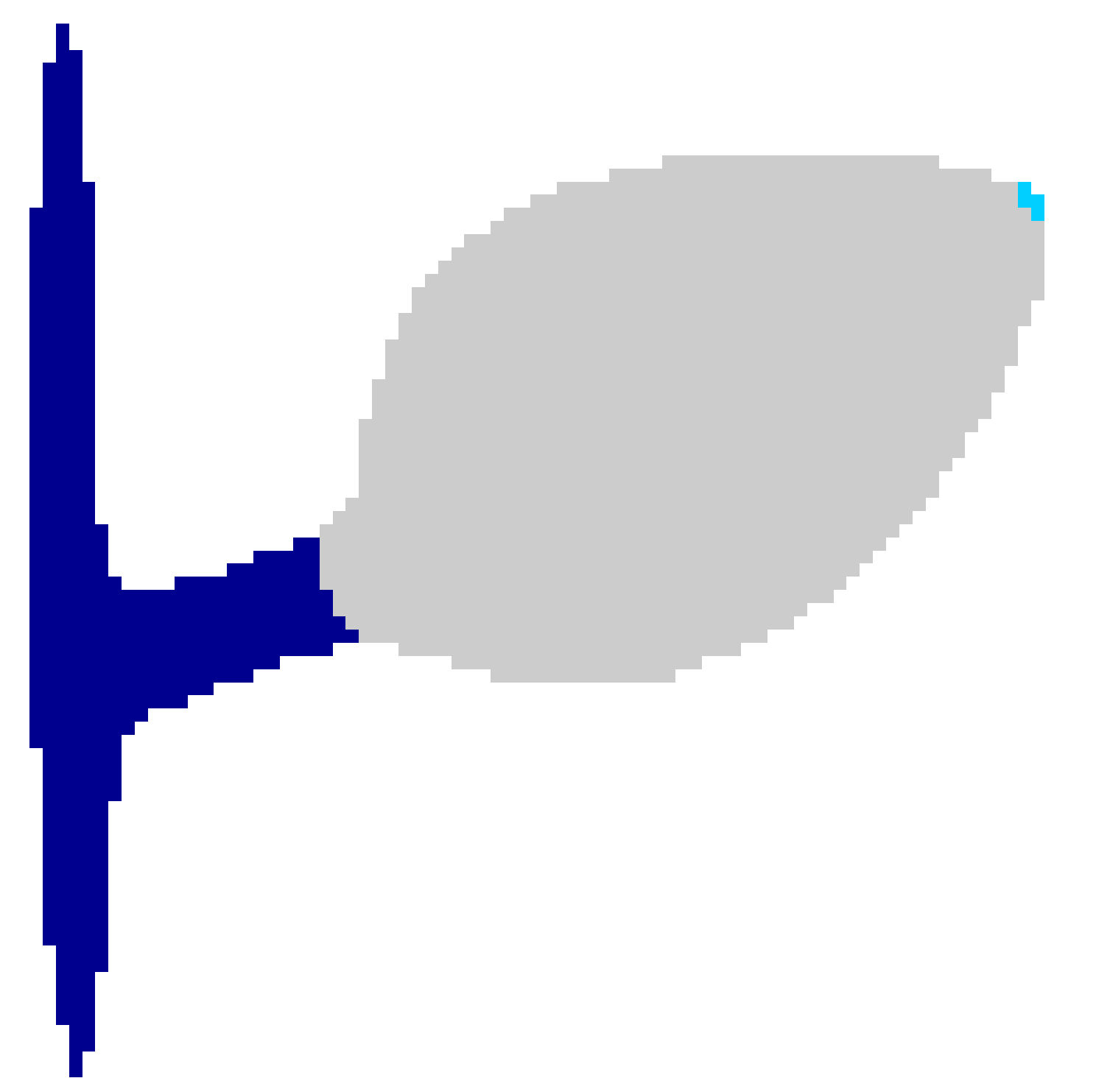}&
\includegraphics[height=3cm]{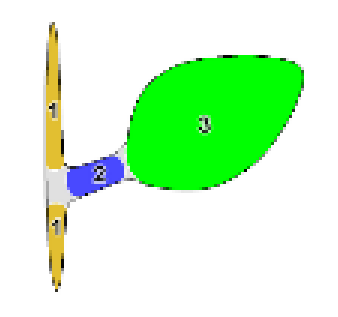}&
\includegraphics[height=2.7cm]{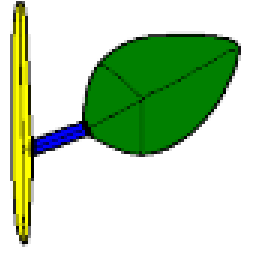}\\
(a)& (b)&  (c)
\end{tabular}}
\caption{The effect of significantly reducing the importance of the
global term.  (a) The decomposition with the new method, $c=0.025$.
(b-c) The decompositions presented by Mi and DeCarlo~\cite{Mi07} and
Zeng {\sl et al.}~\cite{Zeng07}, respectively. [(b-c) taken from
the original sources~\cite{Mi07,Zeng07}] } \label{fig:leaf}
\end{figure}

In Fig.~\ref{fig:leaf}, the importance of the global term is
significantly reduced by setting $c=0.025$. The decomposition result
using the new method is shown in (a). For reference, the decomposition
results by the previous methods are shown in (b-c).

\section{Connection to the methods of Tari,
Shah and Pien~\cite{Tari96,Tari97,Tari98},
Aslan and Tari~\cite{Aslan05,Aslantez,Aslan08}, and Gorelick {\sl et al.}}
\label{S2}
\index{PDE}
\index{PDE!Laplace operator}
\index{medial axis!Aslan and Tari}
\index{medial axis!Tari, Shah and Pien}

Recall that the field $\omega$ is the minimizer of the following energy :
\begin{equation*}
 E(\omega) = \sum_{i,j\in \Omega} E_{Reg}^G(\omega_{i,j}) + E_{Reg}^L(\omega_{i,j}) + w_{Bdy} E_{Bdy}(\omega_{i,j}) 
\end{equation*}
 
Let us omit the term $E_{Reg}^G$ that models the global interaction among the
shape pixels to obtain a reduced energy:
\begin{equation}
\sum_{i,j\in \Omega} - \left(  \omega_{i+1,j} \cdot \omega_{i-1,j} +
\omega_{i,j+1} \cdot \omega_{i,j-1} \right)
+ w_{Bdy} \left( \omega_{i,j} - t_{i,j} \right)^2
\label{eq:reducedE}
\end{equation}

By setting the first derivative  of (\ref{eq:reducedE}) with respect
to $\omega_{i,j}$ equal to $zero$, the condition satisfied by the
minimizer is obtained as:
\begin{equation}
(w_{Bdy}-2 )\omega_{i,j} - w_{Bdy} t_{i,j} +4 \omega_{i,j}  - \omega_{i-1,j} -\omega_{i+1,j}-\omega_{i,j-1}-\omega_{i,j+1}=0
\end{equation}

Letting  $(w_{Bdy}-2 ) =\alpha >0 $ gives
\begin{equation}
 \left( 4 \omega_{i,j}  - \omega_{i-1,j} -\omega_{i+1,j}-\omega_{i,j-1}-\omega_{i,j+1} \right)
- \alpha \omega_{i,j} = \left( \alpha +2 \right) t_{i,j}
\label{eq:reduced_cond}
\end{equation}

(\ref{eq:reduced_cond}) is clearly the discretization using central
difference approximation, of the PDE (\ref{eq:tsp_withf}) given
below:
\begin{eqnarray}
\label{eq:reduced} \left( \bigtriangleup  - \alpha  \right) w(x,y)
&=& f(x,y) \label{eq:tsp_withf} \\
  {\mbox{with }} w({\mathbf x}) &=0&  \mbox{ for } {\mathbf x}=(x,y)\in {\partial \Omega} \nonumber
\end{eqnarray}
where  $ \bigtriangleup$ denotes the Laplace operator, and the right hand side inhomogeneity $f(x,y)$ is a scaled distance transform.
(\ref{eq:tsp_withf})  is defined on a planar shape domain $\Omega$
which is a connected, bounded, open domain of ${\mathbf R^2}$.

Interestingly, when the right hand side inhomogeneity
is replaced with a constant function $f(x,y)=1$ and
 $\alpha$ is set to {\sl zero} (i.e. $w_{Bdy}=2$ as in the new method)
 one  obtains the Poisson equation which has been recently proposed by
 Gorelick {\sl et al.}~\cite{Gorelick06} as a shape representation tool.
 On the other hand, when $\alpha>0$ and  $f(x,y)=-\alpha$, one
 obtains the PDE:
\index{PDE!Poisson equation}

\begin{eqnarray}
 \left( \bigtriangleup  - \alpha  \right) v &=&-\alpha \label{eq:tsp}
\\
  {\mbox{with }} v({\mathbf x}) &=0&  \mbox{ for } {\mathbf x}=(x,y)\in {\partial \Omega} \nonumber
\end{eqnarray}
which  has been  proposed earlier by Tari, Shah and
Pien~\cite{Tari96,Tari97,Tari98}. The qualitative behavior of
the $v$ function for small $\alpha$ is essentially the same with
that of the function obtained by solving the
Poisson~\cite{Gorelick06} equation. Both of them are essentially
weighted distance transforms~\cite{MaragosButt2000}
where the local steps between neighboring points are given different
costs. Tari, Shah and Pien~\cite{Tari96} have initially proposed $v$ function as a
linear and
 a computationally efficient  alternative
 to the curve evolution scheme by Kimia, Tanenbaum and Zucker~\cite{Kimia95} by
 showing that the successive level curves of $v$ mimic the motion of curves with
 a curvature dependent speed in the direction of the inward normal~\cite{Tari96}.
\index{curve evolution} \index{curve evolution!efficient
alternative} \index{curve evolution!Kimia, Tanenbaum and Zucker} \index{distance transform}
 Furthermore, they
 demonstrated that the gradient of $v$ along a level
curve approximates the curvature of level curves, thus, suggesting a
robust method for skeleton extraction by locating the extrema of the
gradient of $v$ along the level curves. The skeleton computation
method of Tari, Shah and Pien~\cite{Tari96,Tari97,Tari98} exploits
the connection among morphology, distance transforms and fronts
propagating with curvature dependent speeds. Such connections have
stimulated many interesting approaches in solving shape related
problems~\cite{MaragosButt2000}. The
importance of Tari, Shah and Pien approach is that it is the first
attempt to unify segmentation and local symmetry computation into a
single formulation by exploiting the connection between
(\ref{eq:tsp}) and the  Mumford and Shah~\cite{Mumford89}
segmentation functional (via its Ambrosio and
Tortorelli~\cite{Ambrosio90} approximation). It naturally extends to
shapes in arbitrary dimension~\cite{Tari98}.  (In a related publication~\cite{Tari09}, the author
introduces an additive normalization term to (\ref{eq:tsp}) which
forces the solution to oscillate, yielding the same boundary texture
and gross structure separation. The proposed method is connected to
a variety of morphological ideas as well as to the method of Tari,
Shah and Pien in the variational calculus and PDE setting.) \index{segmentation}
\index{morphology} \index{Ambrosio and Tortorelli} \index{Mumford
and Shah} \index{level curve}

\index{diffusion}
\index{shape!topology}
\index{shape!evolution}

\begin{figure}
\centering
{\footnotesize \begin{tabular}{ccc}
\includegraphics[height=8cm]{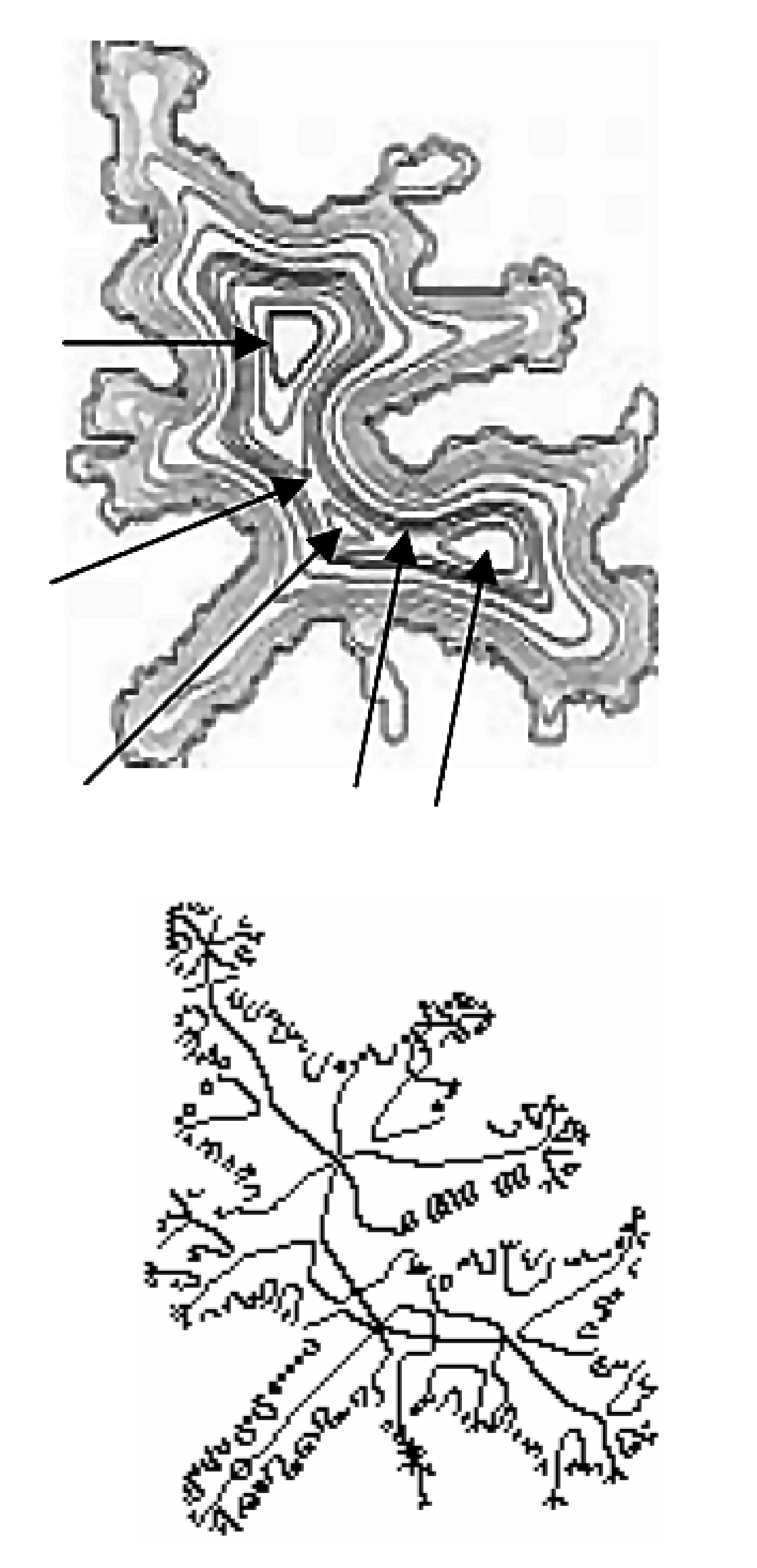} &
\includegraphics[height=8cm]{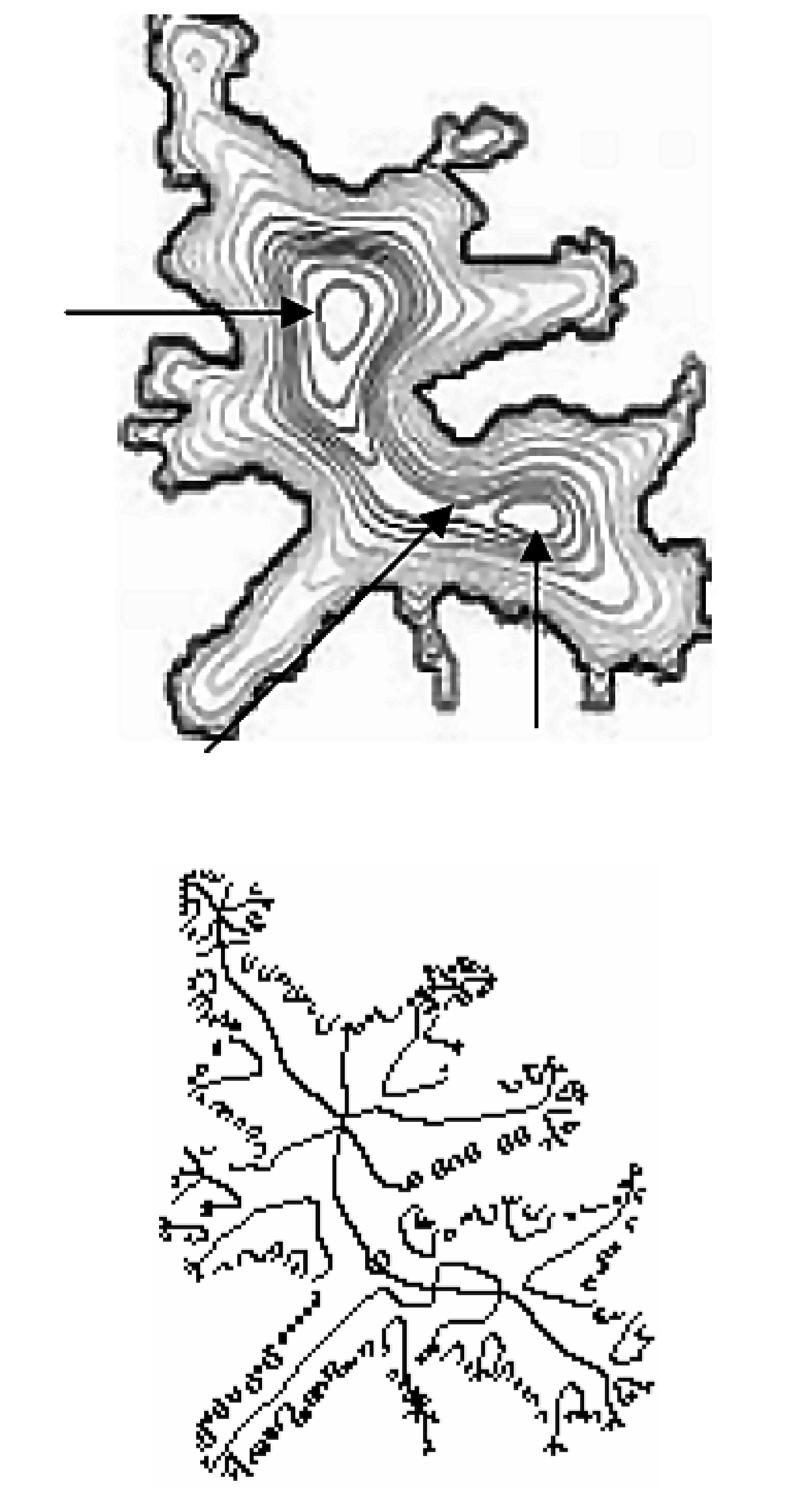} &
\includegraphics[height=8cm]{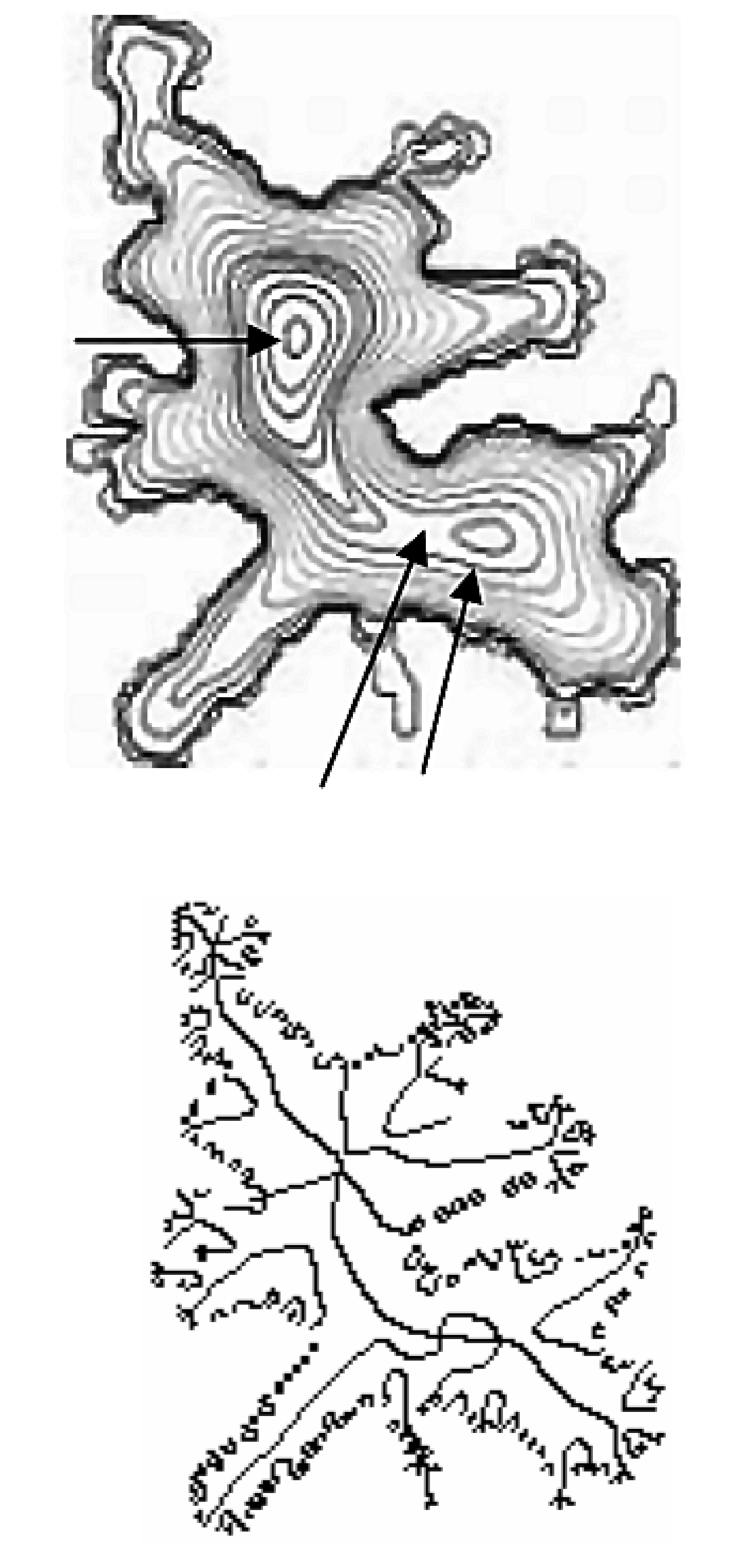} \\
(a) & (b) & (c)
\end{tabular}}
\caption{ Tari, Shah and Pien method using three different $\alpha$
values. (a) $\alpha=1/{4^2}$  (b) $\alpha=1/{8^2}$ (c)
$\alpha=1/{16^2}$. The top row (level curves of $v$) illustrates
 different  topological interpretations  by varying
$\alpha$. The arrows indicate the maxima and the saddle points. The
bottom row displays the skeletons  computed from the respective $v$
functions. [Figures by C. Aslan~\cite{Aslantez,Aslan08}].}
\label{fig:tsp_mri1}
\end{figure}

In Fig.~\ref{fig:tsp_mri1}, the basic method of Tari, Shah and Pien
is illustrated using an example by C. Aslan~\cite{Aslantez,Aslan08}.
At the top row, the level curves of $v$ function,
mimicking the behavior of fronts propagating with curvature
dependent speed, are shown. In each column, a different $\alpha$
value is used when computing $v$ via (\ref{eq:tsp}). As $\alpha$
decreases from left to right, the relative speed of the high
curvature points increases. Consequently, the inner level curves
lose their concavities and become smoother earlier. Thus the value
of $\alpha$ determines the level of smoothing (or {\sl diffusion}).
The arrows indicate the maxima and the saddle points. Topological
interpretation of the shape varies with $\alpha$.  When, in the
first column, $\alpha=1/{4^2}$, the evolving shape boundary gets
split into three curves, each of which shrinks into three distinct
maxima separated by the two saddle points, indicating three parts.
In (b) and (c) $\alpha$ is reduced to $1/{8^2}$ and
$1/{16^2}$, respectively. The level curves lose their
concavities (which imply parts) earlier and evolving curves split
into two parts instead of three.

%
\begin{figure}
\centering
{\footnotesize\begin{tabular}{cc}
\includegraphics[height=4.3cm]{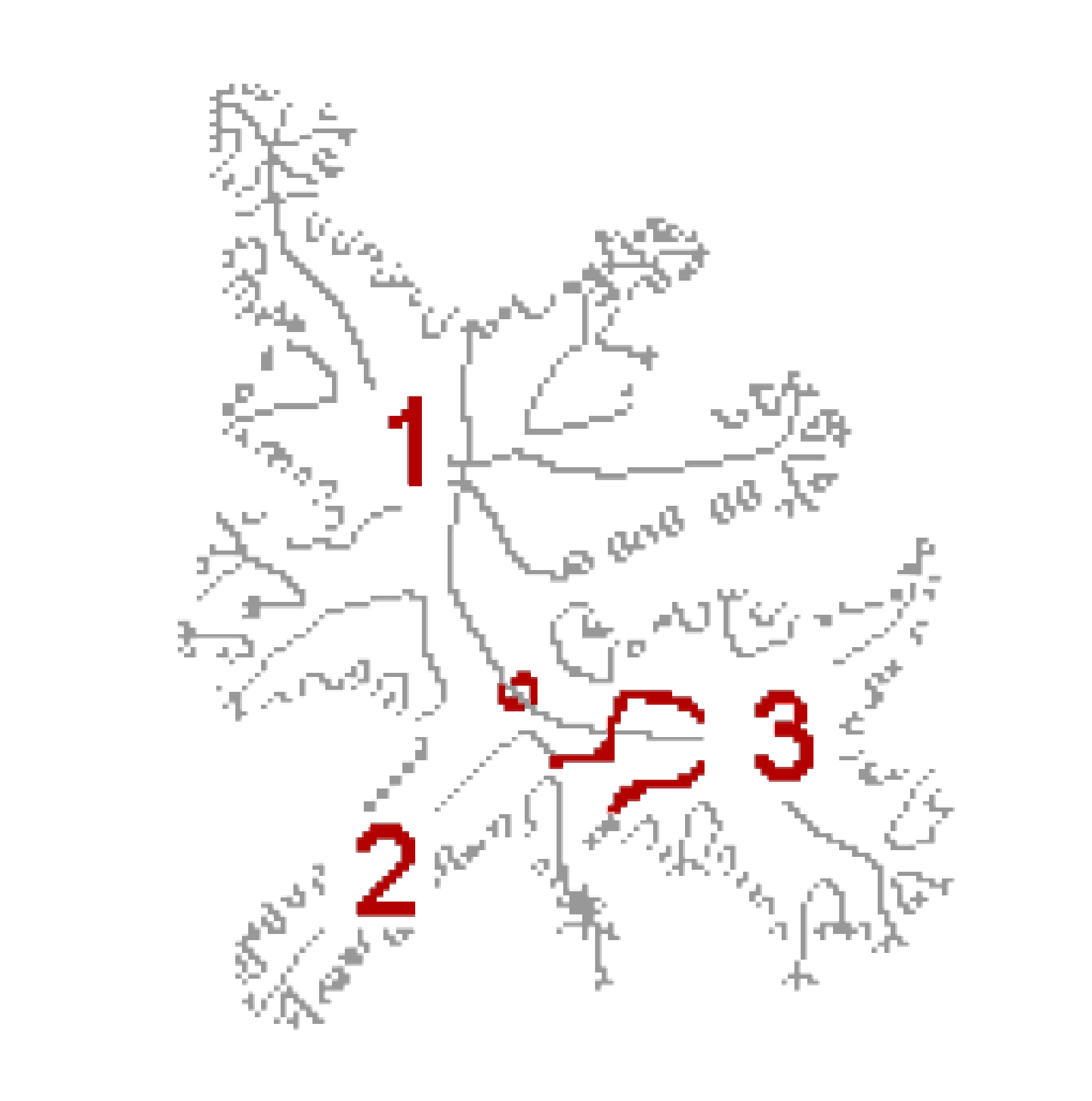} &
\includegraphics[height=4.3cm]{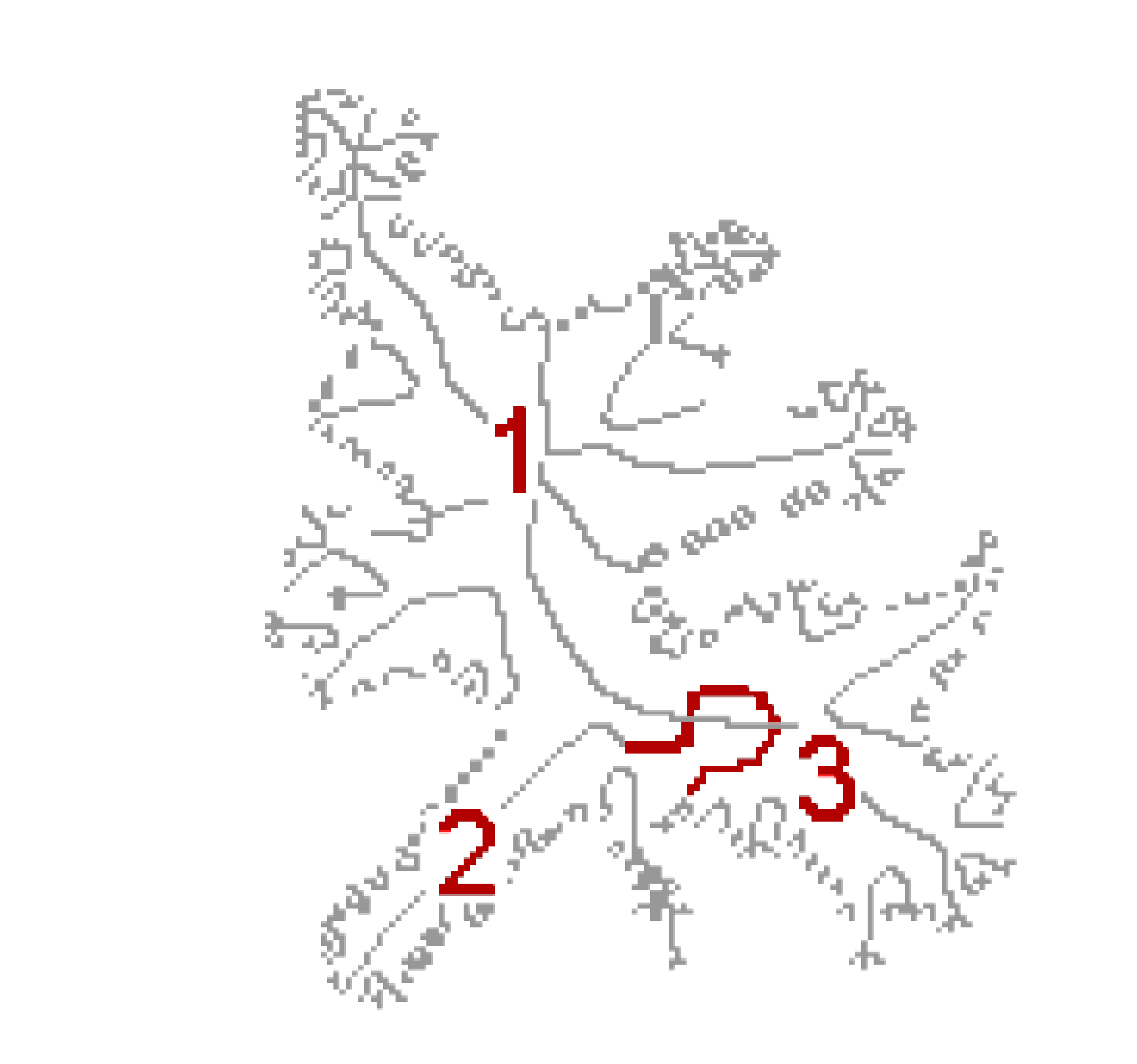} \\
 (a) & (b) \\
\end{tabular}}
\caption{ Un-intuitive skeleton branches (red color) in Tari, Shah
and Pien~\cite{Tari96,Tari97}. (a) $\alpha=1/{8^2}$ (b)
$\alpha=1/{16^2}$. See text for discussion. [Figures by C.
Aslan~\cite{Aslantez,Aslan08}].} \label{fig:tsp_mri2}
\vglue 6pt
\end{figure}

Skeleton branches computed from the respective $v$ functions
(displayed at the bottom row) typically track the evolution of the
indentations and protrusions of the shape. However, some of them exhibit a
pathological behavior that frequently occurs in the Tari, Shah and
Pien method when a limb is close to a neck. Notice that the
skeletons contain branches that do not correspond to any protrusion
or indentation of the shape. Such branches are marked with red color
in Fig.~\ref{fig:tsp_mri2}. Aslan and
Tari~\cite{Aslan05,Aslantez,Aslan08} claim that the reason of this
pathology is {\sl insufficient diffusion}. \index{diffusion}

\index{shape!topological change}
As seen in Fig. \ref{fig:tsp_mri2} increasing the amount of
diffusion by decreasing $\alpha$ makes such branches disappear. In
(a), the computation stopped while the shape was transforming from a
shape with three major blobs to a shape with two major blobs. The
circular branch colored with red is due to the interaction of the
center of parts two  and the neck between parts one and two. As
shown in (b), increasing the amount of diffusion by decreasing
$\alpha$ makes this branch  disappear since the topological change
is complete. This time, the shape is between the state with two
blobs (parts one and two together and part three) and the state with
one blob. The red branch is due to the interaction of the center
point of part
 three and the neck  between parts two and three.

Thus, as a remedy, they propose to increase the diffusion by
gradually decreasing
 $\alpha$ so that almost every shape is forcefully
interpreted as a single blob  ignoring the part structure.
Following this ad-hoc modification, the new $v$ function has been successfully applied in shape matching applications~\cite{Aslantez,Aslan08,Baseski08,Erdem09}.

However, the method can not be applied to shapes which can not be
reduced to a single blob. Such cases include:
\begin{itemize}
\item{shapes with holes;}
\item{thin and long shapes with constant width; }
\item{shapes with more than one equally prominent parts.}
\end{itemize}

The strategy adopted~\cite{Aslan05,Aslantez,Aslan08} for shapes with
two equally prominent parts (dumbbell-like shapes) is to retain
their dumbbell-like character. This ad-hoc solution introduces a
representational instability as the width of the neck that separates
two prominent parts change.

\begin{figure}[ht]
\vglue 6pt
\centering
\begin{tabular}{ccc}
\includegraphics[height=4.5cm]{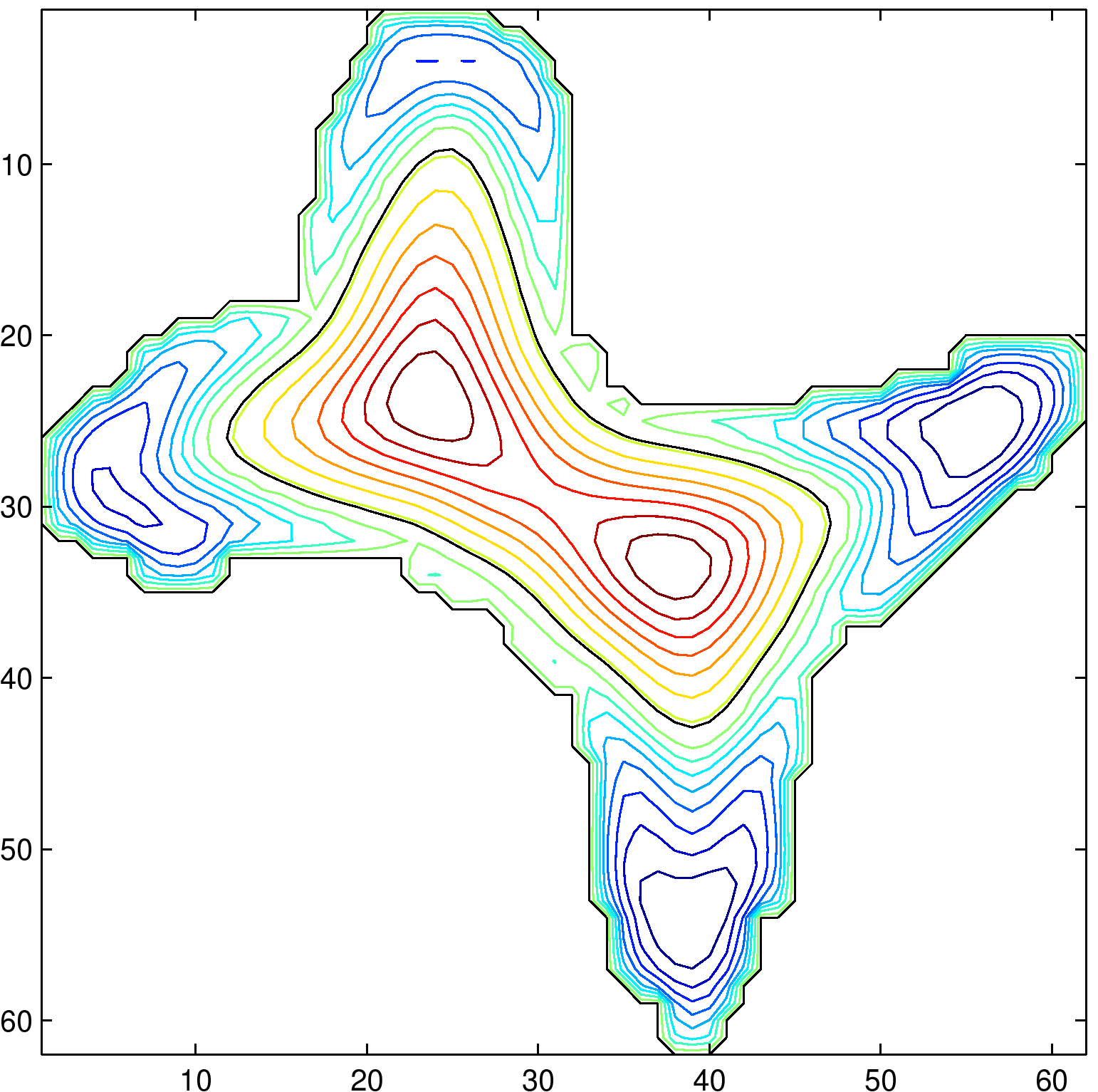}&
\includegraphics[height=4.5cm]{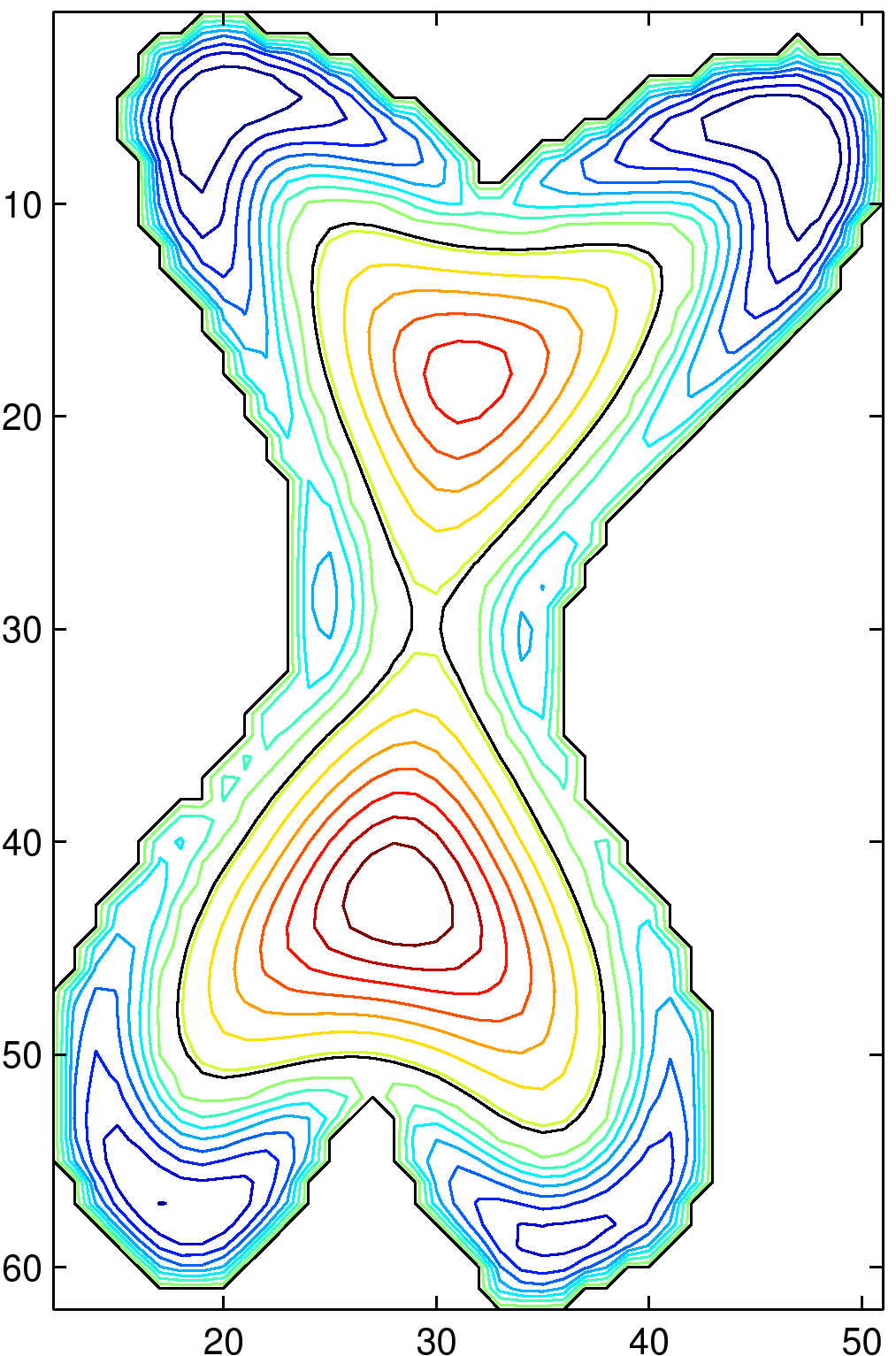}&
\includegraphics[height=4.5cm]{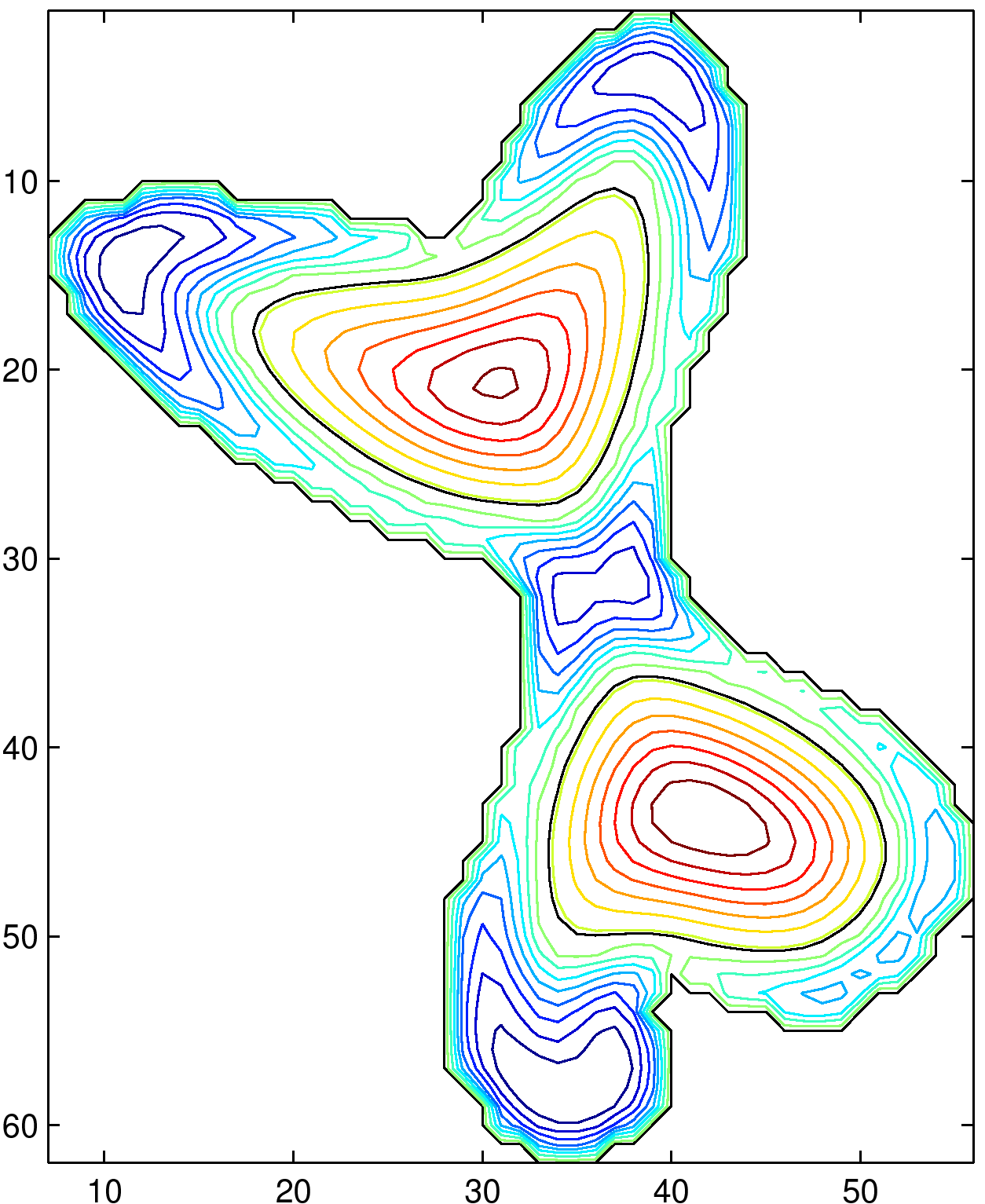}
\end{tabular}
\caption{The level curves of  $\omega$. The inner black level curve
is the zero-level curve. In all of the three cases, $\Omega^+$
denotes the gross structure.} \label{fig:dumbell_lc}
\end{figure}
\vfill\pagebreak

\begin{figure}[t]
\centering
\begin{tabular}{ccccc}
\includegraphics[height=3.6cm]{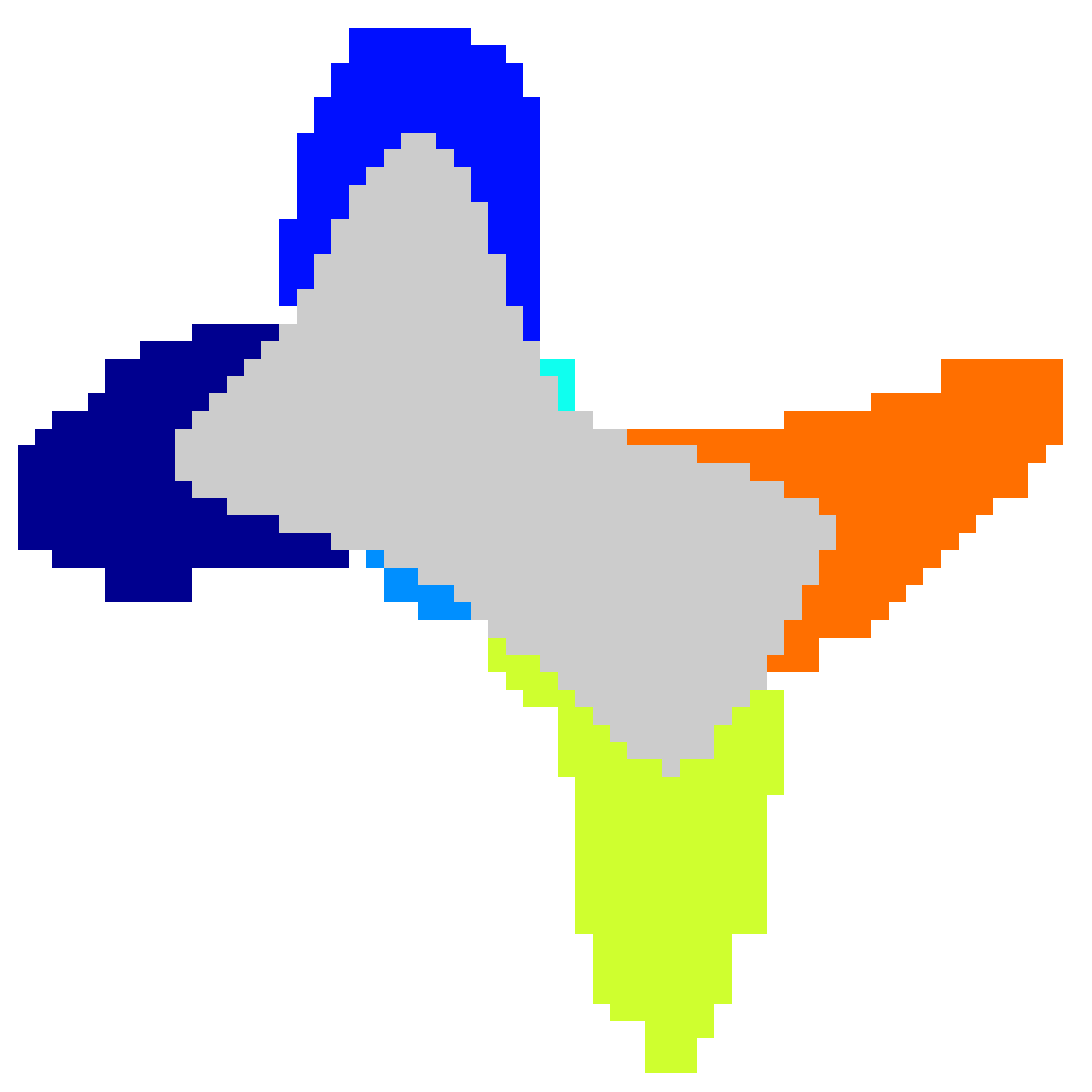}&
\includegraphics[height=3.6cm]{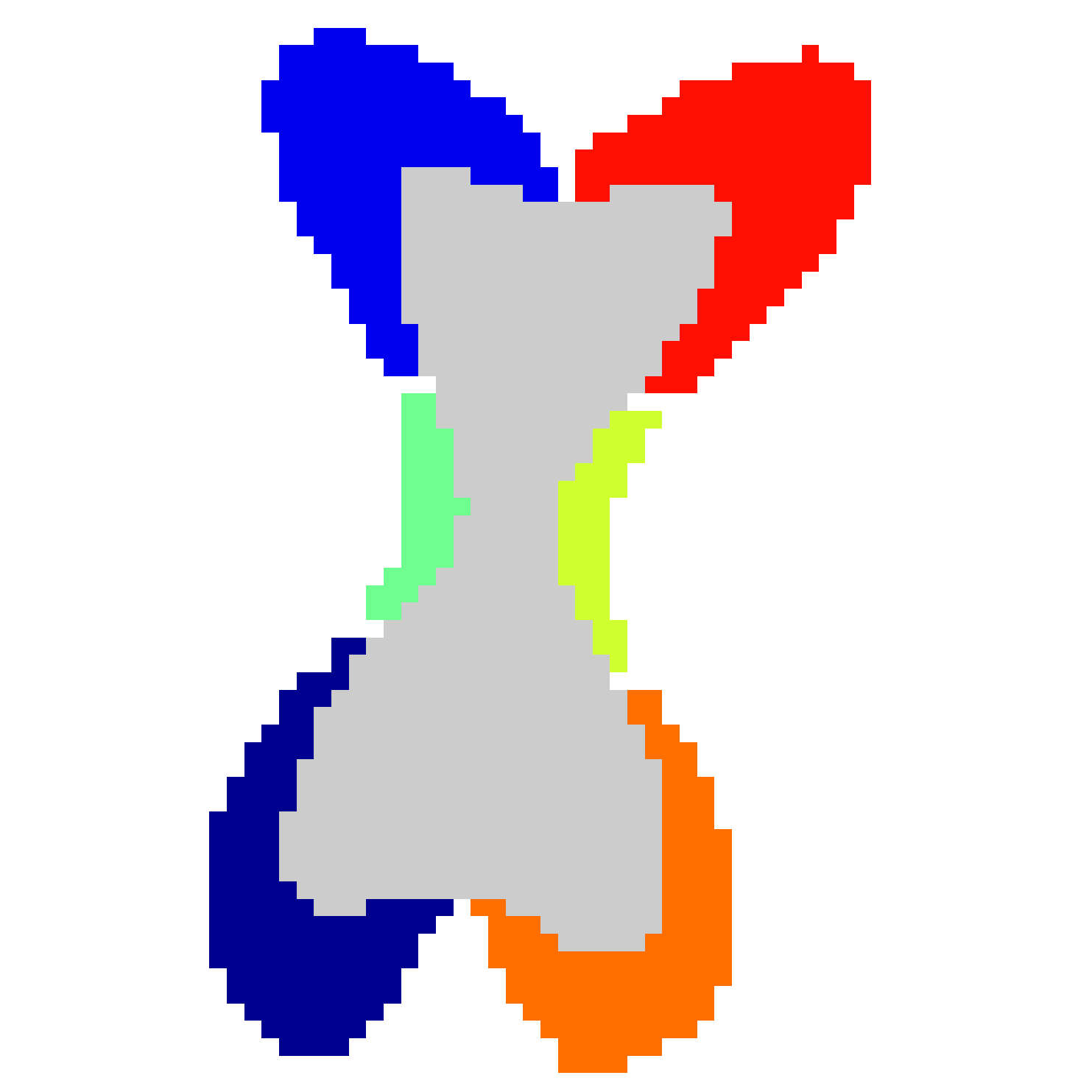}&
\includegraphics[height=3.6cm]{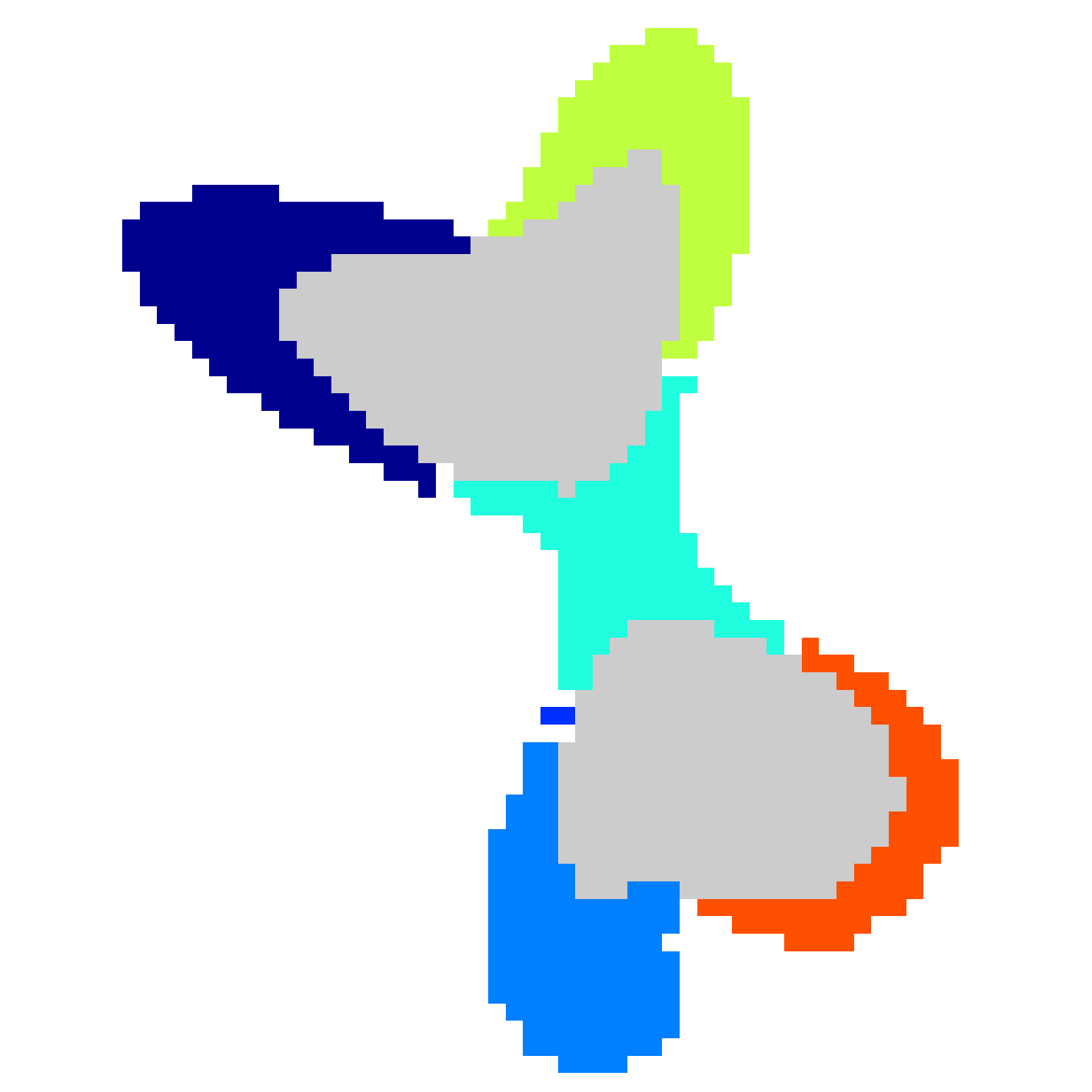}
\end{tabular}
\caption {Parts of the peripheral structure for dumbbell-like shapes
of varying neck thickness. } \label{fig:dumbell_parts}
\end{figure}

In Fig.~\ref{fig:dumbell_lc}-\ref{fig:dumbell_parts}, the level
curves of $w$ are shown for three dumbbell-like shapes with varying
neck thickness. In all of the three cases, $\Omega^+$ denotes the
gross structure. Instead of relying on a single point to be  the
shape center as in the method of Aslan and Tari~\cite{Aslan05,Aslantez,Aslan08},
the new method takes a different attitude. Robustness
is obtained by replacing a point estimate for the center with an
interval estimate. Parts of the peripheral structure are depicted in
Fig.~\ref{fig:dumbell_parts}.

\begin{figure}[b]
\centering
{\footnotesize\begin{tabular}{cc}
\includegraphics[height=2.5cm]{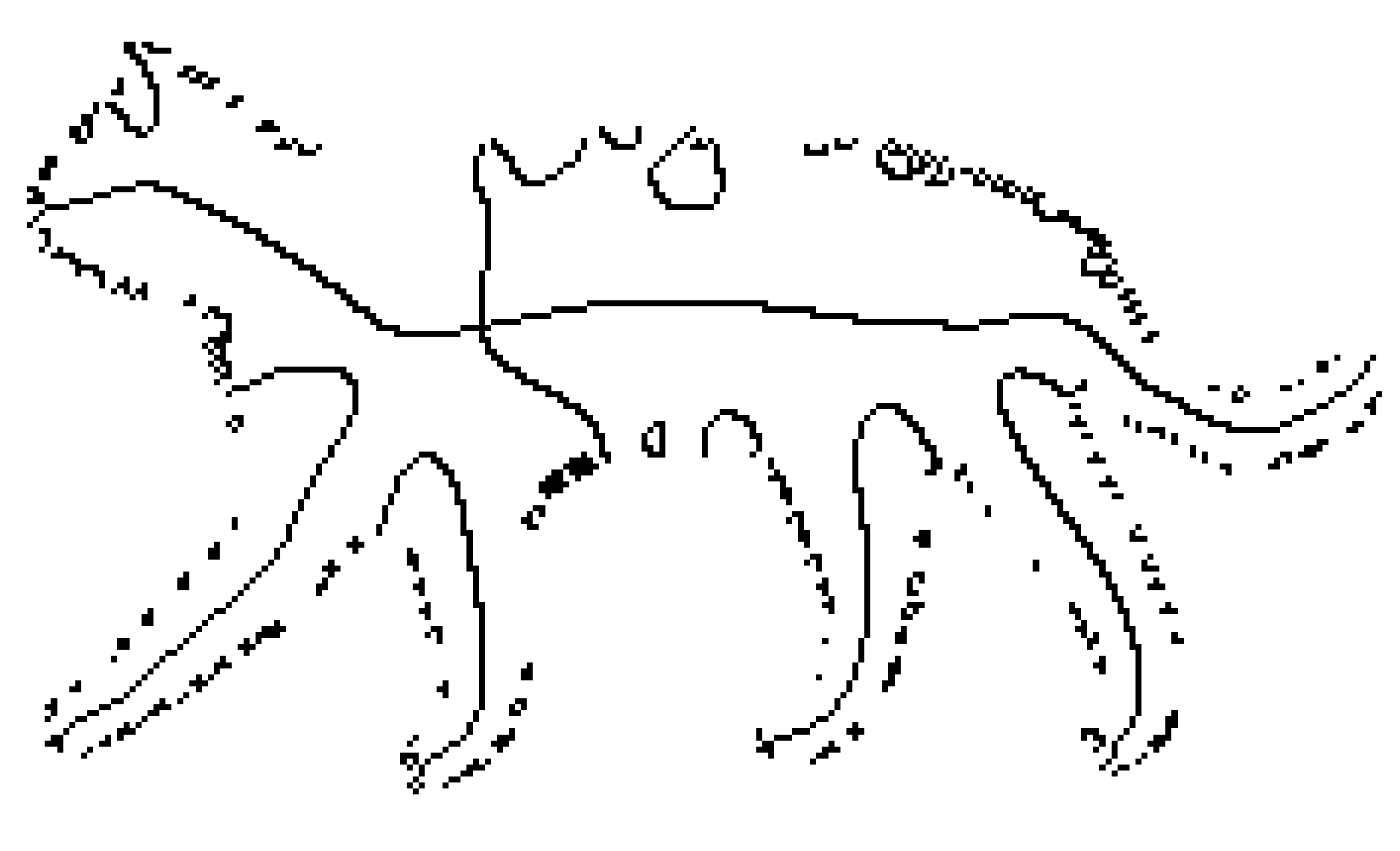} &
\includegraphics[height=2.5cm]{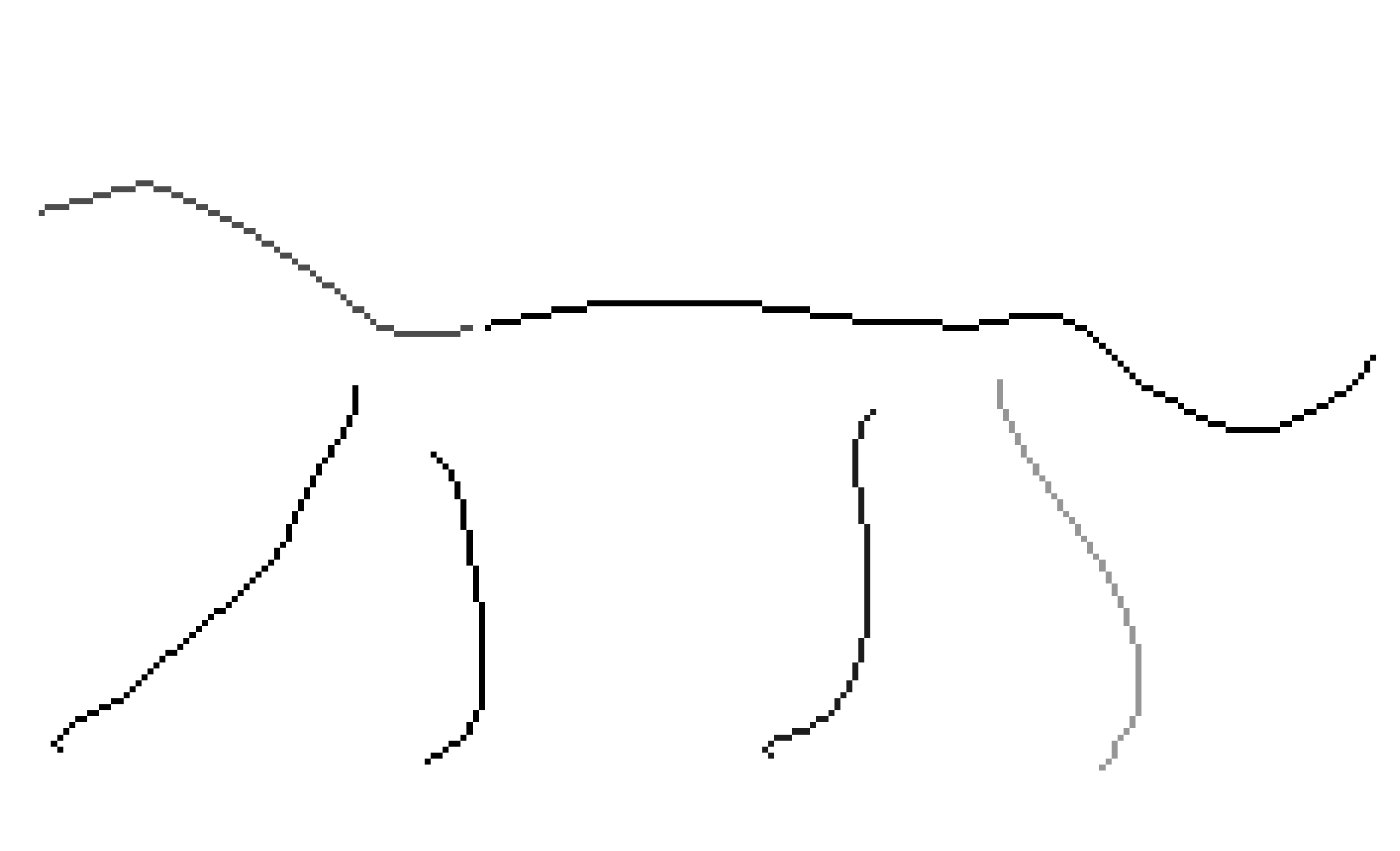} \\
(a) & (b) \\[4pt]
\includegraphics[height=2.5cm]{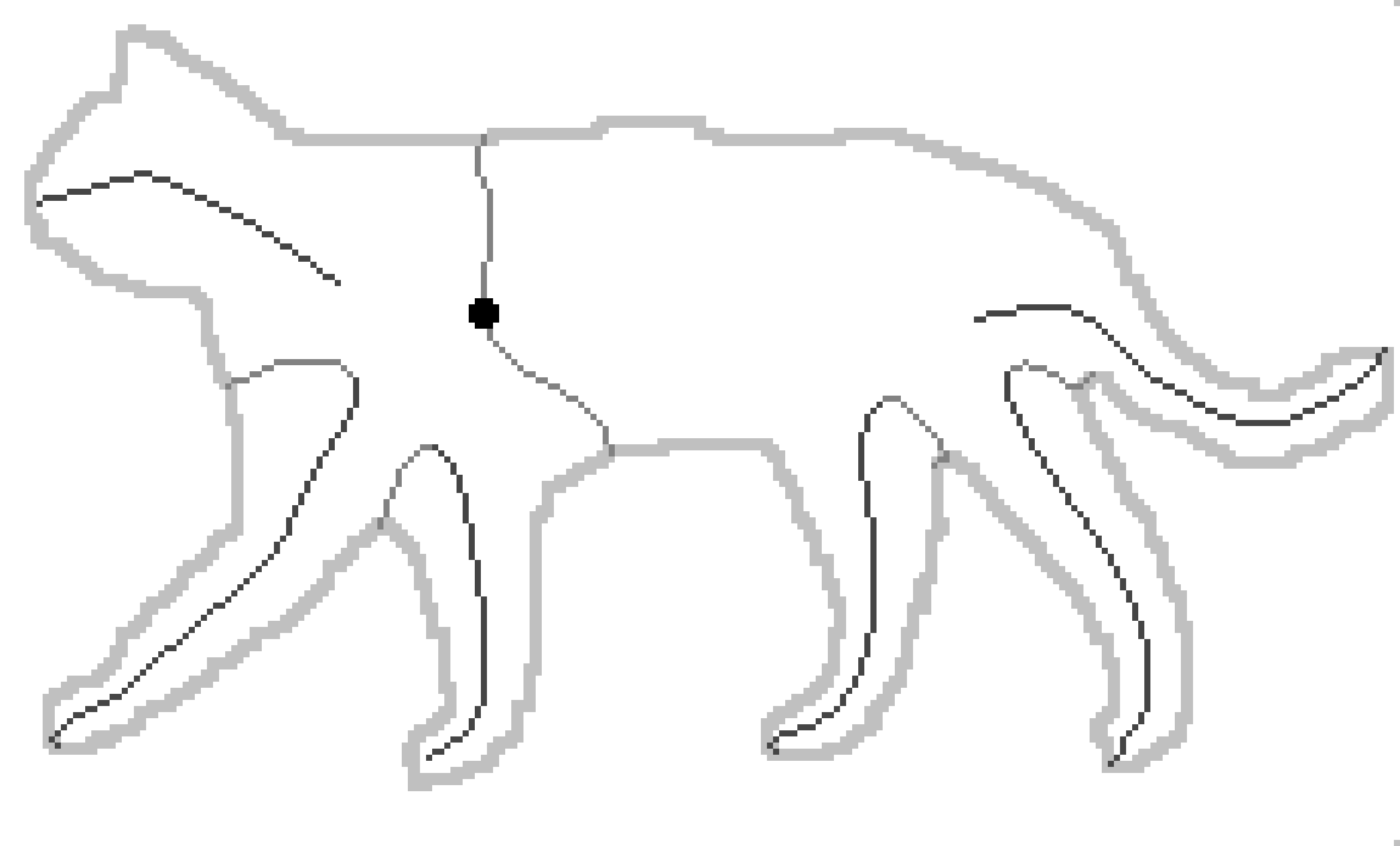}&
\includegraphics[height=2.5cm]{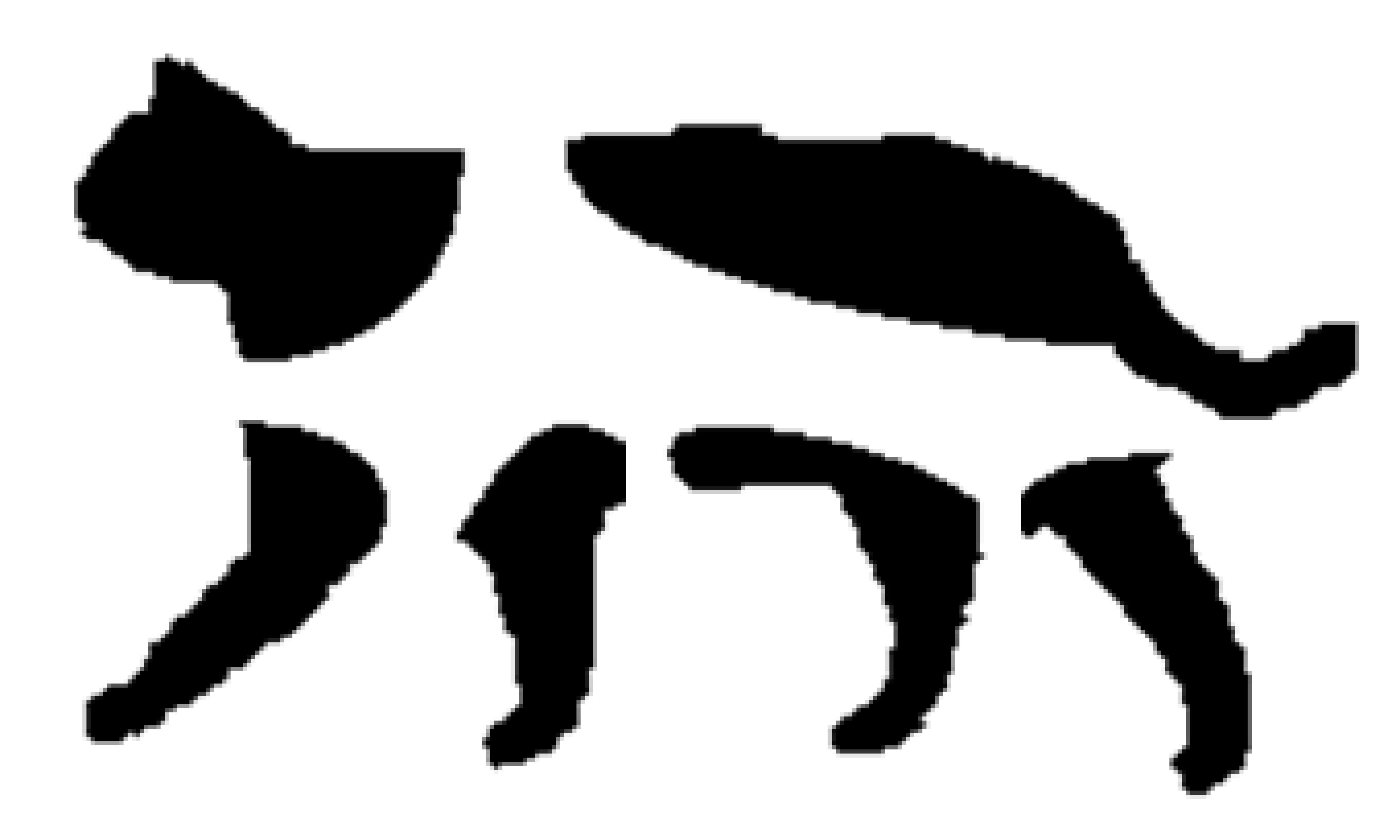}\\
(c) & (d)
\end{tabular}}
\caption{(a) Skeleton points detected with the method of Tari, Shah
and Pien~\cite{Tari97} using the modified
function~\cite{Aslan05,Aslantez,Aslan08}. (b) After pruning and
grouping skeleton points. (c) Disconnecting the major skeleton
branch~\cite{Aslan08}. (d) Parts obtained from disconnected
branches.  [Unpublished result by the author's former student  E. Baseski.] }
\label{fig:Emrekedi}
\end{figure}

\begin{figure}[t]
\centering
\begin{tabular}{cc}
\includegraphics[height=2.5cm]{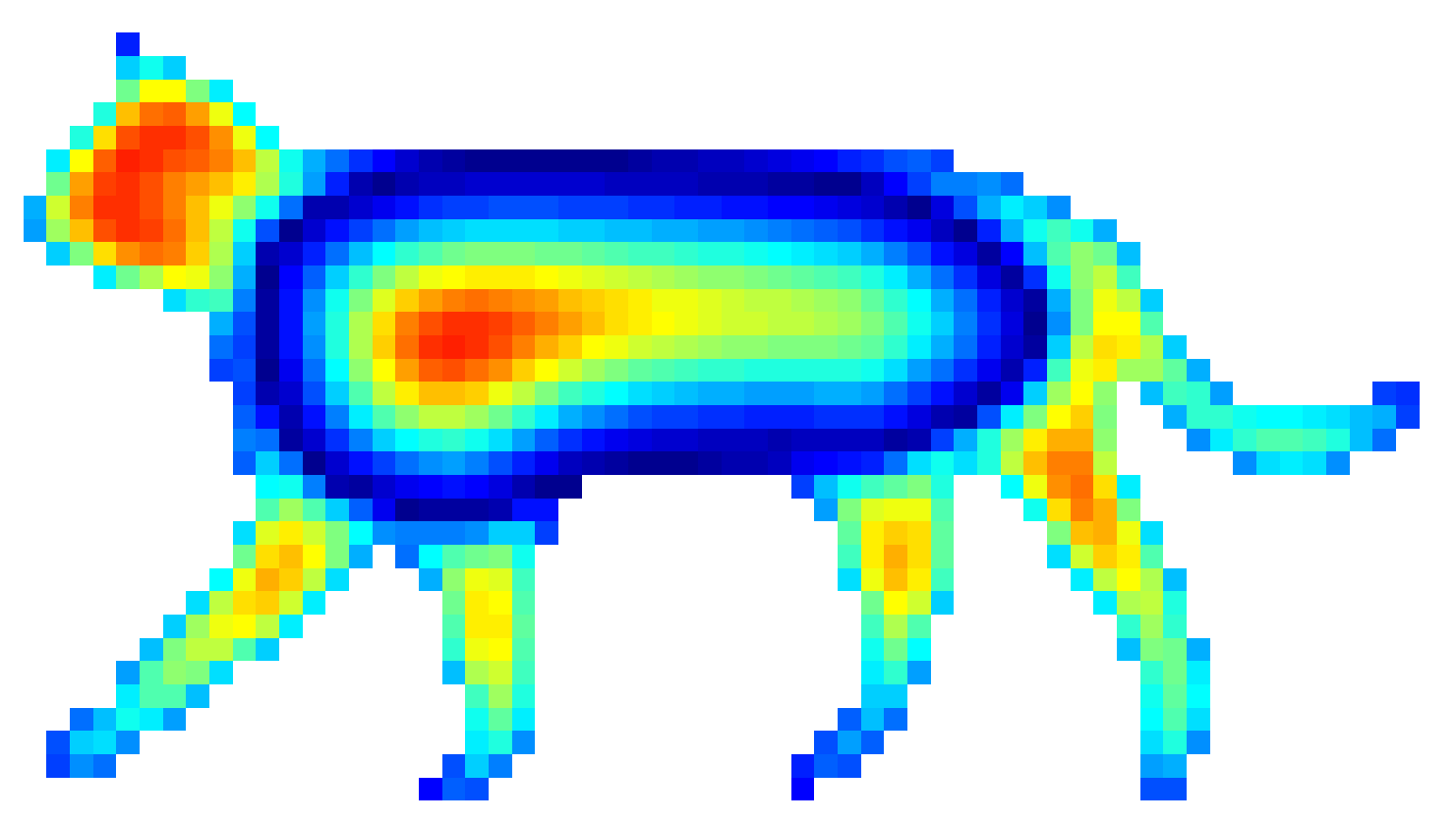} &
\includegraphics[height=2.5cm]{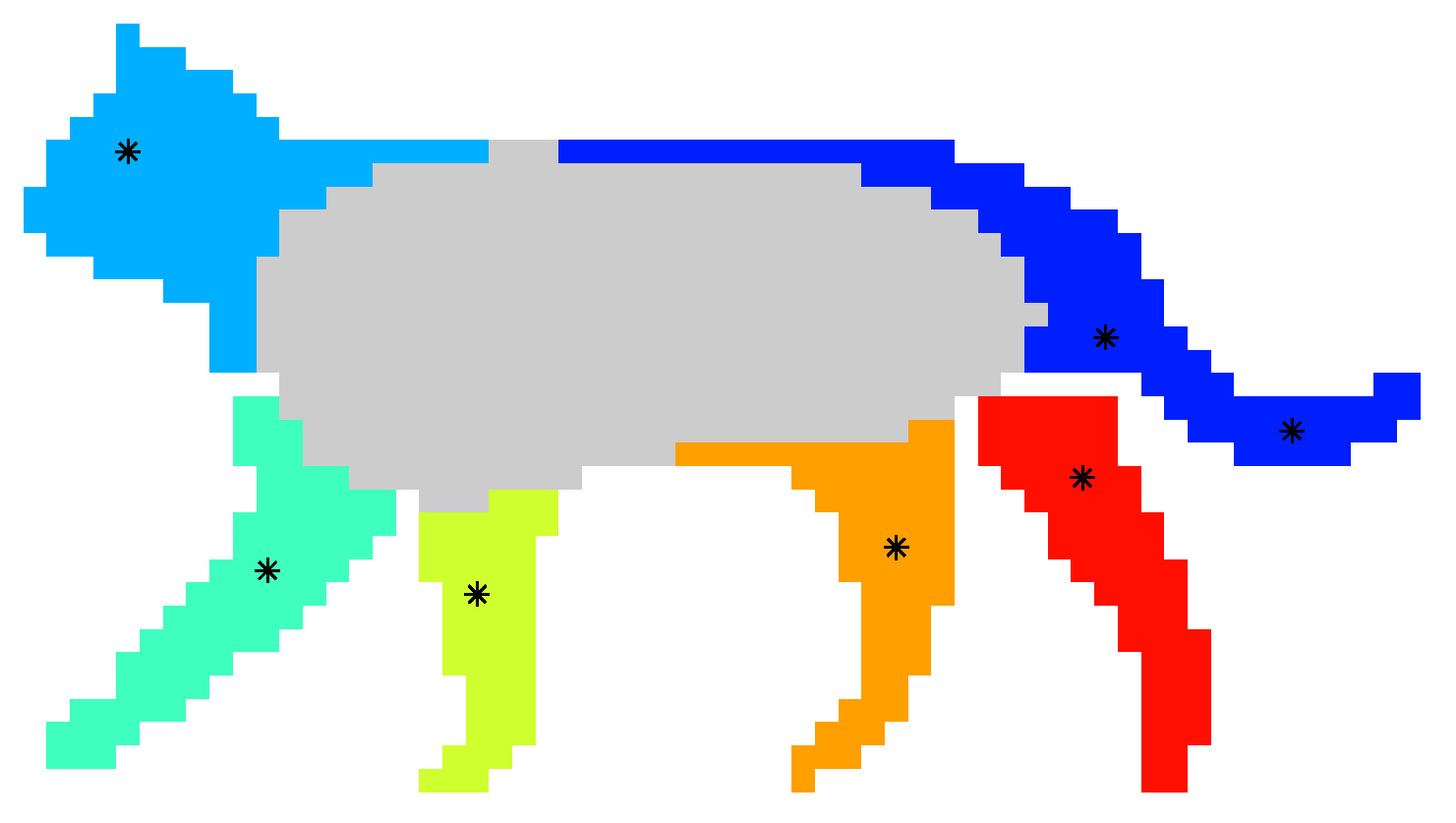}\\
\end{tabular}
\caption{(a) $|\omega|$. (b) Parts.} \label{fig:kediparts}
\end{figure}

\index{medial axis!disconnected skeleton}
The method of Aslan and Tari  implicitly
codes the part structure via disconnected skeleton branches.
Fig.~\ref{fig:Emrekedi} demonstrates the possibility of extracting
parts from these disconnected branches. In (a) skeleton points
detected with the method of Tari, Shah and Pien~\cite{Tari97} using
the modified function~\cite{Aslan05,Aslantez,Aslan08} is shown. In
(b) the result of pruning and grouping procedure is shown.  In (c)
final representation (called the disconnected Aslan
skeleton~\cite{Aslan05,Aslan08,Aslantez} to distinguish it from the
Tari, Shah and Pien skeleton~\cite{Tari96,Tari97}) is depicted. In
(d) parts are extracted, for each disconnected branch, by fitting a
spline that passes through the disconnection point and the nearest
indentations. The parts captured by the new method is presented in
Fig.~\ref{fig:kediparts}  for the same cat shape (reduced
resolution) for comparison.


\printindex
\end{document}